\documentclass[english, 
               brazil, 
               msc] 
               {dcomp-abntex2}

\usepackage{hyperref}
\usepackage{url}
\usepackage{subcaption}
\usepackage{lscape}   
\usepackage{pdfpages}
\usepackage{ragged2e}
\usepackage{amssymb}

\makeindex

\begin{document}

\selectlanguage{english}

\frenchspacing 

\pretextual


\titulo{Analysis and evaluation of deep learning based super-resolution algorithms to improve performance in low-resolution face recognition}

\autor{Angelo Garangau Menezes}
\orientador{Carlos Alberto Estombelo-Montesco}
\curso{Computer Science}

\imprimircapa
\imprimirfolhaderosto*
\newpage
\includepdf{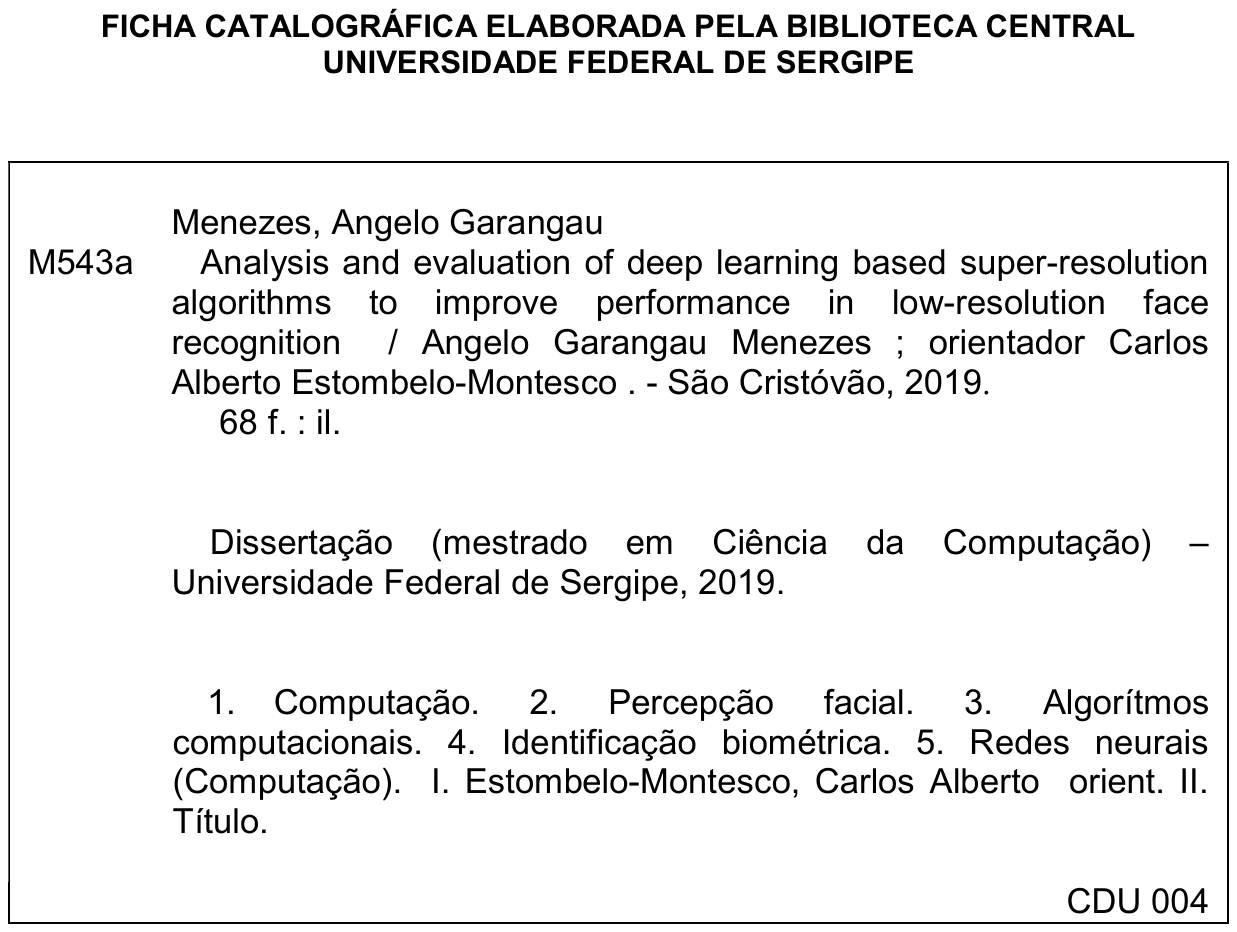}
\newpage
\includepdf{Pre_Textual/ata_angelo.pdf}

    
\begin{agradecimentos}

Firstly, I would like to say that I am thankful to God for creating this perfect simulation that we live, and for supporting me to get this far in this incredible adventure that we call life.

To my family, Roberto, Adilma, and Apolo, for all the support, love, and consideration that I have had all my life. You guys are the reason why I want to become a better version of myself every day.

To my advisor Prof. Dr. Carlos Estombelo, for believing and guiding me through the course of my master's program while also being an incredibly understanding person. Thanks for your friendship and for being hard on me when I needed it.

To the best people in the world that have inspired me, stayed by my side in the hardest moments, and have helped this thesis come to life directly with their support and love, Fernando Melo, Gracieth Cavalcanti, Barbara Sena, and Rita Macedo.

To Prof. Dr. André Carvalho and all the friends that I made in São Paulo while working at USP, for their amazing friendship, research insights, and support while doing some ``balbúrdia'' inside and outside the lab.

To Prof. Dr. Vijay Mago, for having introduced me to the field of data science and made me believe that with the right amount of effort, I would be able to learn anything and produce valuable research. 

To Prof. Dr. Wilson Wang and Dr. Peter Luong, for having shaped me as a researcher and taught me how to overcome all the difficulties that academia could possibly have.

To all the great friends who understood my absence in certain moments and that I hope will always be with me Grace Kelly, Natalia Rosa, Thiago Charles, Raul Rodrigo, Duda Maia, Renan Albuquerque, Felipe Torres, Vinicius Araujo, Ronny Almeida, Davi Santana, Eriana Pinto, Manu Magno, and all the incredible others that I will probably have to pay a beer since their names are not here.

\end{agradecimentos}
\begin{epigrafe}[]
    \vspace*{\fill}
	\begin{flushright}
	
		\textit{\textit{``Quando disser sim para os outros, certifique-se \\ de não estar dizendo não para si mesmo.''}}
		\\
		\textit{(Paulo Coelho)}
		
	\end{flushright}
\end{epigrafe}
\setlength{\absparsep}{18pt} 
\begin{resumo}[Resumo]
 \begin{otherlanguage*}{portuguese}
   
Os cenários de vigilância e monitoramento estão propensos a vários problemas, pois não existe um controle sobre a distância dos possíveis suspeitos para a câmera e geralmente as tarefas envolvem avaliação de imagens em baixa resolução. Para tais situações, a aplicação de algoritmos de \textit{super-resolution} (super-resolução) pode ser uma alternativa adequada para recuperar as propriedades discriminantes das faces dos suspeitos envolvidos.

Embora abordagens gerais de super-resolução tenham sido propostas para aprimorar a qualidade da imagem para a percepção no nível humano, os métodos de super-resolução biométrica buscam a melhor versão da imagem para ``percepção'' do computador, pois seu foco é melhorar o desempenho do reconhecimento automático. Redes neurais convolucionais e algoritmos de aprendizado profundo, em geral, têm sido aplicados a tarefas de visão computacional e agora são o estado da arte em seus vários subdomínios, incluindo classificação, restauração e super-resolução de imagens. No entanto, poucos trabalhos avaliaram os efeitos que os mais recentes métodos de super-resolução propostos podem ter sobre a precisão e o desempenho da verificação de faces em imagens de baixa resolução do mundo real.

Este projeto teve como objetivo avaliar e adaptar diferentes arquiteturas de redes neurais profundas para a tarefa de super-resolução de faces, impulsionada pelo desempenho do reconhecimento de faces em imagens de baixa resolução do mundo real. Os resultados experimentais em um conjunto de dados de monitoramento/vigilância e de avaliação de presença universitária mostraram que arquiteturas gerais de super-resolução podem melhorar o desempenho da verificação de faces utilizando uma redes neural profunda treinada em faces de alta resolução para extração de características. Além disso, como as redes neurais são aproximadores de funções e podem ser treinadas com base em funções objetivo específicas, o uso de uma função de custo personalizada que foi otimizada para extração de características da face mostrou resultados promissores para recuperar atributos discriminantes em imagens de faces em baixa resolução.

   \textbf{Palavras-chave}: Reconhecimento Facial em Baixa Resolução; Super-Resolução; Aprendizado Profundo; Redes Neurais Convolucionais.
 \end{otherlanguage*}
\end{resumo}

\setlength{\absparsep}{18pt} 
\begin{resumo}
 
Surveillance scenarios are prone to several problems since they usually involve low-resolution footage, and there is no control of how far the subjects may be from the camera in the first place. This situation is suitable for the application of upsampling (super-resolution) algorithms since they may be able to recover the discriminant properties of the subjects involved.

While general super-resolution approaches were proposed to enhance image quality for human-level perception, biometrics super-resolution methods seek the best ``computer perception'' version of the image since their focus is on improving automatic recognition performance. Convolutional neural networks and deep learning algorithms, in general, have been applied to computer vision tasks and are now state-of-the-art for several sub-domains, including image classification, restoration, and super-resolution. However, no work has evaluated the effects that the latest proposed super-resolution methods may have upon the accuracy and face verification performance in low-resolution ``in-the-wild'' data. 

This project aimed at evaluating and adapting different deep neural network architectures for the task of face super-resolution driven by face recognition performance in real-world low-resolution images. The experimental results in a real-world surveillance and attendance datasets showed that general super-resolution architectures might enhance face verification performance of deep neural networks trained on high-resolution faces. Also, since neural networks are function approximators and can be trained based on specific objective functions, the use of a customized loss function optimized for feature extraction showed promising results for recovering discriminant features in low-resolution face images.

\end{resumo}

\pdfbookmark[0]{\listfigurename}{lof}
\listoffigures*
\cleardoublepage

\pdfbookmark[0]{\listtablename}{lot}
\listoftables*
\cleardoublepage

   

\begin{siglas}
    \item[CoordConv]{Coordinate Convolution}
    \item[DL]{Deep Learning}
    \item[FPS]{Frames per second}
    \item[FSRCNN]{Fast Super-Resolution Convolutional Neural Network}
    \item[GAN]{Generative Adversarial Network}
    \item[GPU]{Graphical Processor Unit}
    \item[HR]{High-Resolution}
    \item[LFW]{Labeled Faces in the Wild}
    \item[LR]{Low-Resolution}
    \item[MSE]{Mean Squared Error}
    \item[PSNR]{Peak signal-to-noise ratio}
    \item[SR]{Super-Resolution}
    \item[SRCNN]{Super-Resolution Convolutional Neural Network}
    \item[SubCNN]{Subpixel Convolutional Neural Network}
    \item[SRGAN]{Super-Resolution Generative Adversarial Network}
    \item[SOTA]{State-of-the-art}
    \item[SSIM]{Structural Simmilarity}
	\item[UFS]{Federal University of Sergipe}
	
\end{siglas}

\begin{simbolos}
  \item[$ \beta $] Greek letter Beta
  \item[$ \phi $] Greek letter Phi
  \item[$ \mathbf{R} $] Real space
  \item[$ \theta $] Theta
\end{simbolos}
\pdfbookmark[0]{\contentsname}{toc}
\tableofcontents*
\cleardoublepage

\textual
\chapter{Introduction}\label{sec:intro}

An essential ability in human beings that group them as social animals is face perception. Infants tend to prefer to look at faces at a very early age, and across the lifespan, most people spend more time looking at faces than at any other type of object \cite{johnson1991newborns}.

Faces provide a wealth of information that facilitates social communication since humans are able to recognize the identity of other people and interpret their emotional state by analyzing the facial expression and pose. More specifically, regarding identity recognition, there is behavioral and neural evidence that such a feature has its basis on the perception of aspects of facial structure that are invariant across changes \cite{gobbini2007neural, haxby2000distributed}.

Face perception is also related to a high-level visual and memory process that involves the retrieval of the memory of faces and the identity information stored in memory (i.e., person semantic knowledge). This process is developed in such a robust way in human brains that some people are able to recognize others by situations where there are only a few resembling features of a person, such as in caricature drawings and photos with low-resolution \cite{chang2017memory}. The field of research that describes and evaluates the reliable methods for automatic identification of subjects based on their physiological and behavioral characteristics is usually called biometrics~\cite{nguyen2018super}.

As an example of how face biometrics has become an important matter in modern society, situations in surveillance that employ the verification of a watch-list of subjects through CCTV footage have become quite regular for world security standards in airports, malls, and other crowded places. However, as sometimes they do not involve automation, they might become a weak spot as they require an impressive amount of manual work to check the live feed or saved data of several cameras~\cite{rasti2016convolutional}. This is one of the reasons why countries are spending a large number of resources to rapidly grow their technology market related to surveillance in order to have intelligible solutions specifically designed to their needs~\cite{feldstein2019global}.

Even though computers have shown a great ability to also deal with image and face recognition in the last decade, in situations where low-resolution (LR) inputs are employed, they tend to fail as much as humans when trying to identify an individual or reconstruct a higher-resolution representation of the same subject~\cite{nguyen2018super}. These occurrences are the majority in surveillance scenarios since the cheapest and most commonly used cameras can only provide low-quality video footage, and there is no control for the distance between the subjects of interest and the device \cite{rasti2016convolutional}. 

These recognition faults mainly occur because when the resolution drops, the amount of information available for identifying or verifying a subject decreases as well. That leads to a severe degradation for both human perception and machine interpretation. Since there is no standard resolution that can be set for making recognition available~\cite{nguyen2018super}, the development of image upscaling algorithms, commonly known as super-resolution (SR) algorithms, has become an intensive area of research. An example of that is the fact that the pioneering work of this group of algorithms dates back to 1974, when \citeonline{gerchberg1974super} showed that the resolution of a data object could be significantly improved through error energy reduction. Thenceforth, researchers have put a massive effort into investigating SR and its possible range of applications, even knowing that it is fundamentally an ill-posed problem since the details presented in the LR samples are usually not enough to provide a robust reconstruction of the original high-resolution (HR) image~\cite{tian2011survey}.

Deep Learning (DL) algorithms started to be used to solve tasks regarding image classification and reconstruction due to their computational cost being now facilitated by advances in hardware and parallel processing~\cite{krizhevsky2012imagenet}. This group of techniques has become the state-of-the-art (SOTA) rapidly in a great variety of tasks regarding images both for accuracy and applicability~\cite{lecun2015deep}. Also, they have shown excellent performance in image restoration tasks that are related to biometrics such as iris, fingerprint, and face super-resolution for improving recognition performance~\cite{ribeiro2017exploring,li2018deep,kim2019progressive}.

Most of the SR solutions for LR face recognition have relied on the use of convolutional neural networks (CNNs) optimized by a pixel loss~\cite{nguyen2018super}. Nevertheless, there exists nowadays a large pool of network designs and learning strategies that are applied to solve similar computer vision problems~\cite{haris2018task,liu2018intriguing}. Since the goal of SR for face biometrics is to optimize face recognition performance while keeping reasonable perceptual quality, replicating successful strategies from similar computer vision tasks can be a worth research direction. One example of a different strategy that some similar works have applied is the use of different types of convolution operators and customized loss functions to increase performance~\cite{wang2019generative,wang2019deep}.

One of the current issues with SR solutions to the LR face recognition problem is that, researchers often train their SR deep learning models reporting their accuracy results only on the downsampled version of the same or other HR frontal image dataset~\cite{ouyang2018deep, abello2019optimizingSR}. However, it is known that such task becomes more challenging when faces are captured in an unconstrained environment where they can be subject to blurring, motion, non-frontal pose, and other situations that hinder recognition.

The origin of the analysis to be presented in this thesis is related to the lack of recent studies of if and how the state-of-the-art deep learning SR techniques may assist face biometrics in real-world low-resolution scenarios, taking into consideration also different network architectures, learning strategies, and their real applicability and scalability.

\section{Hypotheses}

For the development of this thesis and the proposal of experiments, the following specific hypotheses were elaborated:

\begin{enumerate}
    \item The relationship between image quality metrics and accuracy performance is not significant.
    \item The use of a specific convolution operator that take into account position information (CoordConv) can effectively improve metric performance over normal convolution operators when dealing with super-resolution.
    \item Application of a loss function based on face identity for an upscaling network (FaceLoss) can influence the verification results positively in a face recognition pipeline using DL models.
\end{enumerate}

\section{Objectives}

Taking into account all the possible challenges regarding the discussed topics, the general objective of this thesis is to evaluate the efficiency of a face recognition pipeline in real-world low-resolution scenarios and check whether the recently developed SR algorithms and their variants are capable of enhancing recognition performance in these situations.

The specific objectives are listed below:

\begin{itemize}
    \item Evaluation of the possibility of a correlation between image quality metrics and face verification accuracy in a LR recognition pipeline as considered by hypothesis 1.
    \item Evaluation of different SOTA neural network architectures, also involving different convolution operators as proposed in hypothesis 2, for the super-resolution task driven by face biometrics performance involving faces in real-world LR datasets.
    \item Evaluation of an adapted loss function that optimizes the DL model for better face feature extraction while keeping the SR upsampling characteristic as suggested by hypothesis 3.
\end{itemize}

\section{Thesis Structure}

In order to make an easier read, this thesis brings the technical background before the related work chapter since the discussed topics are from recent research, and a prior overview can be useful for a better comprehension of the concepts. Therefore, this manuscript was structured with the following chapters:

\begin{itemize}
\item {Chapter 1 - Introduction}
\item {Chapter 2 - Technical Background}
\item {Chapter 3 - Related Work}
\item {Chapter 4 - Methodology}
\item {Chapter 5 - Experiments}
\item {Chapter 6 - Results}
\item {Chapter 7 - Final Considerations}
\end{itemize}

\chapter{Technical Background}

This chapter gives a brief technical background overview for the topics discussed in this thesis in order to provide the basics that validate the proposed experiments and hypotheses.

\section{Convolutional Neural Networks and Deep Learning}

Deep learning (DL) is a branch of machine learning that is capable of learning the data representation through the use of a structure of hierarchical layers, similar to the way the brain handles new information. Its concept is mainly applied to supervised learning problems (e.g., where there is a need for mapping an input vector to an output vector), and its core is based on the math behind Artificial Neural Networks~\cite{lecun2015deep}.

Deep Neural Networks can have different architectures based on the nature of the data that is used as input. When image data needs to be processed as input, CNNs have been ideally applied by academia and industry because of its interior architecture properly set to work with high dimensional data and extract its more discriminating features. \cite{lecun2015deep, shi2016real}

\begin{figure}[!htb]
	\begin{center}
		\includegraphics[width=1\textwidth]{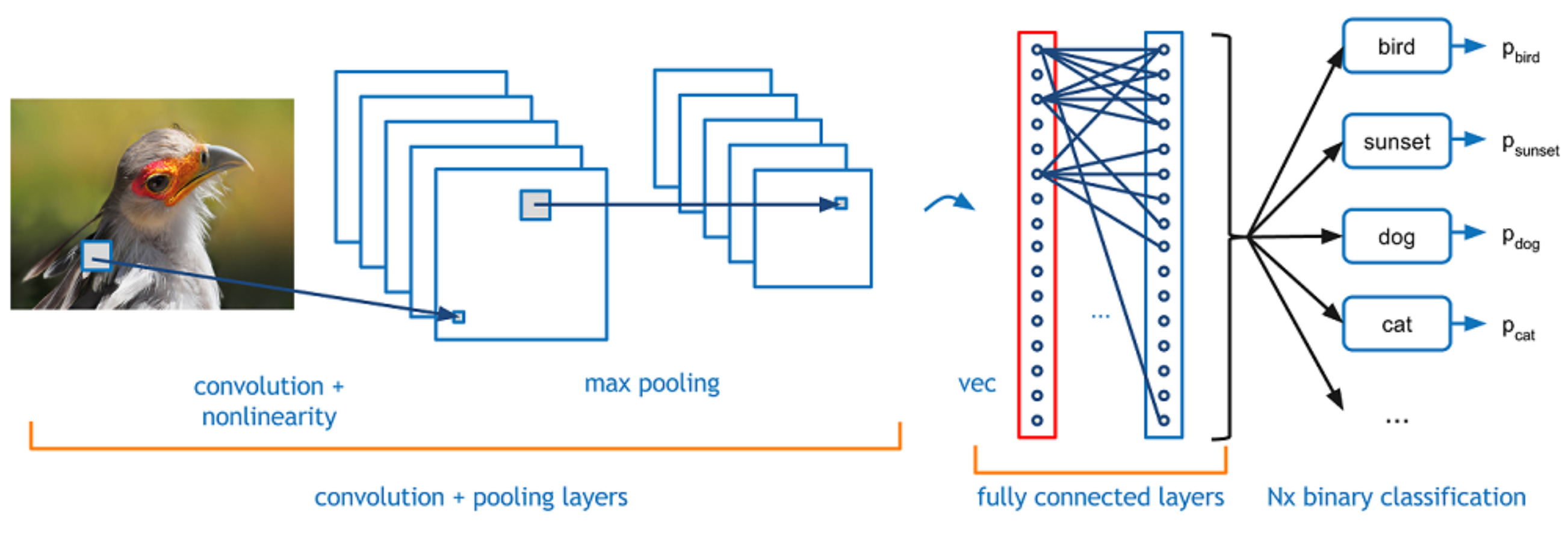}
	\end{center}
    \caption{\label{fig:conv_struc} CNN Architecture Exemplified \cite{blogcnn2019}}
\end{figure}

A typical structure of a CNN can be seen by Figure \ref{fig:conv_struc} where an image is used as input, and the network needs to predict a label for it. The first operation that happens inside the network is on the convolutional layer, where a moving window is applied to a small pixel grid of the image. This moving window, commonly called a kernel, works as a ``filter'' and its task is to multiply its weight values by the original pixel values. All these multiplications are summed up to one number that is going to be placed on the matrix used as input on the following layer.

The CNN per see consists of several stacked convolutional networks mixed with nonlinear and pooling layers that work as feature extractors. Usually, the nonlinear layer is added after each convolution operation, which brings a nonlinear property characteristic to the network through the use of an activation function. The pooling layer will then be placed after the nonlinear layer working directly with the width and height of the image in order to perform a downsampling operation. This step reduces the image data to a more compressed version containing only details that were processed and identified by the previous filter (convolutional) layer. After a series of ``feature extraction'' layers, a fully connected layer is generally stacked upon them in order to map the extracted features to a fixed output.

The learning phase of a CNN happens on the update of the weights presented on every convolutional layer and the weights for the fully connected one. The first often allows the network to identify edges, contours, and shapes that characterize the image while the second is accountable for the classification or regression step. The training is usually performed using variants of gradient-based optimization methods via backpropagation \cite{krizhevsky2012imagenet, lecun2015deep}.

\subsection{Residual Networks}

\begin{figure}[!ht]
	\begin{center}
		\includegraphics[width=.6\textwidth]{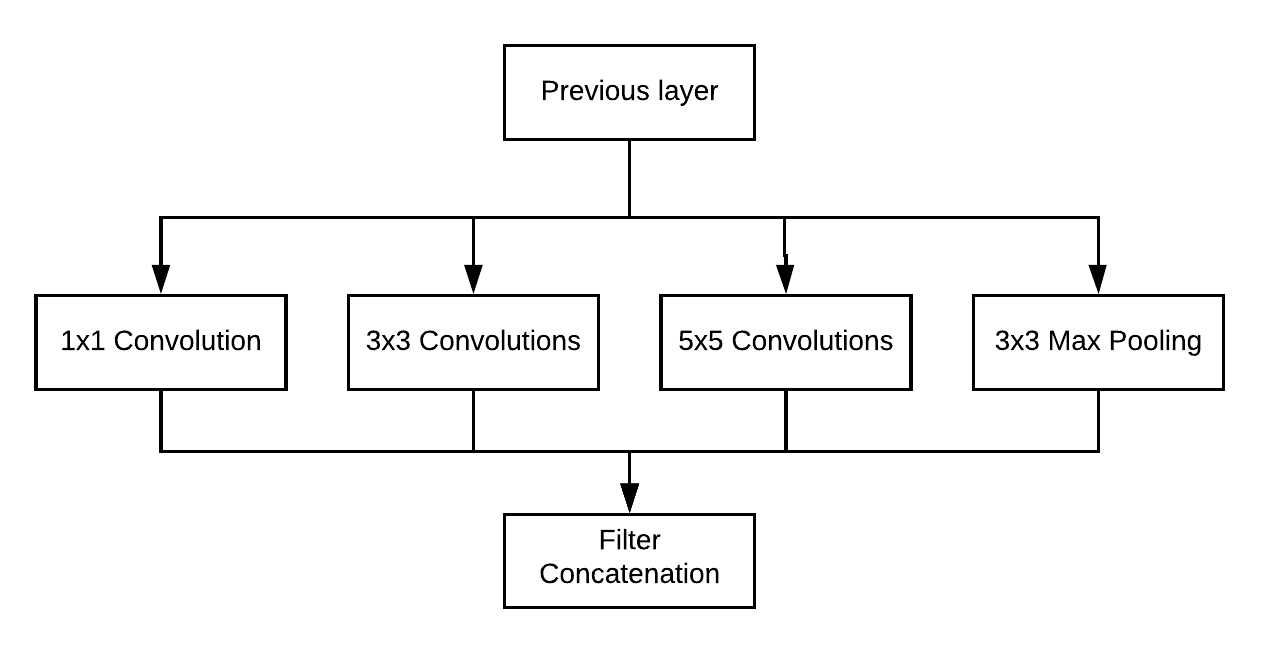}
	\end{center}
    \caption{\label{fig:inception-block} Example of Inception Block. (Source: Author's own)}
\end{figure}

When training large image classifiers, usually there is a considerable variation in the location and size of the object of interest. In order to have a robust feature extractor that identifies features that are globally or locally distributed on the image, the use of different kernel sizes may be needed. With this in mind, \citeonline{szegedy2015going} proposed GoogleNet using large blocks that contained different convolution operators with several kernel sizes. One representation of such block is shown in Figure~\ref{fig:inception-block}.

\begin{figure}[!ht]
	\begin{center}
		\includegraphics[width=1\textwidth]{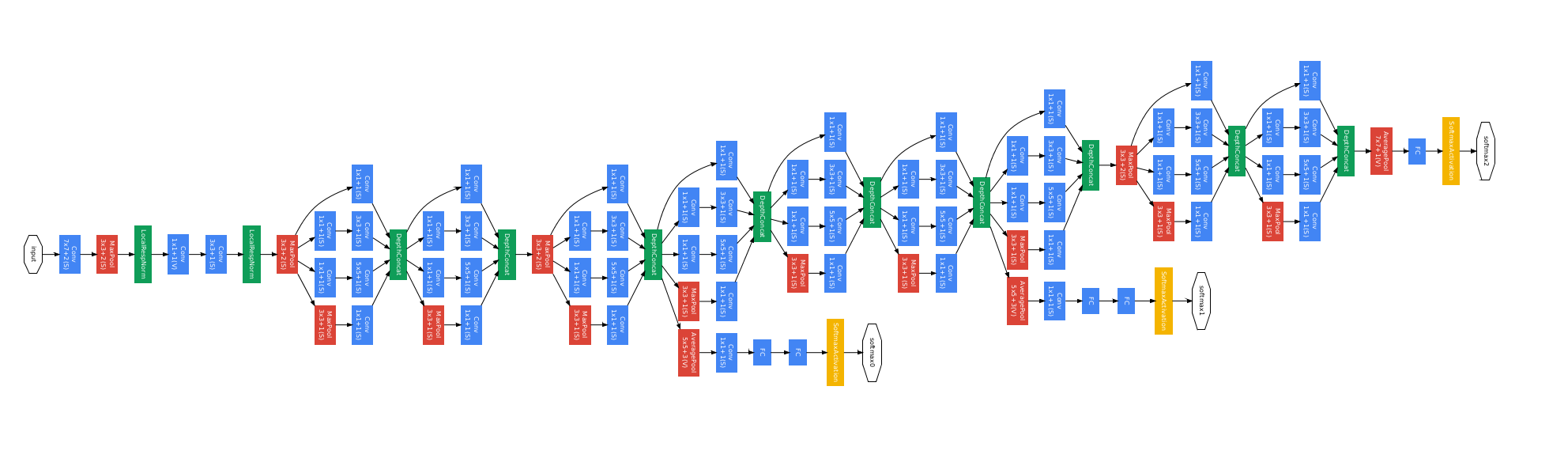}
	\end{center}
    \caption{\label{fig:google-net} GoogleNet architecture. \cite{szegedy2015going}}
\end{figure}

Using several stacks of blocks in a very deep network, they were able to achieve 93.3\% top-5 accuracy on the ImageNet competition with much less computation than the state-of-the-art (SOTA) at that time, VGG16. The final architecture of GoogleNet can be seen in Figure~\ref{fig:google-net}.

Nevertheless, as academia started to implement and test different types of deep architectures, the problem of vanishing gradients became popular. This issue appears because certain activation functions squish an ample input space into the range between 0 and 1. Then, sometimes even when a large change arrives in the input, the output is going to have only a minor change, and consequently, the gradients become too small for updating the weights when backpropagated~\cite{lecun2015deep}.

\begin{figure}[!ht]
	\begin{center}
		\includegraphics[width=.4\textwidth]{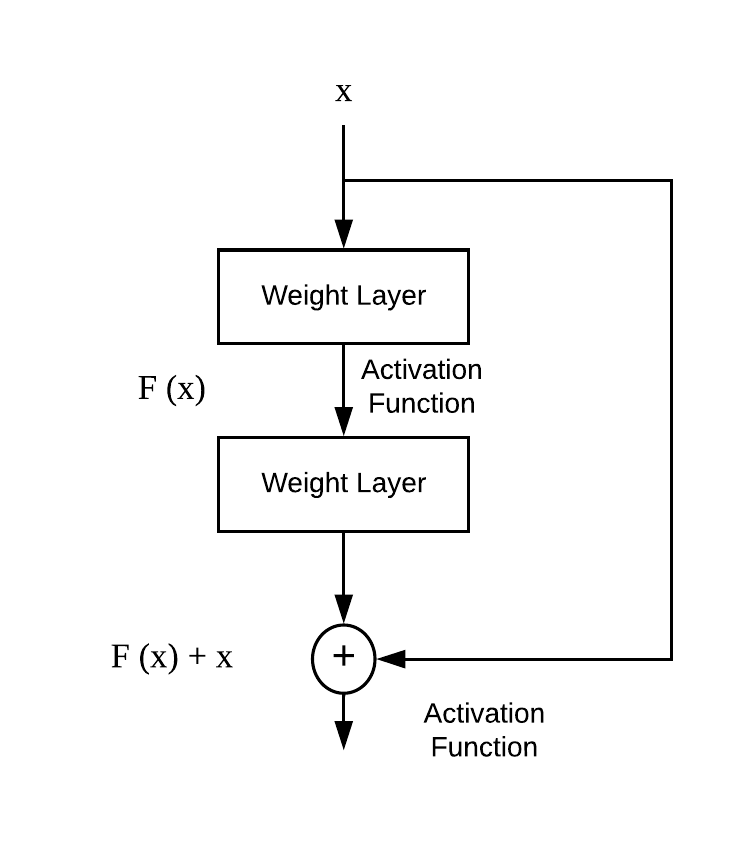}
	\end{center}
    \caption{\label{fig:res-block} Single Residual Block. (Source: Author's own)}
\end{figure}

One solution that researchers found to this problem was to use skip connections. These connections, as shown by Figure~\ref{fig:res-block}, are used to feed posterior layers the same input that previous layers had, which makes the network skip the training of a few layers and learn only the residual between the input and the output~\cite{he2016deep}.

This structure gave the name for the group of residual networks, commonly known as ResNets, and influenced researchers to go even deeper since networks consequently could have more layers and still train in sufficient time. One of the examples of such structure in SOTA applications is in the work of \citeonline{szegedy2017inception}, where inception and residual blocks are combined to create robust feature extractors.

\newpage

\subsection{Generative Adversarial Networks}

Generative Adversarial Networks (GANs) were proposed by \citeonline{goodfellow2014generative} in order to sidestep the common difficulties that involve deep generative models such as approximating intractable probabilistic computations that arise in maximum likelihood estimation and leveraging the benefits of piecewise linear units in the generative context. 

\begin{figure}[!ht]
	\begin{center}
		\includegraphics[width=1\textwidth]{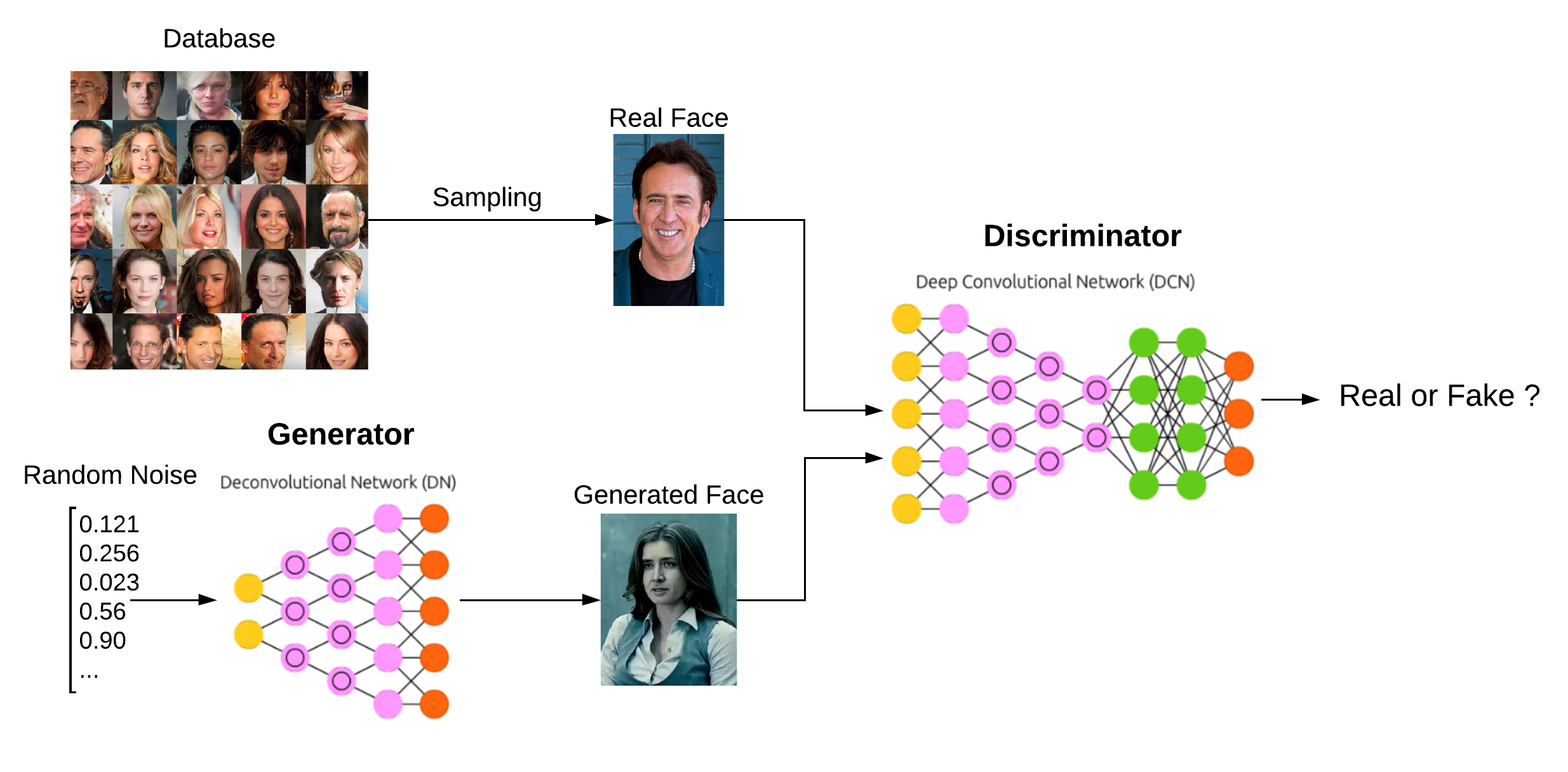}
	\end{center}
    \caption{\label{fig:arch-gan} Architecture of a GAN. (Source: Author's own)}
\end{figure}

In this architecture, a discriminator network $D(x)$, where $x$ is an image, is optimized for distinguishing whether the given input is fake or not, while a generator network $G(x)$, where $x$ can be random noise or even another image, is optimized to generate fake image samples that follow the same distribution of the real image and fool the discriminator from discerning which one is the real~\cite{wang2019generative}. Therefore, in this context, the output for the discriminator network is always a label (real $\rightarrow$ 1, fake $\rightarrow$ 0), and for the generator is always an image. The general idea presented in the learning process is shown in Figure~\ref{fig:arch-gan}.

In other words, $D$ is trained to maximize the probability of assigning the same correct label for both generated and real images, while simultaneously $G$ is trained to minimize $log(1-D(G(z)))$. \citeonline{goodfellow2014generative} described their optimization, also known as adversarial training, as the play of a minimax game with value function $V(D,G)$:

\begin{equation}
    min(G) \, max(D)\, \rightarrow V(D,G) = E_{x \, \sim p_{data}(x)}[logD(x))] + E_{z \, \sim p_{z}(z)}[log(1 - D(G(x)))]
\end{equation}

\begin{flushleft} given $p_{z}(z)$ as the input noise and considering $E_{x}$ and $E_{z}$ the error associated with discriminator and generator, respectively.\end{flushleft}

Since then, GANs attracted growing interests in the research community due to their applicability and versatility. They have been applied to various domains such as natural language processing, time-series synthesis, and computer vision~\cite{yang2017semi,donahue2018synthesizing,bao2017cvae}. In the latter area, they have become the SOTA for several applications such as image-to-image translation, image inpainting, and image SR~\cite{ma2018gan,yu2018generative,ledig2017photo}.

However, since generator and discriminator need to achieve Nash equilibrium during training where neither generator nor discriminator can become too specialist in its task, GANs suffer from major challenges when training such as non-convergence, mode collapse, and diminished gradient~\cite{wang2019generative}. Consequently, they are highly sensitive to hyperparameters. In addition, for obtaining good results with them, their loss functions need to represent well the real optimization problem involved in the task~\cite{johnson2016perceptual}.

\subsection{Coordinate Convolutions}

The convolution operator is widely used in image processing, after learning the ideal filter weights, due to its ability to extract features of content from the training set that may not be in the same angle or place all the time. Such learned characteristic is called translation invariance. However, \citeonline{liu2018intriguing} noted that also due to this feature, regular convolutions in CNNs could perform poorly in tasks that involve coordinate transforms. One example of this problem is the mapping between coordinates in $(x,y)$ cartesian space to coordinates in the pixel space features, where even state-of-the-art architectures would bot be able to obtain more than 90\% of testing accuracy. 

\begin{figure}[!ht]
	\begin{center}
		\includegraphics[width=1\textwidth]{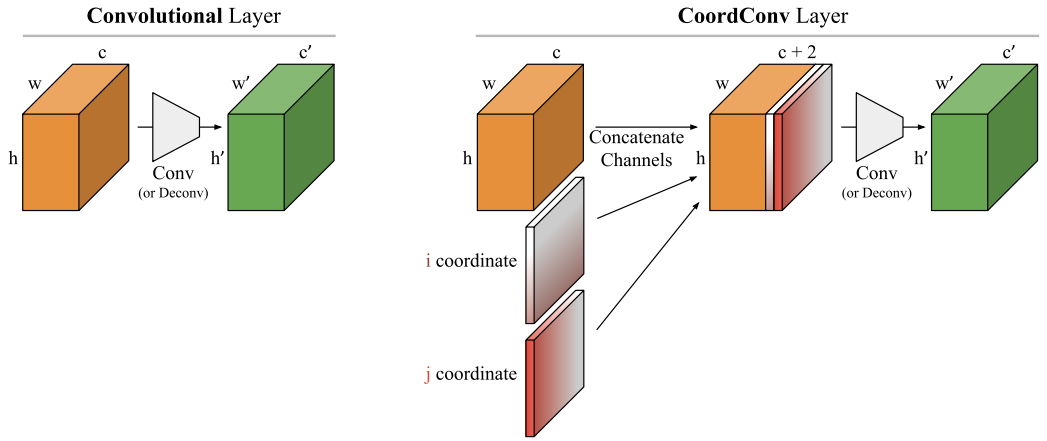}
	\end{center}
    \caption{\label{fig:coord-conv-ex} Differences between normal and coordinate convolutions. \cite{liu2018intriguing}}
\end{figure}

For dealing with problems that require varying degrees of translation dependence or complete translation invariance, \citeonline{liu2018intriguing} proposed an operator called CoordConv, which works by giving the normal convolution operator access to its own input coordinates through the use of extra coordinate channels. This allows the network to check and work with the exact location of pixels inside its grid. This operator allows the network to learn either complete translation invariance or varying degrees of translation dependence, as required by position regression tasks. Their result in the same given position regression task presented perfect generalization, being 150 times faster, and having 10–100 times fewer parameters. The difference between a standard convolution operator to a CoordConv can be visualized in Figure~\ref{fig:coord-conv-ex}.

Since their launch, researchers have explored different applications and scenarios where normal convolutions can be switched to CoordConv for improving performance~\cite{upadhyay2019spinal,xu2019view}. Nonetheless, only \citeonline{zafeirouli2019efficient} so far in literature have reported the improvements that CoordConvs may provide over the use of regular convolutions for SR, which makes it an interesting research direction.

\section{Super-Resolution}

\begin{figure}[!ht]
	\begin{center}
		\includegraphics[width=.5\textwidth]{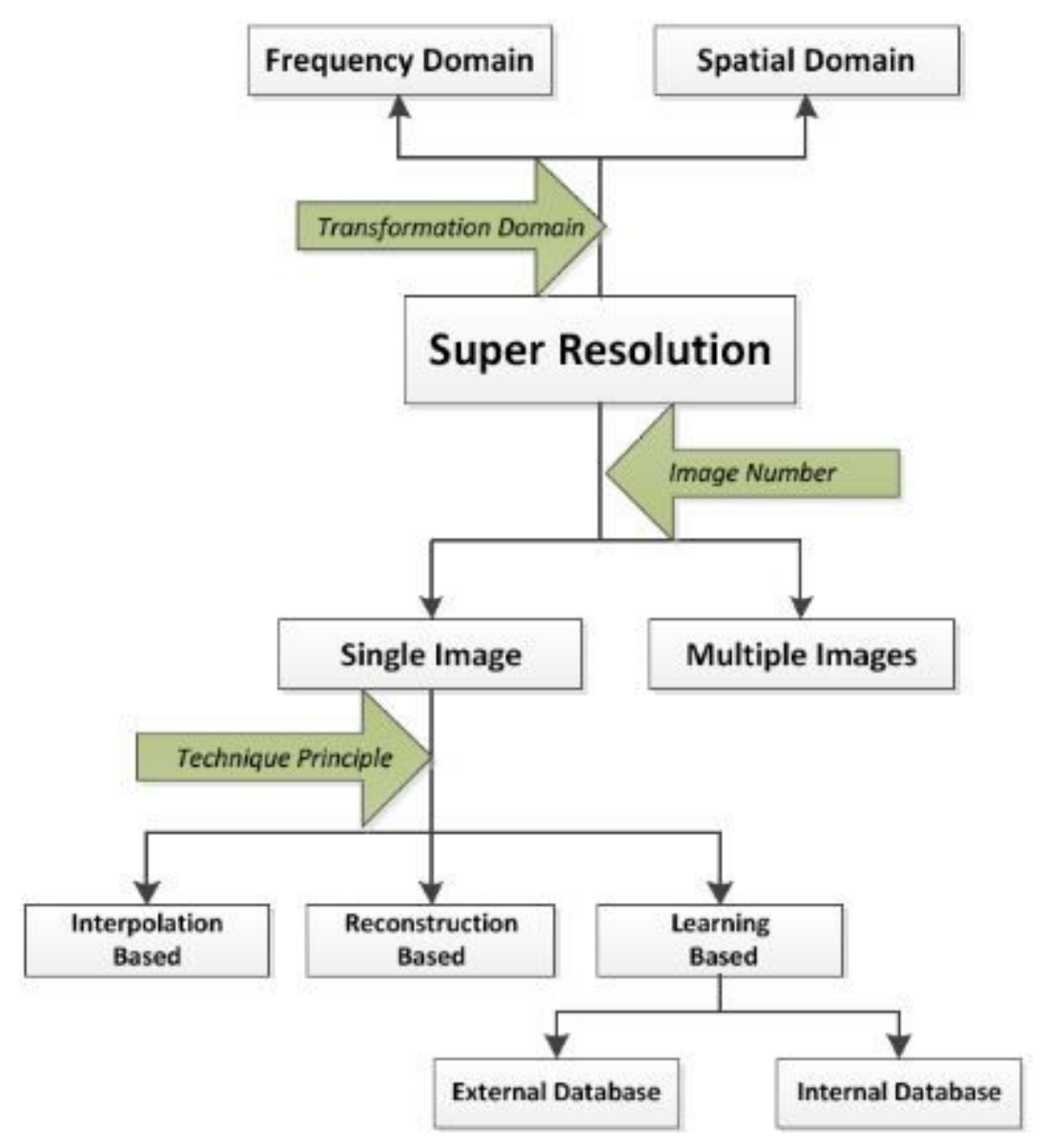}
	\end{center}
    \caption{\label{fig:classes_sr} General classes of SR algorithms. \cite{huang2015short}} 
\end{figure}

Super-resolution can be described as an attempt to generating a higher resolution image out of a lower resolution input. Throughout this domain, researchers have applied different strategies to reconstruct the HR image, which culminated in different classes of SR algorithms being developed depending on a variety of conditions~\cite{huang2015short}. Some of the categories involving SR are shown in Figure \ref{fig:classes_sr}.

The general principle of supervised SR is that a LR image $I_{LR}$ is the result of a degradation process that was applied to its HR version $I_{HR}$ as in:
\begin{equation}
    I_{LR} = D(I_{HR})
\end{equation}

The degradation function $D$ is naturally unknown, but researchers usually associate it with blur, motion, warp, and noise \cite{nguyen2018super}. Therefore, the goal of the SR algorithm is to learn the inverse mapping in such a way that, from a LR input, its HR can be achieved as in:

\begin{equation}
    I_{HR} = F_{SR}(I_{LR};\theta)
\end{equation}

\begin{flushleft}
where $F_{SR}$ is the SR function and $\theta$ its parameters.
\end{flushleft}

\begin{figure}[!ht]
\centering
\begin{subfigure}{.25\textwidth}
  \centering
  \includegraphics[width=1\linewidth]{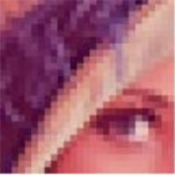}
  \caption{}
  \label{fig:sub1}
\end{subfigure}%
\begin{subfigure}{.25\textwidth}
  \centering
  \includegraphics[width=1\linewidth]{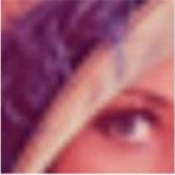}
  \caption{}
  \label{fig:sub2}
\end{subfigure}%
\begin{subfigure}{.25\textwidth}
  \centering
  \includegraphics[width=1\linewidth]{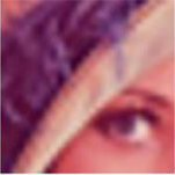}
  \caption{}
  \label{fig:sub3}
\end{subfigure}%
\begin{subfigure}{.25\textwidth}
  \centering
  \includegraphics[width=1\linewidth]{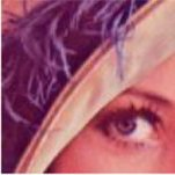}
  \caption{}
  \label{fig:sub4}
\end{subfigure}%
\caption{Visual comparison of general interpolation methods: (\subref{fig:sub1}) Nearest Neighbor (\subref{fig:sub2}) Bilinear (\subref{fig:sub3}) Bicubic
(\subref{fig:sub4}) Original HD image. - (Source: Author's own)}
\label{fig:interpolations}
\end{figure}

The most common and used techniques for upscaling images are the ones based on interpolation such as bicubic, bilinear, or nearest neighbor since their time cost is low, which makes them ideal for real-time applications. An illustration of the results when zooming an image (4x) with each technique is presented in Figure \ref{fig:interpolations}. Although the bicubic interpolation has a higher time complexity, it is the default method for upscaling images in software such as MATLAB and Photoshop. \cite{purkait2014fuzzy, vedadi2014map}

\subsection{Operating Channels}

The human evaluation of the degradation degree in a LR image is based on the perception of the RGB channel. However, when applying SR methods to images, some researchers instead use the YCbCr color space representation. In this space, images are depicted in Y, Cb, Cr channels, denoting the luminance, blue-difference, and red-difference chroma components, respectively~\cite{wang2019deep}. Some works report that using only the Y channel may bring better results than when working with the addition of Cb and Cr channels since they are more blurry than the Y channel by nature, and therefore are less affected by the downsampling process~\cite{dong2015image}. There is no consensus in academia for which channels are better for training and evaluating SR; nevertheless, the most recent architectures tend to operate on RGB channels~\cite{ledig2017photo,Chen2018FSRNetEL}.

\subsection{Super-Resolution Benchmarking}

Even though different works have presented several ways of benchmarking and measuring their image SR results regarding their specific field of application, the most common objective measurements of image quality are Peak Signal to Noise Ratio (PSNR) and Structural Similarity (SSIM). \cite{tian2010task}. PSNR is an estimation of quality based on the mean squared error (MSE) of pixels for every channel between the HR image generated and the ground truth, as can be seen in Equations \ref{eq:psnr} e \ref{eq:mse}. 

\begin{equation}
    PSNR = 10\log_{10}(\frac{S^{2}}{MSE})
    \label{eq:psnr}
\end{equation}

\begin{equation}
    MSE = \frac{\sum_{n,m} (x_{mn} - y_{mn})^2}{m*n}
    \label{eq:mse}
\end{equation}

\begin{flushleft} where: $S$ is the maximum value in the input image data type; $n$ is the number of pixels; $m$ the number of channels; $x_{mn}$ and $y_{mn}$ represent the pixel value described in $n$ with the channel $m$ for the generated and original images respectively.
\end{flushleft}

SSIM is a measurement that considers the visual degradation in quality with more importance through analysis of the homogeneity and phase coherence of the gradient magnitude on the original and reconstructed image. This similarity is based on structure, brightness, and contrast of the images. \cite{begin2006comparison, reibman2006quality} Its mathematical formulation can be seen in Equation \ref{eq:ssim}.

\begin{equation}
    SSIM = \frac{(2 \mu_{x} \mu_{y} + c_{1})(2\sigma_{xy} + c_{2})}{(\mu_{x}^2 + \mu_{y}^2 + c_{1})(\sigma_{x}^2 + \sigma_{y}^2 + c_{2})}    \label{eq:ssim}
\end{equation}

\begin{flushleft}
where: $\mu_{x}$ and $\mu_{y}$ represent the average intensity value of a linked windows for the original and reconstructed image; $c_{1}$ and $c_{2}$ denote the brightness of two images; $\sigma_{x}$ and $\sigma_{y}$ formulate the variance of the two sets of intensity for both images; $\sigma_{xy}$ presents the correlation between these two sets.
\end{flushleft}

\subsection{Deep Learning for Image Super-Resolution}

Deep learning solutions for SR fits into the ``learning-based'' category show in Figure~\ref{fig:classes_sr}. In the last few years, DL methods have become the most explored approach for performing SR tasks since they early showed SOTA performance in various benchmarks and competitions~\cite{Agustsson_2017_CVPR_Workshops,Timofte_2018_CVPR_Workshops}. In special, the single image super-resolution (SISR) problem has been the most fundamentally tackled problem within SR, since researchers can make use of already available large datasets scrapped from the internet to train their models~\cite{liu2015faceattributes, Chen2018FSRNetEL}.

\begin{figure}[!ht]
	\begin{center}
		\includegraphics[width=\textwidth]{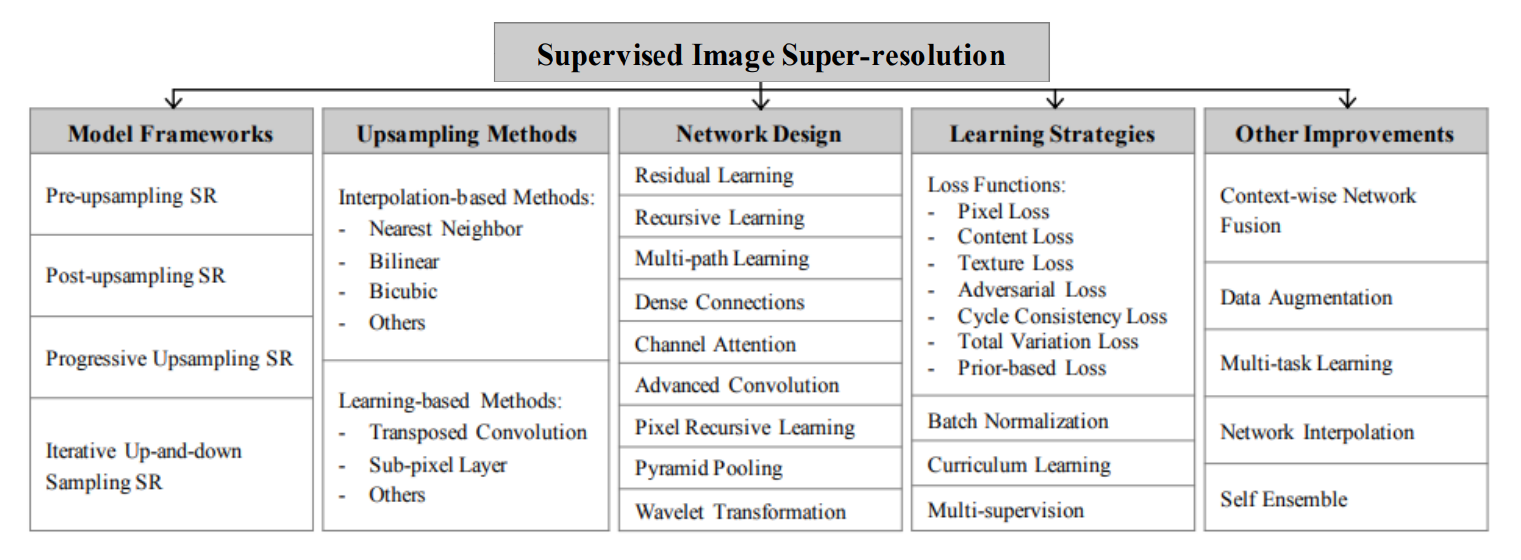}
	\end{center}
    \caption{\label{fig:dl-sr} DL for SR algorithms related topics. (Adapted from \citeonline{wang2019deep})} 
\end{figure}

A variety of methods have been used and incorporated for solving the SR problem, ranging from simpler approaches involving only convolutional layers, to more sophisticated ones with the use of residual blocks, recursive learning and different losses~\cite{wang2019deep}. An overview of the most related directions that researchers have taken when considering working with DL in SR can be analyzed in Figure~\ref{fig:dl-sr}.

For proposing the new architectures assessed in this thesis, different network designs and learning strategies presented in Figure~\ref{fig:dl-sr} were considered, such as the use of ``Residual Learning'' and ``Content Loss''. A more deep review of the works which influenced the directions taken in this manuscript is presented in Chapter~\ref{related-work}.

\section{Face Recognition}

\begin{figure}[!ht]
	\begin{center}
		\includegraphics[width=\textwidth]{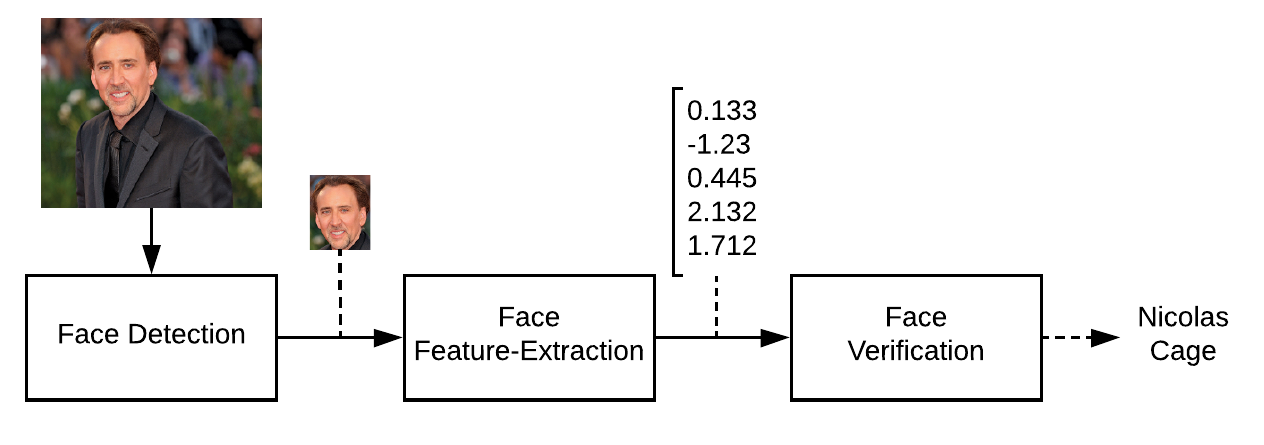}
	\end{center}
    \caption{\label{fig:nic-face} Example of a generic pipeline for face recognition. (Source: Author's own)} 
\end{figure}

The basic steps that involve a general face recognition pipeline are defined in Figure~\ref{fig:nic-face} and described in sequence.

\subsection{Face Detection}

Face detection in an image is the first step in a recognition pipeline because it eliminates unnecessary information from the image. In this way, if the algorithm finds one or more faces, they are extracted from the original image so that they can be analyzed separately~\cite{muttu2015effective}.

The training phase of these algorithms happens with the use of several images containing faces and others without them. Even though this problem presents itself as a simple binary classification, several face detection algorithms need to be trained exhaustively so that they can give good results~\cite{zhang2016joint}. Two measures are responsible for evaluating the quality of face detection algorithms ~\cite{detection2003}: 

\begin{itemize}
    \item False positive: Represents the number of objects that were detected wrongly as faces.
    \item False negative: Represents the number of faces that were not detected.
\end{itemize}

Face detection algorithms are usually divided into four different groups: knowledge, feature, template, and appearance-based models ~\cite{zafeiriou2015survey}. However, as the amount of available data has increased over the years for training such algorithms, the appearance-based methods have overcome the other solutions since they generalize face models from a set of representative samples. A common core for the SOTA algorithms proposed in this group of techniques is the use of CNNs since they derive problem-specific feature extractors from the training examples automatically, without making any assumptions about the features to extract or the areas of the face patterns to analyze due to their spatially invariant characteristic~\cite{zhang2010survey}.

\subsection{Feature Extraction and Face Verification}

The ``real'' recognition step in a face recognition pipeline consists of the representation and extraction of facial features of an image. These features are then input into a mathematical model, which is meant to specify whether the presented face matches one or any previously stored face~\cite{crosswhite2018template}.

The implementation of recognition systems can range from low-throughput to process-intensive methods where, for example, GPUs are required. Some more straightforward methods can make use of metric learning approaches or principal component analysis for dimensionality reduction. On the other hand, the most sophisticated ones are usually based on analysis of probability densities, manifold learning, and deep neural networks, among other methods with a higher computational cost~\cite{wang2018deep}.

For extracting discriminative features of an image that only contains a face (after the pre-processing step), models based on CNN have been the ones most used by SOTA approaches. This architecture is suitable for feature extraction because it takes advantage of local connections to extract the spatial information effectively. Also, their shared weights significantly reduce the number of parameters for training the network, which consequently reduces its size~\cite{chen2016deep}. An effective way to create accurate face recognition models is through the application of Transfer Learning~\cite{lecun2015deep} using available pre-trained models. These models are often trained in datasets with millions of faces and, through the use of their intern representations, it is possible to extract discriminative features of an input face directly~\cite{cao2018vggface2}.

Once extracted all the features of the involved subjects, the system needs to decide whether the person is whom he/she claims to be. This step is called face verification and different machine learning approaches can be employed to perform it depending on how many dimensions the obtained feature space may have~\cite{faceli2011inteligencia}. These approaches can be differentiated by how their functions create the decision boundaries on the feature hyperplane. However, in the context of face recognition, when only one or few training samples are provided, methods based on distance metrics have shown the best results regarding computational complexity and accuracy~\cite{nguyen2010cosine,schroff2015facenet}. 

\section{Final Considerations}

In this chapter, an overview of the main topics discussed in this thesis was provided. It is important to reinforce that most of the trends regarding DL in the fields of face recognition and super-resolution have only emerged in the past five years through empirical experimentation with different architectures. This statement indicates that most of the theory behind why these models have performed better than others is still in the development phase and will probably lead in more exploration and changes in the following years.

In the next chapter, different SOTA works with respect to SR and LR face recognition are discussed. Their evaluation was essential to extract meaningful insights for proposing the hypotheses and objectives of this thesis.

\chapter{Related Work} \label{related-work}

This chapter presents some of the related works evaluated during the development of this thesis. 

\section{Super-Resolution}

\citeonline{baker2000hallucinating} proposed the first SR work to be applied to faces in 2000. They created an algorithm that was used to learn priors on the spatial distribution of the image gradient for frontal images of faces. At that time, they stated that the high-frequency details inferred by the probabilistic models were ``hallucinated'' by the model.

The work of \citeonline{tian2010task} presented objective and subjective measures for evaluating how SR impacts different image processing and computer vision tasks. Their findings reflected the conflicts between objective and subjective measures since the former tends to penalize the model that enhanced the image according to computer vision standards, and the latter tends more to changes that improve the image quality based on the human vision system.

\citeonline{dong2015image} were the first to propose the use of CNNs for the SR problem. Their architecture was called ``Super-Resolution Convolutional Neural Network'' (SRCNN) and provided superior accuracy compared with other SOTA example-based methods at the time. In their work, LR images are pre-upsampled using traditional methods (e.g., bicubic interpolation) to the desired size, and then a deep CNN with three layers is applied to the coarse image for reconstructing the high-frequency details. This work became later the baseline for all works that involve DL based algorithms in SR. One advantage of this method (and all pre-upsampling methods) is that they can take input images of any arbitrary size and perform the SR task. The downsides may be the introduction of noise and blurring and, since most operations are performed with images in a high-dimensional space, time and memory costs can be higher than other frameworks.

\citeonline{dong2016accelerating} designed a compact hourglass-shape CNN structure using the basic SRCNN structure for faster inference and improved accuracy in SR called FSRCNN. They proposed an architecture with a deconvolution layer at the end of the network for mapping the original LR image directly to the super-resolved output, an iterative up-and-down sampling in the mapping layers, and the use of smaller filter sizes with more mapping layers. The results pointed out an increase in performance of over 40x for inference time while presenting a superior restoration quality when compared against the naive SRCNN architecture. 

For the work of \citeonline{shi2016real}, the authors presented a strategy to solve the necessity to upscale the LR with interpolation methods or using a single filter before feature extracting and mapping. They presented a modified CNN architecture with en efficient sub-pixel convolutional layer for ``post-upsampling'' where the feature extraction could happen in the LR space before being upscaled. This architecture was capable of performing real-time SR in 1080p videos on a single K2 GPU.

In the work of \citeonline{ledig2017photo}, a deep generative adversarial network using residual convolutional blocks was applied for image SR. Their approach achieved SOTA results in upscaling photo-realistic natural images by a factor of 4. To accomplish such results, instead of only optimizing the network by image similarity in pixel space, the authors proposed a perceptual loss function, which consisted of an adversarial loss and a content loss. The adversarial loss was responsible for pushing the upscaled solution to the natural image manifold using a discriminator network trained to differentiate between the super-resolved images and the original photo-realistic ones. Besides, they proposed the use of a content loss motivated by perceptual similarity. This similarity was calculated from the comparison of extracted semantic features from an ImageNet pre-trained network. Also, they evaluated the impact of applying several image losses together, such as adversarial, content, mean-squared-error, and total-variance, which inspired this thesis in investigating a different task-specific learning strategy.

The work of \citeonline{haris2018task} presented an approach to detect objects in LR images using an end-to-end strategy with the training of a CNN to perform the SR steps, and also aid detection. In this approach, a specific multi-objective loss function was developed for CNN training, where individual weights for each part of the loss were used in order to optimize the learning process based on each desired task. The goal behind the work was to assess how much improvement in resolution would assist a recognition/detection task in the input image.

\citeonline{Chen2018FSRNetEL} presented an end-to-end approach to perform SR on face images using prior geometric face features as prior information. The authors divided the training process into several stages where different encoder-decoder network architectures were applied to extract geometric features from the faces to aid the task of SR. The deep network produced in this work, FSRNet, presented results that today are the SOTA for SR in face images. However, when dealing with real-world SR of LR face images in the wild, obtaining priors is a hard and computationally expensive task that hinders its implementation in the face biometric context.

\citeonline{zafeirouli2019efficient} proposed an efficient, lightweight model leveraged by the benefits of a recursive progressive upsampling architecture to tackle the SR problem. This work recognized that SR tasks involve spatial representations and transformations, and exploited the pixel position information to reinforce the reconstruction task using the CoordConv operator. They obtained comparative results with SOTA implementations in four SR benchmarks. More importantly, their results also showed accuracy performance improvements with the use of the coordinate convolutional layer for the SR task while keeping low computational complexity, which motivated the application of this operator for proposing the new architectures described in Chapter~\ref{sec:results}.

\section{Low-Resolution Face Recognition}

\citeonline{hennings2008simultaneous} presented an approach for simultaneous SR and face feature extraction for recognition of LR faces by treating face features (e.g., Eigenfaces, Fisherfaces) as prior information in the SR method. They evaluated their approach against matching gallery and probe images in the LR and applying the pure SR approach to check for matches in the high-dimensional domain. They concluded that their approach could produce better recognition performance since the focus of the SR shifted to recognition instead of reconstruction. This particular work inspired this thesis for using features extracted from the face as an optimization strategy for the SR models.

The work of \citeonline{rasti2016convolutional} proposed a system that super-resolves a face image before the face feature extraction and recognition phases. They used a deep CNN to upscale the image followed by a Hidden Markov Model and Single Value Decomposition based face recognition model. They experimented in two general and one small surveillance database and pointed out that such upscaling phase could result in a 6 to 10\% increase in performance for face recognition. The increase in accuracy performance reported in this paper influenced the elaboration of the hypothesis that DL based SR could assist positively face recognition in real-world LR data.  

\citeonline{berger2016boosting} proposed a two-step neural approach for face SR with the focus of improving face recognition. They employed a generic SR CNN network based on the work of \citeonline{peyrard2015comparison} trained on the Labeled Faces in the Wild (LFW) dataset \cite{huang2008labeled} and then refined the HR output with localized SR steps using autoencoders trained in patches of the images on the LFW. The localized SR step focused on locally reconstructing image patches at crucial face landmark points (e.g., eyes, nose, mouth) via dictionary learning. They claimed that the image reconstruction had a +2.80dB improvement, while the recognition performance also had a 3.94\% increase compared to the same results on x4 bicubic interpolation. However, they lacked tests in real-world LR datasets in order to check if their model would be able to keep the high performance within the wild data. Also, as having two networks in cascade is computationally expensive, this CNN architecture may not be ideal for real-world surveillance situations.

\citeonline{wang2016studying} presented an attempt to deal with the problem of very low-resolution recognition, where the region of interest could be smaller than 16x16 pixels. Their approach achieves feature enhancement and recognition simultaneously through the use of a deep SR network for pre-training with a carefully selected loss function for matching between LR and HR face images. The recognition step employed a deep neural network for classification trained on a different dataset that shared similar features with the one used for evaluation. They report a rise of 1.71\% in top-1 accuracy on the famous UCCS surveillance dataset, which is no longer publicly available.

\citeonline{abdollahi2019exploring} explored factors for improving LR face recognition using DL classifiers in two real-world surveillance datasets. Instead of focusing on SR approaches, they proposed two strategies to overcome the lack of fine information on the face images: increase the crop on a detected face before upsampling the image to match the input size of the classifier and match the resolution between gallery and probe images. For classification, they evaluated several ResNet-50 and SENet-50 architectures for feature extraction trained on VGGFace2 and the MS-Celeb-1M datasets. Along with a nearest neighbor classifier, they were able to achieve SOTA results in Rank-1 verification for the ICB-RW and SCFace surveillance datasets. Their work inspired this thesis in also investigating different face crop sizes and using the ICB-RW dabase as benchmark for LR face recognition.

\citeonline{elsayed2018unsupervised} evaluated the effects that SR and face alignment may have on accuracy for LR face recognition using an unsupervised approach. They proposed experiments where a LR version of the LFW dataset was frontalized and fed to a simple SR network based on SRCNN. Later they made use of an unsupervised recognition model using speed up robust features and local binary features. They tested only on the LFW data and reported that SR and face alignment increased the recognition performance.

\citeonline{ataer2019verification} proposed a two-stage architecture for simultaneous feature extraction and super-resolution. They trained a VGG-based deep face recognition network to be used as a feature extractor and trained an SR network to decrease the L1 distance between the features extracted from the VGG network for real and generated images. The evaluation procedure presented two DL based SR networks and showed that this setup increases recognition performance. However, they only evaluated the results in LR frontal images that were acquired after the downsampling of four HR datasets.

\citeonline{li2019low} presented results for LR face recognition in the wild with the evaluation of different deep learning SR network architectures in two originally LR datasets. They trained a VGG network using LR versions and HR versions of images from a HR dataset to extract features and then applied different classifiers for validation. Their best results were obtained by pre-training an SR architecture based on GAN in LR images of datasets with similar features to the ones used for evaluation. This trick helped their models reach close to SOTA recognition performance in the used datasets.

\citeonline{abello2019optimizingSR} explored the use of a loss function defined as the L2 error between face features from an super-resolved face and the ground truth for improving LR face recognition. For feature extraction, they used the pre-trained network with Inception architecture proposed by \citeonline{schroff2015facenet}. They reported that this loss was able to give improvements for both image quality and recognition performance. Nevertheless, they only tested their system in LR versions of the HR dataset used for training.  

\section{Final Considerations}

In this chapter, several works that present the SOTA for SR and LR face recognition are described with their results.
After analyzing what the trends present in the SOTA were, several architectures for SR were chosen for evaluation in real-world LR images. Also, the discussed related works led this thesis to a research direction that included the assessment of the use of different convolution operators and loss functions for better image quality and recognition accuracy.

For the next chapter, the methodology behind the experiments proposed in this thesis for assessing the hypotheses established in Chapter~\ref{sec:intro} are presented.

\chapter{Methodology}

This chapter describes the materials and methods used for experimenting with different architectures for super-resolving images and improving face recognition.

\section{Datasets}

\subsection{VGGFace2}

\begin{figure}[!ht]
	\begin{center}
		\includegraphics[width=1\textwidth]{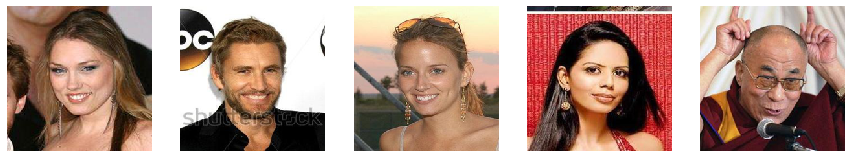}
	\end{center}
    \caption{\label{fig:vgg} Example of face images in the VGGFace2 dataset. (Adapted from \citeonline{cao2018vggface2})} 
\end{figure}

The VGGFace2 is an in the wild dataset of faces that contains 3.31 million images from 9131 celebrities downloaded from Google Image Search and shows significant variations in pose, age, lighting, and background~\cite{cao2018vggface2}. One advantage of using this dataset for training a robust image classifier/feature extractor is the fact that approximately 20\% of its images have pixel resolution lower than 50 pixels, which leads the model to have a better feature representation for low-resolution face images \cite{abdollahi2019exploring}. A set of five images from this dataset can be seen in Figure \ref{fig:vgg}.

\subsection{CelebA}

\begin{figure}[!ht]
	\begin{center}
		\includegraphics[width=1\textwidth]{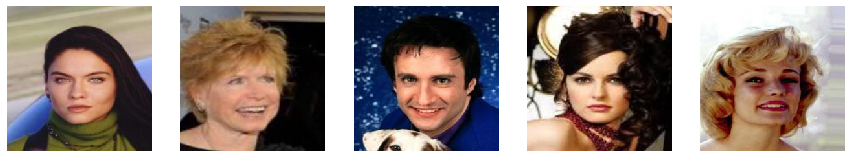}
	\end{center}
    \caption{\label{fig:celeba} Example of face images in the CelebA dataset. (Adapted from \citeonline{liu2015faceattributes})}
\end{figure}

The CelebFaces Attributes Dataset (CelebA) is a large-scale face attributes dataset with over 200,000 images of 10,117 celebrities around the world~\cite{liu2015faceattributes}. It presents a vast diversity of poses and background clutter. Also, it is commonly used for training SOTA SR networks because of its rich features and size~\cite{Chen2018FSRNetEL, yu2018face, kim2019progressive}. The first 18,000 images of this dataset were used for training, and the following 2,000 were used for validating the results for the SR networks according to the specified image quality metrics.  A set of 5 samples from this dataset is shown in Figure \ref{fig:celeba}

\subsection{Quis-Campi Dataset (ICB-RW)}

The Quis-Campi dataset is a growing biometric database comprising 3000 images from 320 subjects automatically acquired by an outdoor visual surveillance system, with subjects on-the-move and at-a-distance (up to 50 m). The system used for image acquisition has a master wide camera for subject detection and tracking, and a slave pan-tilt-zoom (PTZ) camera, as the foveal sensor, for extracting the facial region at a high-magnification state.

In the context of face verification, they supply three high-quality images of the subject in a controlled environment to be used as gallery data and several images of the same subject on the move inside an university campus to be used as probe data. One strong feature of this dataset is that all probe images present variation in illumination, pose, focus, expression, motion-blur, and occlusion~\cite{neves2017quis}.

Part of this dataset was published to promote the International Challenge on Biometric Recognition in the Wild (ICB-RW) competition, and that is why most of the results present the same benchmarking setup used in the competition. The ICB-RW challenge provided three face images to be used as gallery and five probe images for each of 90 subjects. 

\begin{figure}[!ht]
	\begin{center}
		\includegraphics[width=.6\textwidth]{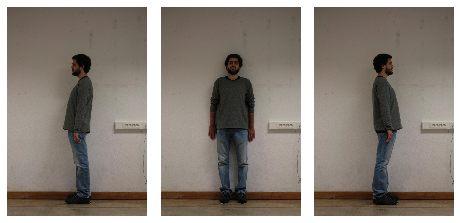}
	\end{center}
    \caption{\label{fig:icb-f} Example of gallery images in the Quis-Campi dataset. (Adapted from \citeonline{neves2017quis})}
\end{figure}

\begin{figure}[!ht]
	\begin{center}
		\includegraphics[width=1\textwidth]{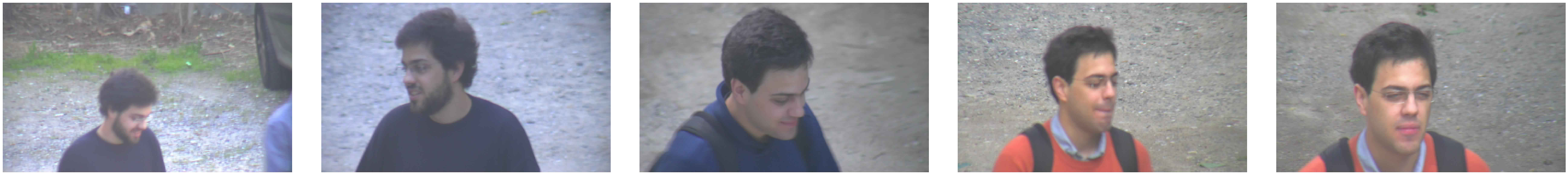}
	\end{center}
    \caption{\label{fig:icb-probe} Example of probe images in the Quis-Campi dataset. (Adapted from \citeonline{neves2017quis})}
\end{figure}

As the goal was to evaluate the performance of the proposed network architectures against other works in a real-word LR scenario, the Quis-Campi dataset was adapted to resemble the ICB-RW challenge to the maximum once the latter was not available. Therefore, since not all the subjects on the dataset had enough images to be selected, the first 90 subjects out of the 320 that had available the three gallery and five probe images were picked. This approach to making an equivalent representation of ICB-RW was considered instead of taking 90 random samples out of the 320 since, at the time of the 2016 competition, the Quis-Campi dataset did not have all of their subjects registered, and according to \citeonline{neves2016icb}, the new samples were registered and added automatically to the database. The images chosen for one of the subjects can be seen in Figures \ref{fig:icb-f} and \ref{fig:icb-probe}.

\vspace{5mm}
\subsection{Federal University of Sergipe Classroom Attendance Dataset}

\begin{figure}[!ht]
	\begin{center}
		\includegraphics[width=1\textwidth]{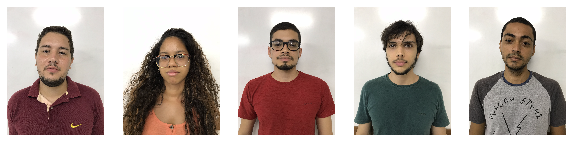}
	\end{center}
    \caption{\label{fig:ufs-class-1} Example of gallery images for the UFS-Classroom Attendance dataset. \cite{joao2019Automatic}}
\end{figure}

\begin{figure}[!ht]
	\begin{center}
		\includegraphics[width=1\textwidth]{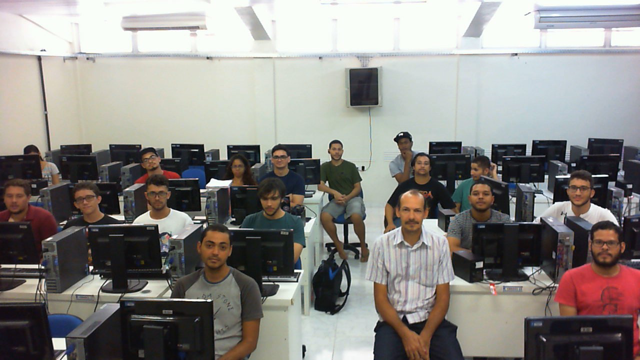}
	\end{center}
    \caption{\label{fig:ufs-class-2} Example of a probe image for the UFS-Classroom Attendance dataset. \cite{joao2019Automatic}}
\end{figure}

This dataset was formulated with the goal of creating an automated attendance system for classes within the computer science department of the Federal University of Sergipe (UFS)~\cite{joao2019Automatic}. The dataset is composed of one high-resolution frontal image of each student, referred to as a gallery image, and three probe images of a whole class taken in slightly different angles with a 1.2MP Webcam. For this thesis, three classes with different amount of students were used for the evaluation since this dataset presents a challenging LR uncontrolled environment ideal for testing real-world face recognition pipelines. An example of the gallery and probe images present on the dataset can be seen in Figures \ref{fig:ufs-class-1} and \ref{fig:ufs-class-2}, respectively.

\section{Data Pre-Processing}

For detecting and extracting the faces from the presented datasets, a pre-trained Multi-task Cascaded Convolutional Neural Network (MTCNN) was used since it has shown SOTA results on a variety of benchmarks for face detection and face alignment while keeping real-time results~\cite{zhang2016joint}.

For training and evaluating the SR networks, every image from the CelebA dataset was scaled to a [0,1] range, and then underwent a process of ``crappification'' to create an LR pair to be used since a paired supervised learning approach was adopted. Each cropped face was resized to 160x160 and saved as the HR sample. Then, for obtaining a ``crappy'' version of it, the same cropped face was resized to 40x40 pixels, and saved using JPEG compression with a quality factor that varied randomly from 10 to 70 (where 1 is the minimum, 75 the standard, and 95 the maximum quality) as advised by \citeonline{howard2018fastai} during the \textit{FastAI} course. This compression approach helps to simulate the data distribution that may be present in real-world LR surveillance footage. The resolution was chosen according to the input size of the deep learning model used for feature extraction (160x160x3), and due to limited computational resources, only the $4x$ SR upscaling setting was experimented.

\begin{figure}[!ht]
	\begin{center}
		\includegraphics[width=.6\textwidth]{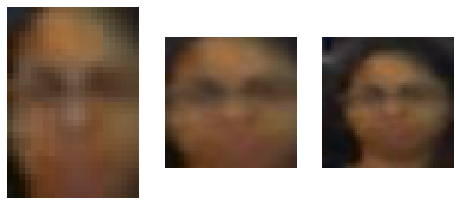}
	\end{center}
    \caption{\label{fig:margin-ex-1} Example of a probe face image from the UFS-Classroom Attendance dataset saved in the three settings. (Adapted from \citeonline{joao2019Automatic})}
\end{figure}

\begin{figure}[!ht]
	\begin{center}
		\includegraphics[width=.6\textwidth]{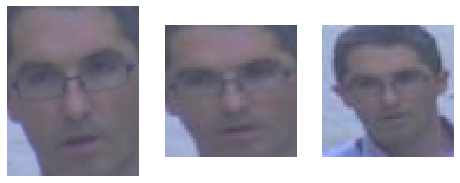}
	\end{center}
    \caption{\label{fig:margin-ex-2} Example of a probe face image from the Quis-Campi dataset saved in the three settings. (Adapted from \citeonline{neves2017quis})}
\end{figure}

As the primary task was to evaluate how SR may influence verification performance in real-world LR in-the-wild scenarios and faces detected in the Quis-Campi and UFS-Classroom Attendance datasets had a large variation in size due to differences in data acquisition equipment, every face detected (gallery and probe) was saved as an image in three different settings: without any change to the size, with bicubic resizing to 40x40 pixels, and with bicubic resizing to 40x40 pixels with a 1.3 crop margin to the borders. This last setup is able to increase the amount of information within the image and, added to the employed resolution matching step, can increase recognition performance for LR samples as validated by \citeonline{abdollahi2019exploring}. An example of the three cases for each dataset can be seen in Figures \ref{fig:margin-ex-1} and \ref{fig:margin-ex-2}.

\section{Transfer Learning}
\subsection{Face Feature Extraction}

Inspired by the work of \citeonline{schroff2015facenet}, a pre-trained network to extract feature embeddings of the faces for further comparison was employed. The chosen deep network architecture was the Inception-ResNet-V1 \cite{szegedy2017inception} trained on the VGGFace2 dataset. This network was able to achieve the SOTA accuracy of 0.9965 on the LFW benchmark. Compared to the Inception network architecture employed similarly in the Facenet paper, the Inception-ResNet-V1 network achieves faster convergence without adding additional computation complexity due to its residual connections. This network was trained for mapping an 160x160x3 image ($\mathbb{R}^{HxWxC}$) to a vector ($\phi(Image)$) in a feature space of 512x1 dimensions ($\mathbb{R}^{512}$).

\subsection{Face Verification}

The verification is performed by applying the nearest neighbor algorithm to check the distance among embeddings for the selected probe and gallery images. The metric employed for verification of closeness is the cosine similarity used previously by the winner of the ICB-RW competition~\cite{neves2016icb}, and also by \citeonline{abdollahi2019exploring}. A description of the metric can be seen in Equation \ref{eq:nn}. 

\begin{equation}
    Nearest \, Neighbor = 1 - \frac{\phi(I_{Face1}) \,\cdot\,\phi(I_{Face2})}{\left \|\phi(I_{Face1}) \right \|_{2} \, \left \|\phi(I_{Face2}) \right\|_{2}}
    \label{eq:nn}
\end{equation}

Most approaches that deal with the LR face recognition problem, as it was reviewed in Chapter~\ref{related-work}, try to train from scratch a robust network that may be able to overcome the difficulties of such task. However, this study approaches the problem from a different point of view since it takes advantage of pre-trained large and robust classifiers with the addition of a specially designed upscaling step including an SR network previous to the feature extraction and verification steps.

\begin{figure}[!htb]
	\begin{center}
		\includegraphics[width=1\textwidth]{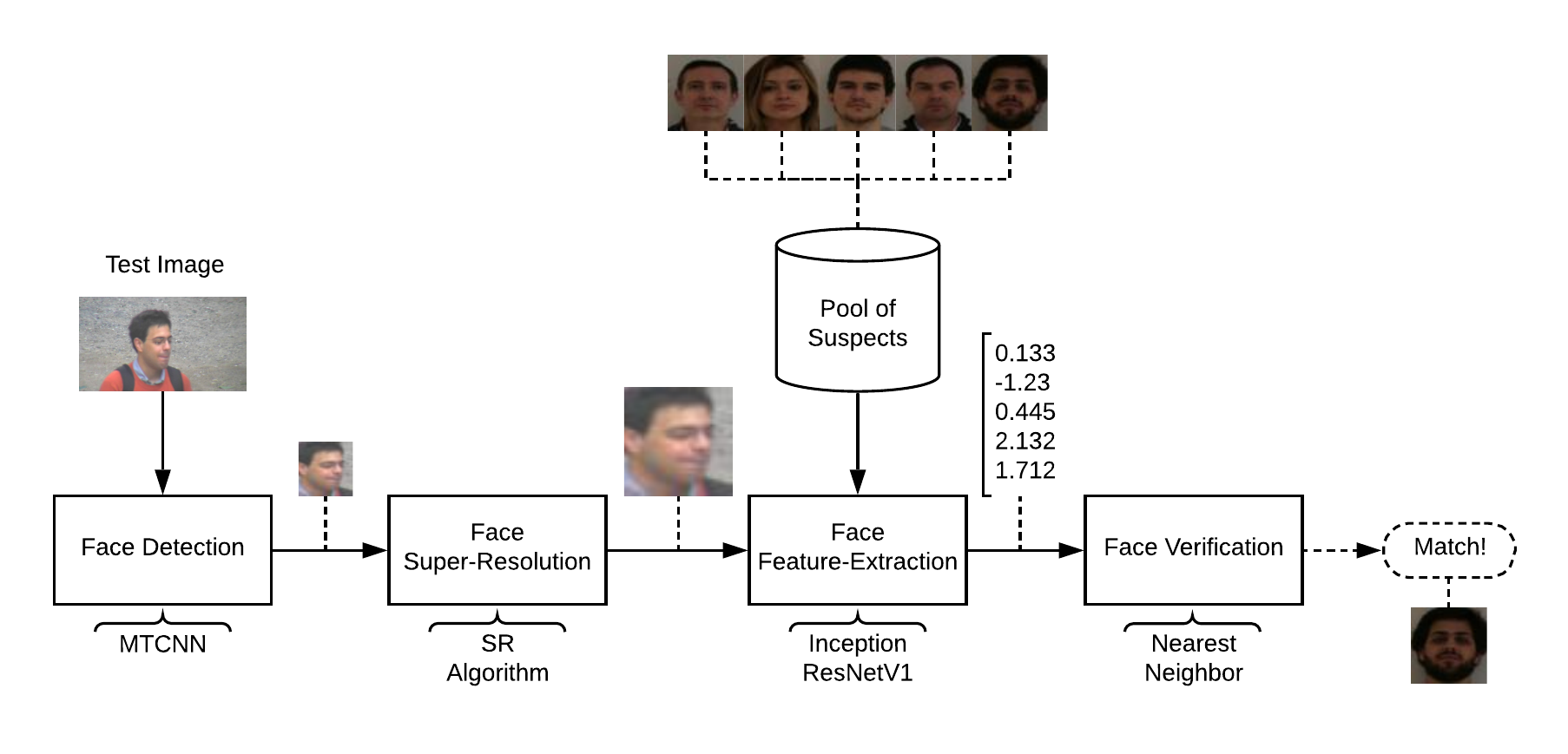}
	\end{center}
    \caption{\label{fig:pipeline} Pipeline for face verification in the ICB-RW. (Source: Author's own)}
\end{figure}

A general description of the whole pipeline for verification after the upsampling task, using the ICB-RW challenge as an example, can be seen in Figure \ref{fig:pipeline}.

\subsection{Face Loss}\label{sec:faceloss}

Mean-squared-error loss optimizes the response of an SR network for generating images with better quality, but it does not take into account if the recognized person has kept the same unique features that may differentiate this person from the others. As a task-driven approach for SR was meant to be developed, the face identity loss commonly used for face normalization~\cite{cole2017synthesizing} and 3D face reconstruction~\cite{gecer2019ganfit} was adopted to guide the SR process for better face recognition accuracy.

The SOTA face feature extractor (Inception-ResNet-V1) was used for checking if the distance between the embedding of the real and the super-resolved image was decreasing within each epoch. To accomplish that, the cosine similarity measure (Equation \ref{eq:nn}) of the face embeddings extracted from the image pairs was added to the standard image loss used for training each SR network. This customized loss ensures that the reconstruction made by the SR network resembles the target identity under various conditions in the feature space. The definition for this loss can be seen in Equation~\ref{eq:faceloss}.

\begin{equation}
    Face \, Loss = 1 - \frac{\phi(I^{SR}) \,\cdot\,\phi(I^{HR})}{\left \|\phi(I^{SR}) \right \|_{2} \, \left \|\phi(I^{HR}) \right\|_{2}}
    \label{eq:faceloss}
\end{equation}

This approach adopts the same concept of the task-driven loss presented by \citeonline{haris2018task}, but it kept the focus on a more robust and general task (face recognition). In addition, the Face Loss presented here differs from the recent work of \citeonline{abello2019optimizingSR} since they applied L2 error, which is more susceptible to outliers, on the feature vectors for the optimization of their SR network. Moreover, in the end, they neither provided an ablation study nor validated their approach for accuracy improvement in real-world LR data.

\section{Final Considerations}

This chapter described the datasets and the steps taken towards finding the best DL architecture for super-resolving images with the goal of improving face recognition. To ensure the models work in real-world LR data, two challenging LR datasets (surveillance and attendance assessment) were used for evaluation. Also, for making sure the super-resolved face images have discriminant features regarding identity, a custom loss function was proposed.

For the next chapter, the experiments of this thesis are presented and explained. Then, their results are broadly discussed according to the hypotheses idealized in Chapter~\ref{sec:intro}.

\chapter{Experiments}

This section provides an overview of the performed experiments with their respective parameterization.

\section{Task 1 - Face Super-Resolution}\label{sec:exp1}

For performing this task, some of the CNN architectures discussed in Section \ref{related-work} were modified and implemented following the implementation details on their papers in order to evaluate the best choice for a face verification pipeline. They were chosen based on previously reported results and computational complexity. The variants that have the name ``Coord'' kept the same original architecture but had the first ``Conv2d'' layer switched to a ``CoordConv''. The variants with ``FaceLoss'' had their training with the addition of the customized loss function presented in Section \ref{sec:faceloss}. The implemented models for SR described in this thesis can be checked on \url{https://github.com/angelomenezes/Pytorch_Face_SR}.

Network Architectures evaluated:
\begin{enumerate}
    \item SRCNN (\citeonline{dong2015image})
    \item SRCNN Coord
    \item Subpixel CNN (\citeonline{shi2016real})
    \item Subpixel CNN Coord
    \item FSRCNN (\citeonline{dong2016accelerating})
    \item FSRCNN Coord
    \item FSRCNN Coord FaceLoss
    \item SRGAN (\citeonline{ledig2017photo})
    \item SRGAN Coord
    \item SRGAN FaceLoss
    \item SRGAN Coord FaceLoss
\end{enumerate}

\begin{itemize}
    \item \textbf{Hyperparameters} $\rightarrow$  SRCNN and SRCNN Coord
    \begin{itemize}
        \item Batch size: 64
        \item Number of epochs: 50
        \item Loss: MSE
        \item Optimizer: Adam
        \item Learning Rate: 0.01 with decay to 10\% of the current learning rate every 15 steps 
    \end{itemize}
    \item \textbf{Hyperparameters} $\rightarrow$ Subpixel CNN and Subpixel CNN Coord
    \begin{itemize}
        \item Batch size: 32
        \item Number of epochs: 50
        \item Loss: MSE
        \item Optimizer: Adam
        \item Learning Rate: 0.01 with decay to 20\% of the current learning rate every 15 steps 
    \end{itemize}
    \item \textbf{Hyperparameters} $\rightarrow$ FSRCNN, FSRCNN Coord and FSRCNN Coord FaceLoss
    \begin{itemize}
        \item Batch size: 32
        \item Number of epochs: 50 (30 for FSRCNN Coord FaceLoss)
        \item Loss: MSE and variant that added FaceLoss
        \item Optimizer: Adam
        \item Learning Rate: 0.001 with decay to 20\% of the current learning rate every 15 steps 
    \end{itemize}
    \item \textbf{Hyperparameters} $\rightarrow$ SRGAN, SRGAN FaceLoss, SRGAN Coord and SRGAN Coord FaceLoss
    \begin{itemize}
        \item Batch size: 32
        \item Number of epochs: 30
        \item Loss: MSE + Adversarial Loss + Perceptual Loss and variant that added FaceLoss 
        \item Optimizer: Adam with $\beta_{1} = 0.5$ and $\beta_{2} = 0.999$
        \item Learning Rate: 0.001 with decay to 20\% of the current learning rate every 15 steps 
    \end{itemize}
\end{itemize}

\section{Task 2 - Watch-List ICB-RW (1x5 Problem)}\label{sec:exp2}

As proposed by \citeonline{neves2016icb}, the ICB-RW challenge used part of the Quis-Campi dataset to evaluate the average Rank-1 identification of a suspect against the ``watch-list'' subjects. For each probe image, the model had to output a similarity score related to each of five possible suspects. An example of this setup is presented in Figure \ref{fig:icb-watchlist}.

\begin{figure}[!ht]
	\begin{center}
		\includegraphics[width=.8\textwidth]{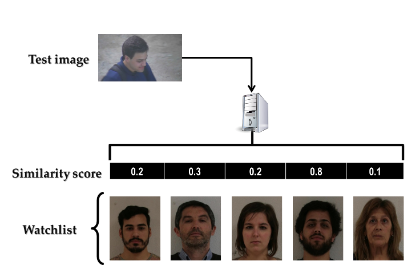}
	\end{center}
    \caption{\label{fig:icb-watchlist} Watch-List setting for the ICB-RW.~\cite{neves2016icb}}
\end{figure}

For this experiment, each individual had its frontal gallery image and a random probe image selected along with four random probe images of different subjects. The challenge is to obtain the highest number of matches according to the least distance given by the nearest neighbor algorithm.


\section{Task 3 - Attendance Evaluation (1xN Problem)}\label{sec:exp3}

For the task of evaluating the attendance inside a classroom, the identity of every student needs to be checked against all entries on the attendance list. The number of students in each class can be seen in Table \ref{tab:nstudents}.

\begin{table}[!htb]
\centering
\caption{Number of students for each class}
\label{tab:nstudents}
\begin{tabular}{cc}
\hline
        & N° of Students \\ \hline
Class 1 & 15             \\
Class 2 & 16             \\
Class 3 & 12             \\ \hline
\end{tabular}
\end{table}

This experiment followed the same verification principle of Task 2. Nevertheless, it is a more challenging situation since this recognition task takes into account all the subjects in the classroom (1 vs. ALL problem).

Each experiment described in this section was either run in a personal computer with an Intel i7-6500U with 16GB of memory and GeForce GTX 950M (4GB) or in a Google Cloud instance with a Skylake processor (8 vCPUs and 52 GB memory) with an NVIDIA Tesla P100.
\chapter{Experimental Results}\label{sec:results}

\section{Experiments}

This section provides an overview of the performed experiments with their respective parameterization.

\subsection{Task 1 - Face Super-Resolution}\label{sec:exp1}

For performing this task, 4 of the CNN architectures discussed in Chapter \ref{related-work} were modified and implemented following the implementation details on their papers in order to evaluate the best choice for a face verification pipeline. These models were chosen based on previously reported results and computational complexity. In this thesis, 7 different new architectures were proposed for evaluation. The variants that have the name ``Coord'' kept the same original architecture but had the first ``Conv2d'' layer switched to a ``CoordConv''. The variants with ``FaceLoss'' had their training with the addition of the customized loss function presented in Section \ref{sec:faceloss}. The implemented models for SR described below can be checked on \url{https://github.com/angelomenezes/Pytorch_Face_SR}.

Network architectures from literature that were evaluated:
\begin{enumerate}
    \item SRCNN (\citeonline{dong2015image}) shown in Figure~\ref{fig:srcnn}.
        \begin{figure}[!htb]
    	\begin{center}
    		\includegraphics[width=.7\textwidth]{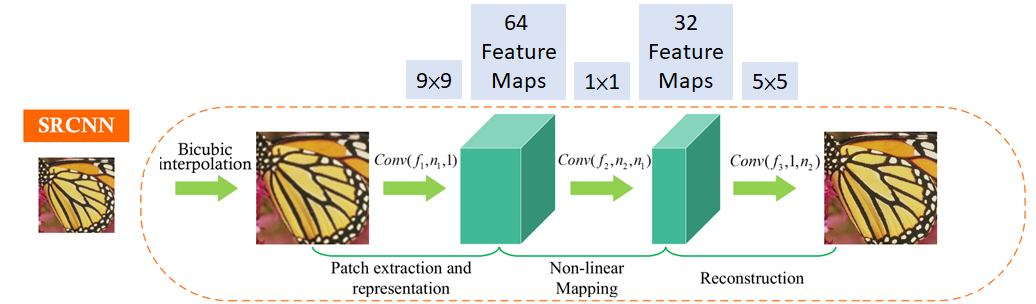}
    	\end{center}
        \caption{\label{fig:srcnn}SRCNN architecture.~\cite{dong2016accelerating}}
    \end{figure}
    \item Subpixel CNN (\citeonline{shi2016real}) shown in Figure~\ref{fig:subcnn}.
    \begin{figure}[!htb]
    	\begin{center}
    		\includegraphics[width=.7\textwidth]{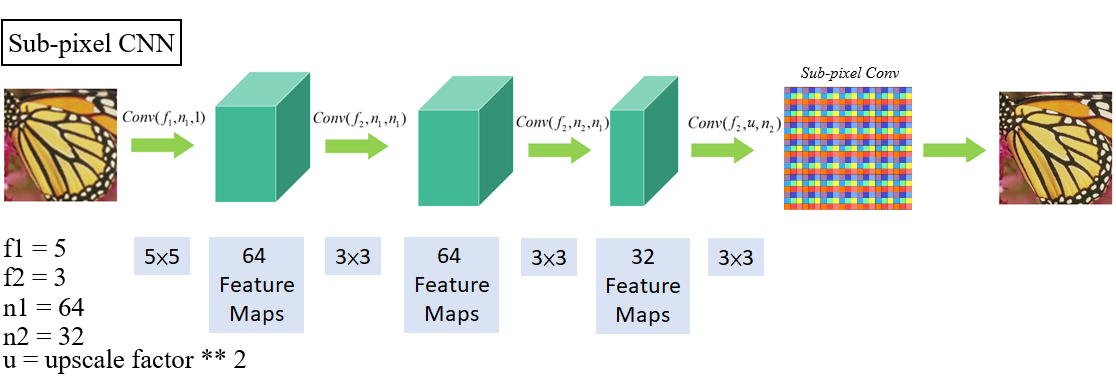}
    	\end{center}
        \caption{\label{fig:subcnn}Subpixel CNN architecture. (Adapted from \citeonline{dong2016accelerating} and \citeonline{shi2016real})}
    \end{figure}
    \item FSRCNN (\citeonline{dong2016accelerating}) shown in Figure~\ref{fig:fsrcnn}.
    \begin{figure}[!htb]
	\begin{center}
		\includegraphics[width=.7\textwidth]{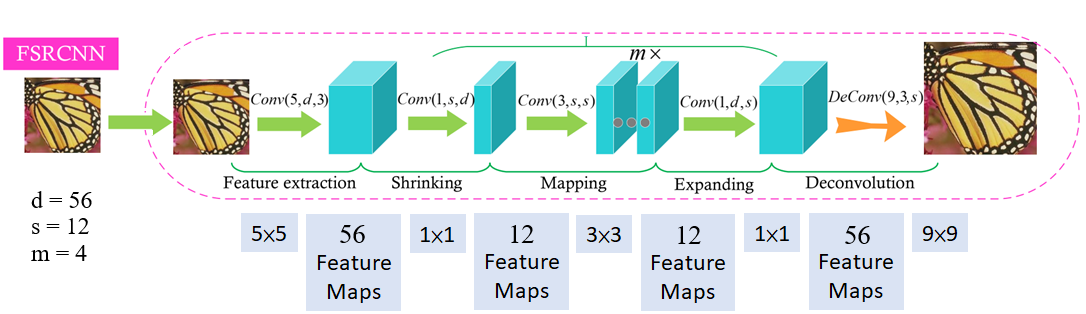}
	\end{center}
    \caption{\label{fig:fsrcnn}FSRCNN architecture.~\cite{dong2016accelerating}}
    \end{figure}
    \item SRGAN (\citeonline{ledig2017photo}) shown in Figure~\ref{fig:srgan}.
    \begin{figure}[!htb]
	\begin{center}
		\includegraphics[width=.75\textwidth]{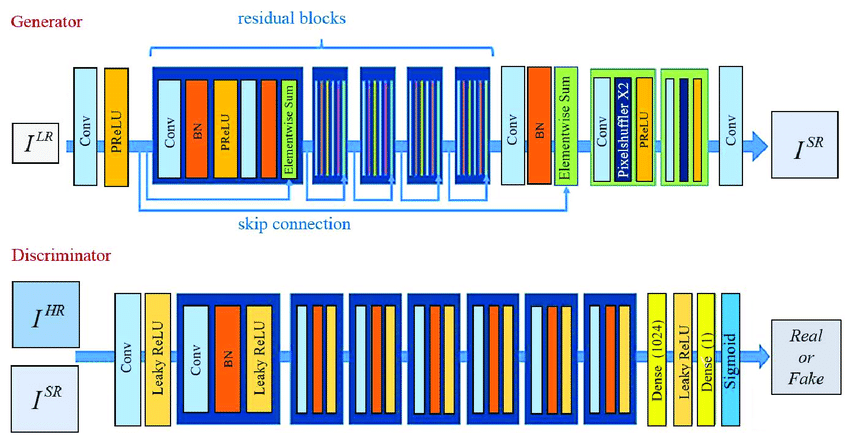}
	\end{center}
    \caption{\label{fig:srgan}SRGAN architecture.~\cite{jiao2019survey}}
    \end{figure}
\end{enumerate}

\newpage

Network architectures with modifications proposed in this thesis for evaluation:
\begin{enumerate}
    \item SRCNN Coord
    \item Subpixel CNN Coord
    \item FSRCNN Coord
    \item FSRCNN Coord FaceLoss
    \item SRGAN Coord
    \item SRGAN FaceLoss
    \item SRGAN Coord FaceLoss
\end{enumerate}

\begin{itemize}
    \item \textbf{Hyperparameters} $\rightarrow$  SRCNN and SRCNN Coord
    \begin{itemize}
        \item Batch size: 64
        \item Number of epochs: 50
        \item Loss: MSE
        \item Optimizer: Adam
        \item Learning Rate: 0.01 with decay to 10\% of the current learning rate every 15 steps 
    \end{itemize}
    \item \textbf{Hyperparameters} $\rightarrow$ Subpixel CNN and Subpixel CNN Coord
    \begin{itemize}
        \item Batch size: 32
        \item Number of epochs: 50
        \item Loss: MSE
        \item Optimizer: Adam
        \item Learning Rate: 0.01 with decay to 20\% of the current learning rate every 15 steps 
    \end{itemize}
    \item \textbf{Hyperparameters} $\rightarrow$ FSRCNN, FSRCNN Coord and FSRCNN Coord FaceLoss
    \begin{itemize}
        \item Batch size: 32
        \item Number of epochs: 50 (30 for FSRCNN Coord FaceLoss)
        \item Loss: MSE and variant that added FaceLoss
        \item Optimizer: Adam
        \item Learning Rate: 0.001 with decay to 20\% of the current learning rate every 15 steps 
    \end{itemize}
    \newpage
    \item \textbf{Hyperparameters} $\rightarrow$ SRGAN, SRGAN FaceLoss, SRGAN Coord and SRGAN Coord FaceLoss
    \begin{itemize}
        \item Batch size: 32
        \item Number of epochs: 30
        \item Loss: MSE + Adversarial Loss + Perceptual Loss and variant that added FaceLoss 
        \item Optimizer: Adam with $\beta_{1} = 0.5$ and $\beta_{2} = 0.999$
        \item Learning Rate: 0.001 with decay to 20\% of the current learning rate every 15 steps 
    \end{itemize}
\end{itemize}

\subsection{Task 2 - Watch-List ICB-RW (1x5 Problem)}\label{sec:exp2}

As proposed by \citeonline{neves2016icb}, the ICB-RW challenge used part of the Quis-Campi dataset to evaluate the average Rank-1 identification of a suspect against the ``watch-list'' subjects. For each probe image, the model had to output a similarity score related to each of five possible suspects. An example of this setup is presented in Figure \ref{fig:icb-watchlist}.

\begin{figure}[!ht]
	\begin{center}
		\includegraphics[width=.8\textwidth]{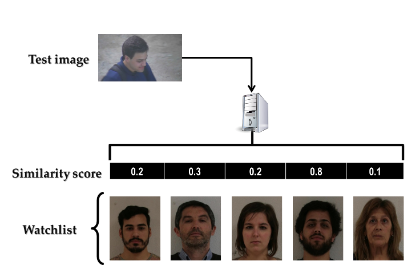}
	\end{center}
    \caption{\label{fig:icb-watchlist} Watch-List setting for the ICB-RW.~\cite{neves2016icb}}
\end{figure}

For this experiment, each individual had its frontal gallery image and a random probe image selected along with four random probe images of different subjects. The challenge is to obtain the highest number of matches according to the smallest distance given by the nearest neighbor algorithm.


\subsection{Task 3 - Attendance Evaluation (1xN Problem)}\label{sec:exp3}

For the task of evaluating the attendance inside a classroom, the identity of every student needs to be checked against all entries on the attendance list. The number of students in each class can be seen in Table \ref{tab:nstudents}.

\begin{table}[!htb]
\centering
\caption{Number of students for each class}
\label{tab:nstudents}
\begin{tabular}{cc}
\hline
        & N° of Students \\ \hline
Class 1 & 15             \\
Class 2 & 16             \\
Class 3 & 12             \\ \hline
\end{tabular}
\end{table}

This experiment followed the same verification principle of Task 2. Nevertheless, it is a more challenging situation since this recognition task takes into account all the subjects in the classroom (1 vs. ALL problem).

Each experiment described in this chapter was either run in a personal computer with an Intel i7-6500U with 16 GB of memory and GeForce GTX 950M (4 GB) or in a Google Cloud instance with a Skylake processor (8 vCPUs and 52 GB memory) with an NVIDIA Tesla P100 (16 GB). All models were implemented and evaluated using Python and the Pytorch library.

\section{Results Evaluation}

When training the SRGAN and its variants, if the discriminator network had its weights updated in the same frequency of the generator, its loss would quickly converge to zero, and both networks would not have any gradients for learning along the epochs. Therefore, in order to make the learning happen, two extra steps were followed according to the tips given on the \textit{FastAI} course~\cite{howard2018fastai}:

\begin{itemize}
    \item \textbf{Generator Pre-Train}: the generator network was pre-trained for 5 epochs using only the MSE loss in order to have some advantage initially against the discriminator.
    \item \textbf{Smart update for the Discriminator}: across the epochs, the discriminator network was only trained (had its weights updated) when its loss was above a threshold (0.5). This step ensures that the network is learning gradually to assess the output of the generator since the update of weights for the discriminant only happened when it was making more ``mistakes'' according to the right labels for each input. All the training losses can be evaluated in Appendix \ref{appendix:graph}.
\end{itemize}

The results for each proposed task are shown and discussed in the following subsections.

\subsection{Task 1 - Face Super-Resolution}

The validation results regarding image quality metrics and inference results for the SR architectures are presented in Table~\ref{tab:psnr-results}. The PSNR was calculated on the RGB channels, and the average time for inference was computed as the average of 10 runs of each algorithm. The perceptual results can be evaluated in the Appendix \ref{appendix:imgs}.

As can be seen in Table \ref{tab:psnr-results}, except for the ``SRGAN Coord FaceLoss'', all the presented architectures achieve real-time performance in a small GPU (GeForce GTX 950M). The best algorithm regarding quality metrics was the FSRCNN with the coordinated convolution operator. Nonetheless, the SRGAN and its variants, even with a low PSNR, had the best human perceptual quality as can be seen by the perceptual clarity of the outputs in the Appendix \ref{appendix:imgs}. Their better performance may be related to the fact that they make use of different losses, which optimizes for a less blur and more textured output (Adversarial and Perceptual Loss).

\begin{table}[!htb]
\centering
\caption{PSNR and SSIM validation results (2000 images from CelebA)}
\label{tab:psnr-results}
\begin{tabular}{cccccc}
\hline
\multicolumn{6}{c}{Validation metric results for 4x Upscaling}                                                                                                                                                                                                       \\ \hline
                                              & PSNR           & SSIM            & \begin{tabular}[c]{@{}c@{}}Avg. Time \\ for Inference (s)\end{tabular} & \begin{tabular}[c]{@{}c@{}}Inference \\ FPS in GPU\end{tabular} & \begin{tabular}[c]{@{}c@{}}Trained on \\ Channel\end{tabular} \\ \hline
SRCNN                                         & 27.95          & 0.7973          & 0.0100                                                                 & 100.00                                                          & Y                                                             \\
{\color[HTML]{FE0000} SRCNN Coord}            & 27.98          & 0.7966          & 0.0153                                                                 & 65.36                                                           & Y                                                             \\
SubCNN                                        & 28.08          & 0.8003          & 0.0035                                                                 & 285.71                                                          & Y                                                             \\
{\color[HTML]{FE0000} SubCNN Coord}           & 28.13          & 0.8022          & 0.0048                                                                 & 208.33                                                          & Y                                                             \\
FSRCNN                                        & 28.45          & 0.8104          & 0.0046                                                                 & 217.39                                                          & RGB                                                           \\
{\color[HTML]{FE0000} FSRCNN Coord}           & \textbf{28.88} & \textbf{0.8175} & 0.0048                                                                 & 208.33                                                          & RGB                                                           \\
{\color[HTML]{FE0000} FSRCNN Coord Face Loss} & 28.78          & 0.8151          & 0.0047                                                                 & 212.77                                                          & RGB                                                           \\
SRGAN                                         & 27.02          & 0.8077          & 0.0336                                                                 & 29.76                                                           & RGB                                                           \\
{\color[HTML]{FE0000} SRGAN Coord}            & 27.28          & 0.8078          & 0.0382                                                                 & 26.18                                                           & RGB                                                           \\
{\color[HTML]{FE0000} SRGAN FaceLoss}         & 26.63          & 0.8083          & 0.0384                                                                 & 26.04                                                           & RGB                                                           \\
{\color[HTML]{FE0000} SRGAN Coord FaceLoss}   & 26.69          & 0.7984          & 0.0404                                                                 & 24.75                                                           & RGB                                                           \\
Bicubic                                       & 27.93          & 0.7881          & 0.0012                                                                 & 833.33                                                          & -                                                             \\ \hline
\multicolumn{6}{r}{$\rightarrow$ {\color[HTML]{FE0000}Red color} highlights the architectures proposed in this thesis.}
\\
\end{tabular}
\end{table}

Architectures that were modified with the CoordConv operator improved its PSNR 100\% of the times with an average increase of 0.16 dB. Their SSIM presented fluctuations, and no significative gains could be measured. This situation may be explained by the fact that all architectures were optimized to decrease the difference in pixels according to the MSE loss, which directly improves the PSNR of the image but often smooths and blurs the output. Such blur decreases the perceptual quality of the image, and consequently, does the same to perceptual quality metrics such as the SSIM. 

The SR models that were optimized using the customized loss function (FaceLoss) presented in general lower image quality metrics than their pairs that used their own objective function. They also presented a longer time for inference, which may be an indication that fewer weights were zero since more computation was measured. Despite that, every evaluated architecture was able to reach real-time inference in the small GPU used for training and testing the models. The interpolation method (Bicubic) was still around three times faster than the fastest SR model. This can indicate that, for situations where processing time weights more than accuracy, interpolation methods can still be a considered direction.

\subsection{Task 2 - Watch-List ICB-RW (1x5 Problem)}

The results for the task of recognizing which suspect was correctly identified by the surveillance camera can be seen in Table \ref{tab:icbrw}. Fine-tuning was not performed. Therefore, the recognition pipeline did not have direct samples that resembled the gallery/probe data of this experiment.

The best obtained results came from SRGAN and its variants. In the setting where the face image was not resized previously, only these algorithms were able to overcome the baseline (Bicubic). When the cropped face had its resolution reduced for simulating lower resolution scenarios, they still were the best performing group, but other networks were also able to beat the interpolation method.

\begin{table}[!htb]
\centering
\caption{Rank-1 Accuracy (in \%) for Face Recognition task (1xN) on 90 subjects from the Quis-Campi dataset (ICB-RW)}
\label{tab:icbrw}
\begin{tabular}{cccc}
    \hline
    \multicolumn{4}{c}{\begin{tabular}[c]{@{}c@{}}Results Quis-Campi / ICB-RW \\ (N = 5 suspects).\end{tabular}} \\ \hline
                                 & No Resize No Margin       & Size 40 No Margin       & Size 40 Margin 1.3      \\ \hline
    SRCNN                        & 74.89                   & 61.78                   & 64.67                   \\
    {\color[HTML]{FE0000}SRCNN Coord}                  & 78.22                   & 62.89                   & 67.11                   \\
    SubCNN                       & 64.67                   & 55.33                   & 57.56                   \\
    {\color[HTML]{FE0000}SubCNN Coord}                 & 67.33                   & 58.44                   & 62.22                   \\
    FSRCNN                       & 72.00                   & 60.22                   & 65.56                   \\
    {\color[HTML]{FE0000}FSRCNN Coord}                 & 78.22                   & 64.00                   & 69.11                   \\
    {\color[HTML]{FE0000}FSRCNN Coord Face Loss}       & 78.44                   & 63.33                   & 67.33                   \\
    SRGAN                        & \textbf{85.78}          & 77.33                   & \textbf{72.00}          \\
    {\color[HTML]{FE0000}SRGAN Coord}                  & 83.80                   & \textbf{78.40}          & 68.90                   \\
    {\color[HTML]{FE0000}SRGAN FaceLoss}               & 85.11                   & 78.22                   & 71.78                   \\
    {\color[HTML]{FE0000}SRGAN Coord FaceLoss}         & 84.89                   & 76.00                   & 71.11                   \\
    Bicubic                      & 83.11                   & 62.00                   & 62.44                   \\
    \hline
    \multicolumn{4}{r}{$\rightarrow$ {\color[HTML]{FE0000}Red color} highlights the architectures proposed in this thesis.}
    \\
\end{tabular}
\end{table}

As the resolution of the faces on the probe data for the ICB-RW had naturally almost the same resolution as the ones for the gallery (around 200x200), the 1.3 margin did not have much effect on the accuracy results, which was also noticed in the work of \citeonline{abdollahi2019exploring}. As in this experiment the sizes of the gallery and probe data were always matched before upsampling, the 40x40 resizing may have caused a drastic loss of high-frequency details and discriminative features, which can be noticed by the decrease in accuracy ratings.

Every accuracy result on this task, except for the SubCNN and its variant, overcame the results of \citeonline{ghaleb2018deep}, the best performing system in the ICB-RW challenge at that time. He was able to achieve a Rank-1 IR rate of 71.7\%, which differs from the proposed SRGAN and SRGAN FaceLoss by a margin of 17.08\% and 16.41\%, respectively. These two architectures would also beat the results of \citeonline{abdollahi2019exploring}, who was able to achieve a Rank-1 rate of 84.22\%, the highest registered so far.

\subsection{Task 3 - Attendance Evaluation (1xN Problem)}

The results for the task of evaluating which students are present in each of the classrooms can be seen in Tables \ref{tab:class1}, \ref{tab:class2}, and \ref{tab:class3}. Similarly to the other task, neither training nor fine-tuning was performed using gallery/probe data for this experiment.

For Classroom 1, the best performing algorithm for all settings was the SRGAN and its variants. The FSRCNN model presented the highest obtained accuracy, but it did not keep a performance consistency. For the setting where no resize was employed, all architectures were able to beat the baseline. Yet, when the margin was applied to increase the amount of information within the image, bicubic interpolation overcame even the SRGAN and two of its variants. 

\begin{table}[!htb]
\centering
\caption{Accuracy (in \%) for Face Recognition task in Classroom 1 (1xN)}
\label{tab:class1}
\begin{tabular}{cccc}
\hline
\multicolumn{4}{c}{\begin{tabular}[c]{@{}c@{}}Results UFS Classroom 1 \\ (N = 15 students)\end{tabular}} \\ \hline
                                  & No Resize No Margin            & Size 40 No Margin           & Size 40 Margin 1.3           \\ \hline
SRCNN                             & 62.50               & 66.67                       & 83.33                        \\ 
{\color[HTML]{FE0000}SRCNN Coord}                       & 64.58                        & 70.83                       & 85.42                        \\
SubCNN                            & 68.75                        & 64.58                       & 66.67                        \\ 
{\color[HTML]{FE0000}SubCNN Coord}                      & 70.83                        & 66.67                       & 68.75                        \\ 
FSRCNN                            & 66.67                        & 72.92                       & 83.33                        \\ 
{\color[HTML]{FE0000}FSRCNN Coord}                      & 66.67                        & 66.67                       & \textbf{89.58}               \\
{\color[HTML]{FE0000}FSRCNN Coord Face Loss}            & 64.58                        & 68.75                       & \textbf{89.58}               \\ 
SRGAN                             & \textbf{85.42}               & 81.25                       & 75.00                           \\ 
{\color[HTML]{FE0000}SRGAN Coord}                       & 81.25                        & 79.17                       & 81.25                        \\ 
{\color[HTML]{FE0000}SRGAN FaceLoss}                    & \textbf{85.42}               & \textbf{83.33}              & 83.33                        \\ 
{\color[HTML]{FE0000}SRGAN Coord FaceLoss}              & 81.25                        & 81.25                       & 79.17                        \\ 
Bicubic                           & 58.33                        & 66.67                       & 83.33                        \\ \hline
\multicolumn{4}{r}{$\rightarrow$ {\color[HTML]{FE0000}Red color} highlights the architectures proposed in this thesis.}
\\
\end{tabular}
\end{table}

For Classroom 2, it was possible to conclude that the images had a high degree of degradation since the highest accuracy was around 73\%, and the setting without resizing and margin adjustment had around 55\%. The best obtained results came from the SRGAN with coordinate convolution in the setting where the margin was adjusted. All the other SRGAN related models happened to hit the same accuracy, which gives a hint that they might have similar weights.

For Classroom 3, probe images might have had a higher resolution than in previous classroom experiments since the results for the setting without margin adjustment presented the highest rate, similar to the results on the simulated ICB-RW benchmark. The most consistent models were the proposed architectures based on SRGAN with results around 80\%, which overcame the baseline in every possible setting. However, when the size and margin were adjusted, FSRCNN presented the best results and the outcome for the other models became similar.

\begin{table}[!htb]
\centering
\caption{Accuracy (in \%) for Face Recognition task in Classroom 2 (1xN)}
\label{tab:class2}
\begin{tabular}{cccc}
\hline
\multicolumn{4}{c}{\begin{tabular}[c]{@{}c@{}}Results UFS Classroom 2 \\ (N = 16 students)\end{tabular}} \\ \hline
                                  & No Resize No Margin            & Size 40 No Margin           & Size 40 Margin 1.3           \\ \hline
SRCNN                             & 26.56                        & 23.99                       & 58.06                        \\ 
{\color[HTML]{FE0000}SRCNN Coord}                       & 26.56                        & 31.32                       & 50.92                        \\ 
SubCNN                            & 23.99                        & 28.94                       & 41.39                        \\ 
{\color[HTML]{FE0000}SubCNN Coord}                      & 21.61                        & 16.85                       & 43.77                        \\ 
FSRCNN                            & 33.88                        & 26.56                       & 51.10                         \\
{\color[HTML]{FE0000}FSRCNN Coord}                      & 31.50                         & 26.56                       & 55.86                        \\ 
{\color[HTML]{FE0000}FSRCNN Coord Face Loss}            & 33.88                        & 26.56                       & 55.86                        \\ 
SRGAN                             & 39.01                        & 45.97                       & 67.95                        \\ 
{\color[HTML]{FE0000}SRGAN Coord}                       & 38.83                        & 36.45                       & \textbf{72.71}               \\ 
{\color[HTML]{FE0000}SRGAN FaceLoss}                    & 48.72                        & \textbf{48.72}              & 67.95                        \\ 
{\color[HTML]{FE0000}SRGAN Coord FaceLoss}              & \textbf{53.66}               & 38.64                       & 67.95                        \\ 
Bicubic                           & 26.74                        & 26.56                       & 55.49                        \\ \hline
\multicolumn{4}{r}{$\rightarrow$ {\color[HTML]{FE0000}Red color} highlights the architectures proposed in this thesis.}
\\
\end{tabular}
\end{table}

\begin{table}[!htb]
\centering
\caption{Accuracy (in \%) for Face Recognition task in Classroom 3 (1xN)}
\label{tab:class3}
\begin{tabular}{cccc}
\hline
\multicolumn{4}{c}{\begin{tabular}[c]{@{}c@{}}Results UFS Classroom 3 \\ (N = 12 students)\end{tabular}} \\ \hline
                                  & No Resize No Margin            & Size 40 No Margin           & Size 40 Margin 1.3           \\ \hline
SRCNN                             & 40.00                           & 40.00                          & 66.67                        \\ 
{\color[HTML]{FE0000}SRCNN Coord}                       & 36.67                        & 56.67                       & 56.67                        \\ 
SubCNN                            & 40.00                           & 46.67                       & 60.00                           \\ 
{\color[HTML]{FE0000}SubCNN Coord}                      & 40.00                           & 43.33                       & 60.00                           \\ 
FSRCNN                            & 46.67                        & 43.33                       & \textbf{80.00}                  \\ 
{\color[HTML]{FE0000}FSRCNN Coord}                      & 46.67                        & 43.33                       & 73.33                        \\ 
{\color[HTML]{FE0000}FSRCNN Coord Face Loss}            & 40.00                           & 33.33                       & \textbf{80.00}                  \\ 
SRGAN                             & 70.00                           & 63.33                       & 70.00                           \\ 
{\color[HTML]{FE0000}SRGAN Coord}                       & 76.67                        & 76.67                       & 73.33                        \\ 
{\color[HTML]{FE0000}SRGAN FaceLoss}                    & 83.33                        & 76.67                       & 70.00                           \\ 
{\color[HTML]{FE0000}SRGAN Coord FaceLoss}              & \textbf{86.67}               & \textbf{80.00}                 & 70.00                           \\ 
Bicubic                           & 46.67                        & 40.00                          & 66.67                        \\ \hline
\multicolumn{4}{r}{$\rightarrow$ {\color[HTML]{FE0000}Red color} highlights the architectures proposed in this thesis.}
\\
\end{tabular}
\end{table}

It was possible to check that, for all experiments in this task, increasing the amount of information within the image with a 1.3 margin resulted in an increase in accuracy for most algorithms. This increase did help the SR network to provide more discriminative face images for the feature extractor since the average accuracy results were higher in general for such setting.

\begin{figure}[!htb]
\centering
\begin{subfigure}{.5\textwidth}
  \centering
  \includegraphics[width=1\linewidth]{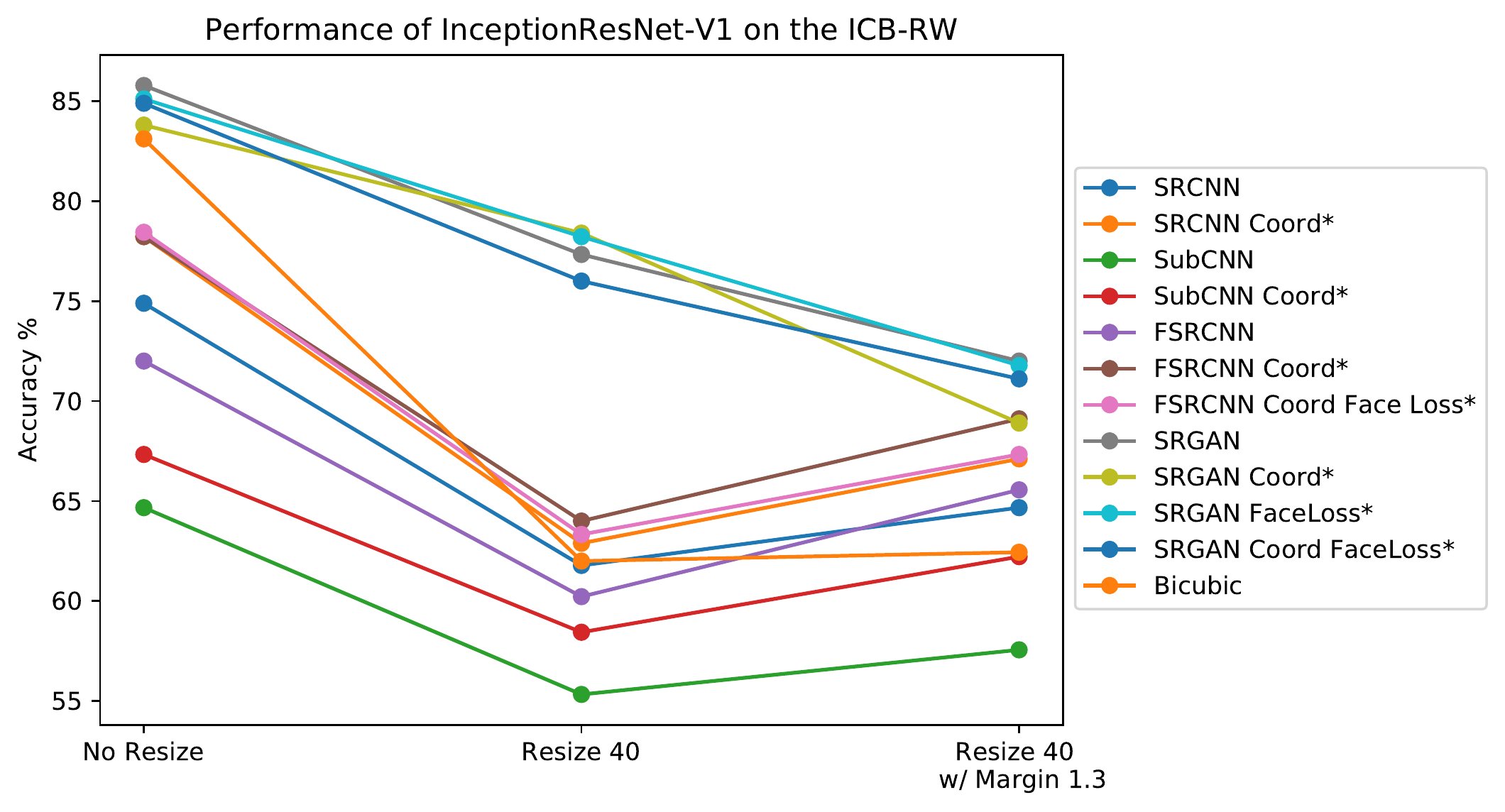}
\end{subfigure}%
\begin{subfigure}{.5\textwidth}
  \centering
  \includegraphics[width=1\linewidth]{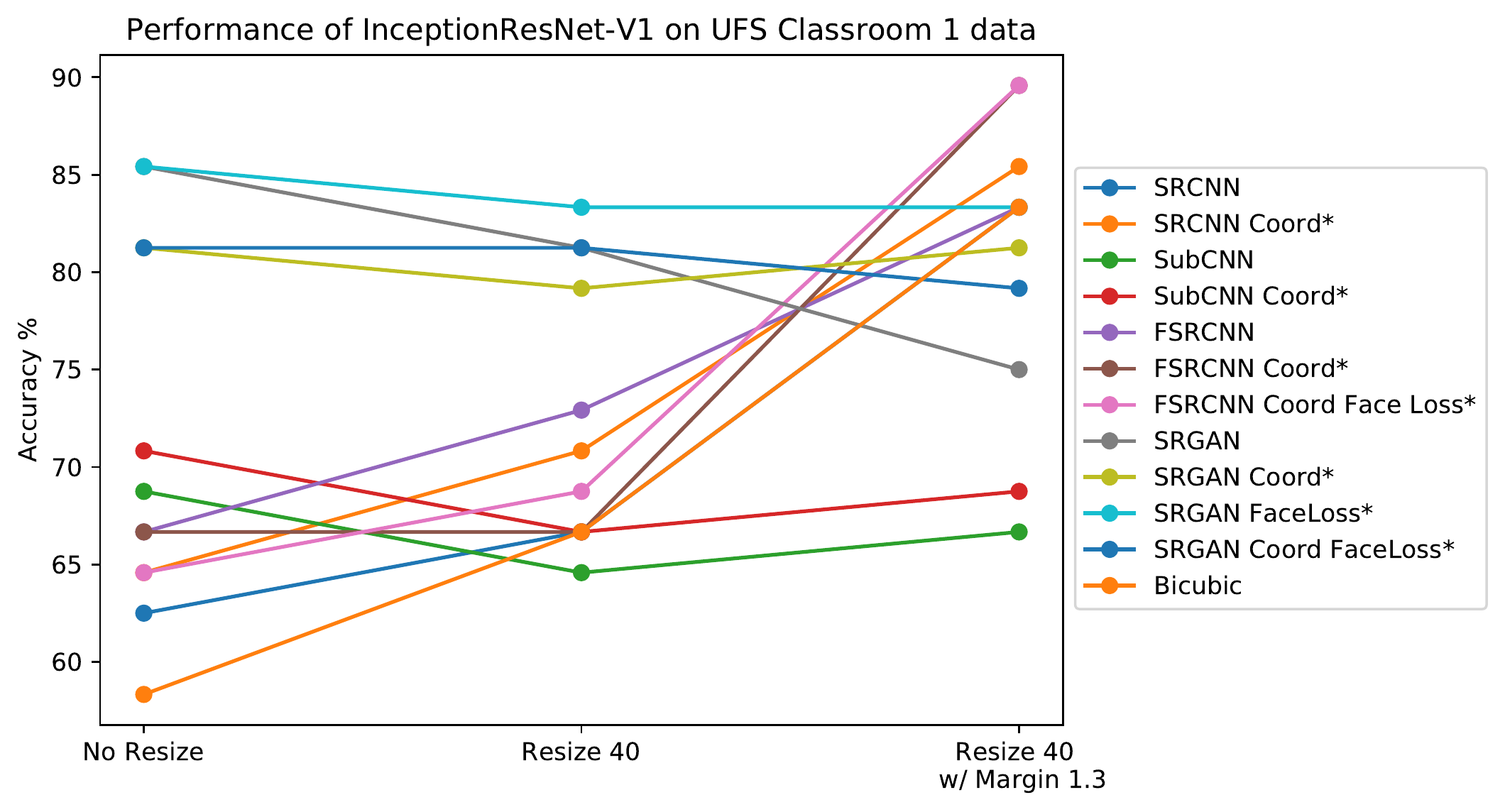}
\end{subfigure}
\caption{Performance results for accuracy on ICB-RW and UFS Clasroom 1 data. Proposed architectures have an asterisk in their names.(Source: Author's own)}
\label{fig:icb-rw}
\end{figure}

\begin{figure}[!htb]
\centering
\begin{subfigure}{.5\textwidth}
  \centering
  \includegraphics[width=1\linewidth]{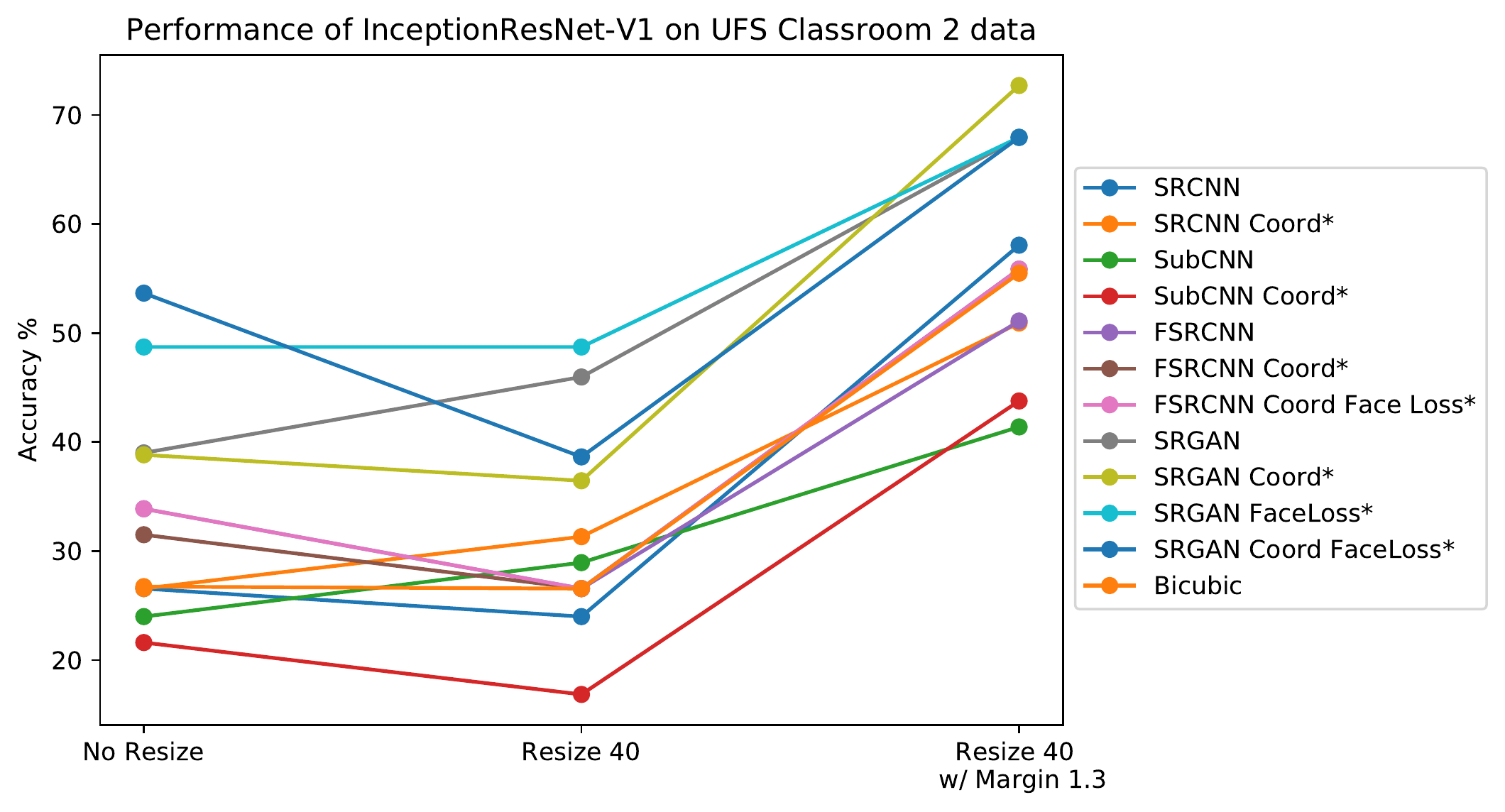}
\end{subfigure}%
\begin{subfigure}{.5\textwidth}
  \centering
  \includegraphics[width=1\linewidth]{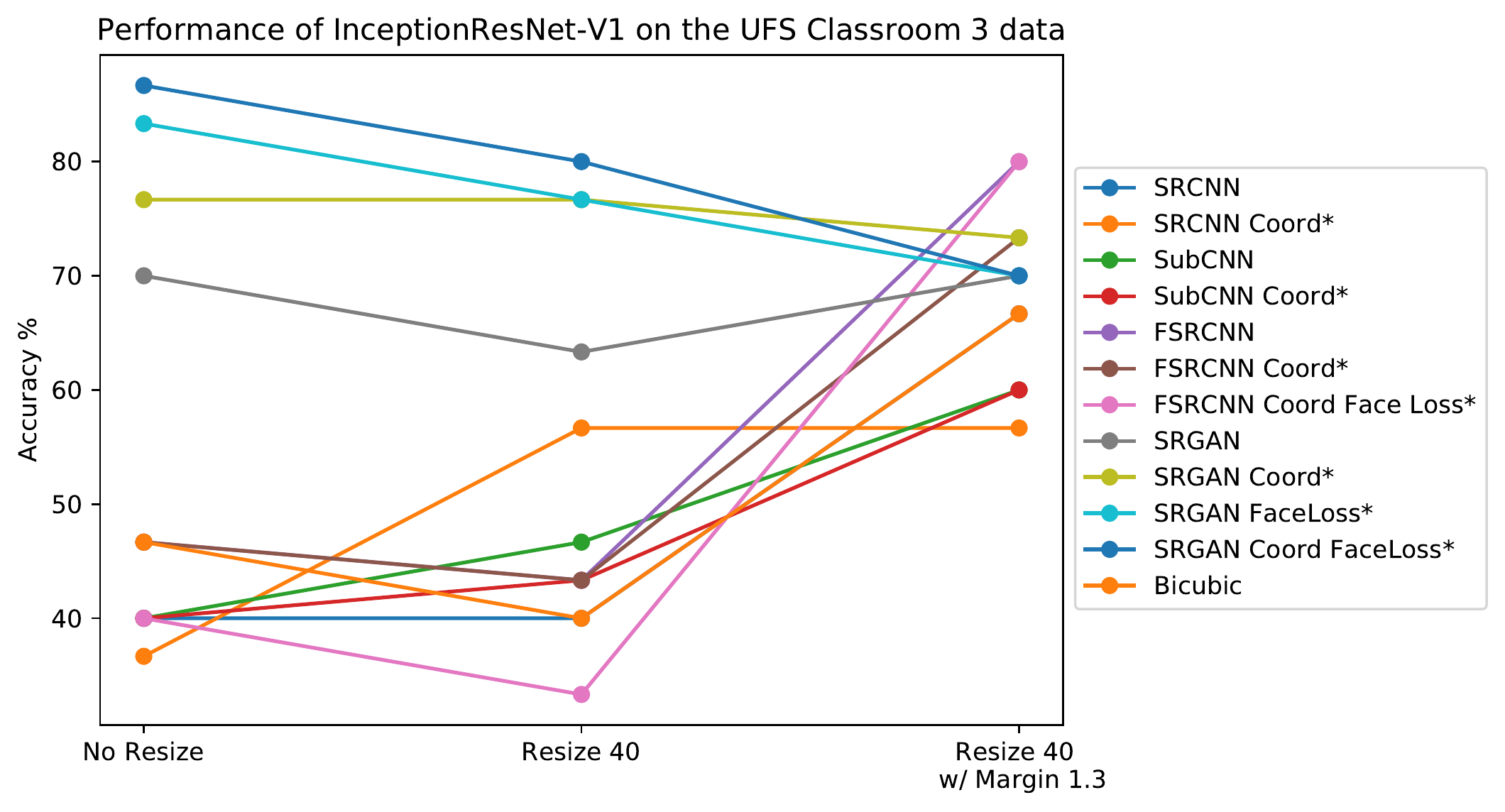}
\end{subfigure}
\caption{Performance results for accuracy on UFS Clasroom 2 and UFS Classroom 3 data. Proposed architectures have an asterisk in their names. (Source: Author's own)}
\label{fig:ufs-class}
\end{figure}

\subsection{Hypotheses Discussion}

For checking if there is correlation between image quality metrics and accuracy performance, the application of a correlation test was necessary. Since it was not possible to confirm if the original image data distribution approached normality, the Spearman Correlation Coefficient was calculated since it is specific for nonparametric data. Its results can be seen in Table~\ref{tab:correlation}.

\begin{table}[!htb]
\centering
\caption{Spearman Correlation PSNR/SSIM vs. Accuracy}
\label{tab:correlation}
\begin{tabular}{ccc}
\hline
                                                                              & PSNR       & SSIM   \\ \hline
\begin{tabular}[c]{@{}c@{}}Spearman \\ Correlation\\ Coefficient\end{tabular} & -0.3625    & 0.1159 \\
p-value                                                                       & 5.671 e-07 & 0.121  \\ \hline
\end{tabular}
\end{table}

As can be seen by Table \ref{tab:correlation}, the null hypothesis (there is no dependency) can be rejected in the case of SSIM, yet it was not the case for the PSNR since its p-value was less than 0.05. However, this result implies that there is a negative correlation involving PSNR. This outcome can be explained by the fact that the best performing models for accuracy (SRGAN and its variants) had the worst PSNR results when compared to the other models, which occasionally performed poorly in the face verification step.

Regarding accuracy, all architectures that took advantage of coordinate convolutions kept or increased its accuracy performance 72\% of the time for all experiments. However, even with the positive results, after applying the Wilcoxon Signed-Rank Test to check if the hypothesis of having CoordConvs brought substantial gains, the p-value value was equal to 0.083. This p-value was not sufficient to reject the null hypothesis, which may indicate that either the results data distributions were the same or there was not sufficient data to point their difference. Since the evaluation of different architectures was meant to be applied to real-world LR data, which in this case is limited, this CoordConv operator still needs to be further explored to have a real measure of its potential for the SR task. Even so, it has shown already promising results for general SR in face biometrics and can be employed for different architectures performing similar tasks. 

Architectures that were optimized with the specially designed for feature extraction ``FaceLoss'' kept or increased its accuracy performance 77\% of the time for all experiments. Also, they presented the highest average accuracy across all experiments and most of the best results, as it is shown in Figures~\ref{fig:icb-rw} and \ref{fig:ufs-class}. The Wilcoxon Signed-Rank Test presented a p-value of 0.03, which means that the null hypothesis (results have same distribution) can be rejected. Therefore, architectures with this loss had results from a different data distribution when compared to the same models trained on their general standard losses.

According to Figures~\ref{fig:icb-rw} and \ref{fig:ufs-class}, the verification accuracy increased when the 1.3 margin was applied to naturally LR images (UFS Classroom data) and decreased when probe images were in HR and already matched the gallery resolution (ICB-RW data). This suggests that a simple system that monitors what to apply to detected faces depending on their size could be elaborated to take advantage of this characteristic and provide significant improvements for a LR face recognition pipeline.

In general, deeper architectures (from literature and adapted) had better performance over the shallower ones and the baseline for the executed experiments. Even the simplest SR models (SRCNN, SubCNN, FSRCNN and their variants) were able to beat bicubic interpolation by at least a tiny margin, as shown in Figures~\ref{fig:icb-rw} and \ref{fig:ufs-class}. Nevertheless, this small margin may not justify the use of a simple DL model for upscaling images prior to recognition in real-world situations since the loss of FPS is still substantial and needs to be taken into consideration.

\section{Final Considerations}

In this chapter, the experiments elaborated for validating the initial hypotheses were presented, and the results regarding the proposed architectures were discussed. The use of deep SR models for enhancing image features before verification proved to be a beneficial step in the low-resolution face recognition pipeline. The models that made use of adversarial training (GANs) with different loss functions presented not only the best results regarding visual perception, as seen in Appendix~\ref{appendix:imgs}, but also recognition performance since they were able to produce a clearer image for feature extraction.

The final chapter presents the final considerations taking into account the whole thesis and the conclusions.

\chapter{Conclusions}

Super-resolution has shown vastly on recent works, and also reaffirmed in this thesis, both its ability to enhance the clarity and visual aspects of images and its potential to improve the accuracy performance of face recognition systems.

In this work, several state-of-the-art deep learning architectures for SR were implemented, modified and evaluated with the objective of enhancing face recognition performance in two naturally LR datasets. The application of SR models with the proposed change that addressed the use of an operator that takes into account the position information resulted in an equal or better accuracy 72\% of the time over the use of the same architectures without adaptation. Meanwhile, applications of SR architectures that were optimized with the loss that prioritized better feature extraction obtained a comparable or improved accuracy 77\% of the time.

The deeper networks (SRGAN and its variants) presented results that were both perceptually good for the human eye and the evaluation of the recognition criteria. The Inception ResNet V1 feature extractor with the proposed SRGAN FaceLoss architecture, the best performing SR model, had an average accuracy in all experiments of 73.54\%, which overcame the bicubic interpolation baseline by 17\%. Also, this same setup was able to achieve 85.11\% in the simulated ICB-RW dataset, which is an indication that such strategy might be able to defeat the SOTA model for the original benchmark since its accuracy was around 84\%.

The study performed in this thesis also confirmed that most of the other recently proposed SR deep learning architectures, even the not so deep ones, could be effective for recovering discriminant features of LR face images in real-world settings. In addition, the deeper network presented real-time capabilities when using a small GPU, which can facilitate their implementation for real-world surveillance systems.

Regarding the specific objectives elaborated for this thesis, some points are important to be highlighted:
\begin{itemize}
    \item Even though SR algorithms are usually optimized for upscaling an image and obtaining good PSNR metrics, not always this super-resolved image is going to present the most discriminative features for face recognition.
    \item Coordinated Convolutions presented gains for both image quality metrics and verification accuracy in the pipeline when applied to SR network architectures. However, they still need further studies since the data distribution for the results of networks with and without them presented high similarities, as confirmed by the Wilcoxon Signed-Rank Test.
    \item The use of a custom loss function to enhance the discriminative face features in images allows solid gains to the accuracy performance of a LR face recognition pipeline. 
\end{itemize}

As future work, a simple system can be proposed for applying the best margin to the crop size of a detected face to ensure there is enough information available for feature extraction. Notwithstanding, a comparative study should be necessary for evaluating in which size the recognition accuracy would start to drop.

Also, different strategies may be employed to tackle the deficiencies presented in an SR pipeline for LR face recognition. One of the problems is the need for a fixed upscale factor when training the SR network. This challenge may be tackled by the use of a meta-upscaling strategy based on the recent work of \citeonline{hu2019meta} where, for example, the weights of the upscaling network may be predicted based on the knowledge acquired by meta-features extracted from similar face datasets.

Another challenge to be solved is that some deep generative SR architectures may not be able to achieve real-time performance in CPU or mobile devices due to higher computational complexity. To overcome that, the use of distillation methods to prune these networks may improve time for inference to the cost of losing some performance~\cite{zhang2018adaptive}. 


\bibliography{Bibliografia}

\providecommand{\abntreprintinfo}[1]{%
 \citeonline{#1}}
\setlength{\labelsep}{0pt}\begin{thebibliography}{}
\providecommand{\abntrefinfo}[3]{}
\providecommand{\abntbstabout}[1]{}
\abntbstabout{v<VERSION> }

\bibitem[Abello and Jr. 2019]{abello2019optimizingSR}
\abntrefinfo{Abello and Jr.}{ABELLO; JR.}{2019}
{ABELLO, A.~A.; JR., R. H. Optimizing super resolution for face recognition.
  In:  SBC. \emph{SIBGRAPI Conference on Graphics, Patterns and Images
  (SIBGRAPI)}. [S.l.], 2019.}

\bibitem[Aghdam et al. 2019]{abdollahi2019exploring}
\abntrefinfo{Aghdam et al.}{AGHDAM et al.}{2019}
{AGHDAM, O. A. et al. Exploring factors for improving low resolution face
  recognition. In:  \emph{Proceedings of the IEEE Conference on Computer Vision
  and Pattern Recognition Workshops}. [S.l.: s.n.], 2019. p.~0--0.}

\bibitem[Agustsson and Timofte 2017]{Agustsson_2017_CVPR_Workshops}
\abntrefinfo{Agustsson and Timofte}{AGUSTSSON; TIMOFTE}{2017}
{AGUSTSSON, E.; TIMOFTE, R. Ntire 2017 challenge on single image
  super-resolution: Dataset and study. In:  \emph{The IEEE Conference on
  Computer Vision and Pattern Recognition (CVPR) Workshops}. [S.l.: s.n.],
  2017.}

\bibitem[Ataer-Cansizoglu et al. 2019]{ataer2019verification}
\abntrefinfo{Ataer-Cansizoglu et al.}{ATAER-CANSIZOGLU et al.}{2019}
{ATAER-CANSIZOGLU, E. et al. Verification of very low-resolution faces using an
  identity-preserving deep face super-resolution network.
\emph{arXiv preprint arXiv:1903.10974}, 2019.}

\bibitem[Baker and Kanade 2000]{baker2000hallucinating}
\abntrefinfo{Baker and Kanade}{BAKER; KANADE}{2000}
{BAKER, S.; KANADE, T. Hallucinating faces.
\emph{fg}, Citeseer, v.~2000, p. 83--88, 2000.}

\bibitem[Bao et al. 2017]{bao2017cvae}
\abntrefinfo{Bao et al.}{BAO et al.}{2017}
{BAO, J. et al. Cvae-gan: fine-grained image generation through asymmetric
  training. In:  \emph{Proceedings of the IEEE International Conference on
  Computer Vision}. [S.l.: s.n.], 2017. p. 2745--2754.}

\bibitem[Begin and Ferrie 2006]{begin2006comparison}
\abntrefinfo{Begin and Ferrie}{BEGIN; FERRIE}{2006}
{BEGIN, I.; FERRIE, F.~P. Comparison of super-resolution algorithms using image
  quality measures. In:  IEEE. \emph{The 3rd Canadian Conference on Computer
  and Robot Vision (CRV'06)}. [S.l.], 2006. p. 72--72.}

\bibitem[Berger, Peyrard and Baccouche 2016]{berger2016boosting}
\abntrefinfo{Berger, Peyrard and Baccouche}{BERGER; PEYRARD; BACCOUCHE}{2016}
{BERGER, G.; PEYRARD, C.; BACCOUCHE, M. Boosting face recognition via neural
  super-resolution. In:  \emph{ESANN}. [S.l.: s.n.], 2016.}

\bibitem[Cao et al. 2018]{cao2018vggface2}
\abntrefinfo{Cao et al.}{CAO et al.}{2018}
{CAO, Q. et al. Vggface2: A dataset for recognising faces across pose and age.
  In:  IEEE. \emph{2018 13th IEEE International Conference on Automatic Face \&
  Gesture Recognition (FG 2018)}. [S.l.], 2018. p. 67--74.}

\bibitem[Chang et al. 2017]{chang2017memory}
\abntrefinfo{Chang et al.}{CHANG et al.}{2017}
{CHANG, C.-H. et al. Memory and perception-based facial image reconstruction.
\emph{Scientific reports}, Nature Publishing Group, v.~7, n.~1, p.~6499, 2017.}

\bibitem[Chen et al. 2016]{chen2016deep}
\abntrefinfo{Chen et al.}{CHEN et al.}{2016}
{CHEN, Y. et al. Deep feature extraction and classification of hyperspectral
  images based on convolutional neural networks.
\emph{IEEE Transactions on Geoscience and Remote Sensing}, IEEE, v.~54, n.~10,
  p. 6232--6251, 2016.}

\bibitem[Chen et al. 2018]{Chen2018FSRNetEL}
\abntrefinfo{Chen et al.}{CHEN et al.}{2018}
{CHEN, Y. et al. Fsrnet: End-to-end learning face super-resolution with facial
  priors.
\emph{2018 IEEE/CVF Conference on Computer Vision and Pattern Recognition}, p.
  2492--2501, 2018.}

\bibitem[Cole et al. 2017]{cole2017synthesizing}
\abntrefinfo{Cole et al.}{COLE et al.}{2017}
{COLE, F. et al. Synthesizing normalized faces from facial identity features.
  In:  \emph{Proceedings of the IEEE Conference on Computer Vision and Pattern
  Recognition}. [S.l.: s.n.], 2017. p. 3703--3712.}

\bibitem[Crosswhite et al. 2018]{crosswhite2018template}
\abntrefinfo{Crosswhite et al.}{CROSSWHITE et al.}{2018}
{CROSSWHITE, N. et al. Template adaptation for face verification and
  identification.
\emph{Image and Vision Computing}, Elsevier, v.~79, p. 35--48, 2018.}

\bibitem[Deshpande 2017]{blogcnn2019}
\abntrefinfo{Deshpande}{DESHPANDE}{2017}
{DESHPANDE, A. \emph{A Beginner`s Guide To Understanding Convolutional Neural
  Networks}. 2017.
  \url{https://adeshpande3.github.io/A-Beginner's-Guide-To-Understanding-Convolutional-Neural-Networks/}.
Accessed: 2019-11-25.}

\bibitem[Donahue, McAuley and Puckette 2018]{donahue2018synthesizing}
\abntrefinfo{Donahue, McAuley and Puckette}{DONAHUE; MCAULEY; PUCKETTE}{2018}
{DONAHUE, C.; MCAULEY, J.; PUCKETTE, M. Synthesizing audio with generative
  adversarial networks.
\emph{arXiv preprint arXiv:1802.04208}, 2018.}

\bibitem[Dong et al. 2015]{dong2015image}
\abntrefinfo{Dong et al.}{DONG et al.}{2015}
{DONG, C. et al. Image super-resolution using deep convolutional networks.
\emph{IEEE transactions on pattern analysis and machine intelligence}, IEEE,
  v.~38, n.~2, p. 295--307, 2015.}

\bibitem[Dong, Loy and Tang 2016]{dong2016accelerating}
\abntrefinfo{Dong, Loy and Tang}{DONG; LOY; TANG}{2016}
{DONG, C.; LOY, C.~C.; TANG, X. Accelerating the super-resolution convolutional
  neural network. In:  SPRINGER. \emph{European conference on computer vision}.
  [S.l.], 2016. p. 391--407.}

\bibitem[ElSayed et al. 2018]{elsayed2018unsupervised}
\abntrefinfo{ElSayed et al.}{ELSAYED et al.}{2018}
{ELSAYED, A. et al. Unsupervised face recognition in the wild using
  high-dimensional features under super-resolution and 3d alignment effect.
\emph{Signal, Image and Video Processing}, Springer, v.~12, n.~7, p.
  1353--1360, 2018.}

\bibitem[Faceli et al. 2011]{faceli2011inteligencia}
\abntrefinfo{Faceli et al.}{FACELI et al.}{2011}
{FACELI, K. et al. Intelig{\^e}ncia artificial: Uma abordagem de aprendizado de
  m{\'a}quina.
2011.}

\bibitem[Feldstein 2019]{feldstein2019global}
\abntrefinfo{Feldstein}{FELDSTEIN}{2019}
{FELDSTEIN, S. The global expansion of ai surveillance.
\emph{Carnegie Endowment. https://carnegieendowment.
  org/2019/09/17/global-expansion-of-ai-surveillance-pub-79847}, 2019.}

\bibitem[Gecer et al. 2019]{gecer2019ganfit}
\abntrefinfo{Gecer et al.}{GECER et al.}{2019}
{GECER, B. et al. Ganfit: Generative adversarial network fitting for high
  fidelity 3d face reconstruction. In:  \emph{Proceedings of the IEEE
  Conference on Computer Vision and Pattern Recognition}. [S.l.: s.n.], 2019.
  p. 1155--1164.}

\bibitem[Gerchberg 1974]{gerchberg1974super}
\abntrefinfo{Gerchberg}{GERCHBERG}{1974}
{GERCHBERG, R. Super-resolution through error energy reduction.
\emph{Optica Acta: International Journal of Optics}, Taylor \& Francis, v.~21,
  n.~9, p. 709--720, 1974.}

\bibitem[Ghaleb et al. 2018]{ghaleb2018deep}
\abntrefinfo{Ghaleb et al.}{GHALEB et al.}{2018}
{GHALEB, E. et al. Deep representation and score normalization for face
  recognition under mismatched conditions.
\emph{Ieee Intelligent Systems}, IEEE, v.~33, n.~3, p. 43--46, 2018.}

\bibitem[Gobbini and Haxby 2007]{gobbini2007neural}
\abntrefinfo{Gobbini and Haxby}{GOBBINI; HAXBY}{2007}
{GOBBINI, M.~I.; HAXBY, J.~V. Neural systems for recognition of familiar faces.
\emph{Neuropsychologia}, Elsevier, v.~45, n.~1, p. 32--41, 2007.}

\bibitem[Goodfellow et al. 2014]{goodfellow2014generative}
\abntrefinfo{Goodfellow et al.}{GOODFELLOW et al.}{2014}
{GOODFELLOW, I. et al. Generative adversarial nets. In:  \emph{Advances in
  neural information processing systems}. [S.l.: s.n.], 2014. p. 2672--2680.}

\bibitem[Haris, Shakhnarovich and Ukita 2018]{haris2018task}
\abntrefinfo{Haris, Shakhnarovich and Ukita}{HARIS; SHAKHNAROVICH; UKITA}{2018}
{HARIS, M.; SHAKHNAROVICH, G.; UKITA, N. Task-driven super resolution: Object
  detection in low-resolution images.
\emph{arXiv preprint arXiv:1803.11316}, 2018.}

\bibitem[Haxby, Hoffman and Gobbini 2000]{haxby2000distributed}
\abntrefinfo{Haxby, Hoffman and Gobbini}{HAXBY; HOFFMAN; GOBBINI}{2000}
{HAXBY, J.~V.; HOFFMAN, E.~A.; GOBBINI, M.~I. The distributed human neural
  system for face perception.
\emph{Trends in cognitive sciences}, Elsevier, v.~4, n.~6, p. 223--233, 2000.}

\bibitem[He et al. 2016]{he2016deep}
\abntrefinfo{He et al.}{HE et al.}{2016}
{HE, K. et al. Deep residual learning for image recognition. In:
  \emph{Proceedings of the IEEE conference on computer vision and pattern
  recognition}. [S.l.: s.n.], 2016. p. 770--778.}

\bibitem[Hennings-Yeomans, Baker and Kumar 2008]{hennings2008simultaneous}
\abntrefinfo{Hennings-Yeomans, Baker and Kumar}{HENNINGS-YEOMANS; BAKER;
  KUMAR}{2008}
{HENNINGS-YEOMANS, P.~H.; BAKER, S.; KUMAR, B.~V. Simultaneous super-resolution
  and feature extraction for recognition of low-resolution faces. In:  IEEE.
  \emph{2008 IEEE Conference on Computer Vision and Pattern Recognition}.
  [S.l.], 2008. p.~1--8.}

\bibitem[Howard et al. 2018]{howard2018fastai}
\abntrefinfo{Howard et al.}{HOWARD et al.}{2018}
{HOWARD, J. et al. \emph{fastai}. [S.l.]: GitHub, 2018.
  \url{https://github.com/fastai/fastai}.}

\bibitem[Hu et al. 2019]{hu2019meta}
\abntrefinfo{Hu et al.}{HU et al.}{2019}
{HU, X. et al. Meta-sr: A magnification-arbitrary network for super-resolution.
  In:  \emph{Proceedings of the IEEE Conference on Computer Vision and Pattern
  Recognition}. [S.l.: s.n.], 2019. p. 1575--1584.}

\bibitem[Huang and Liu 2015]{huang2015short}
\abntrefinfo{Huang and Liu}{HUANG; LIU}{2015}
{HUANG, D.; LIU, H. A short survey of image super resolution algorithms.
\emph{Journal of Computer Science Technology Updates}, v.~2, n.~2, p. 19--29,
  2015.}

\bibitem[Huang et al. 2008]{huang2008labeled}
\abntrefinfo{Huang et al.}{HUANG et al.}{2008}
{HUANG, G.~B. et al. Labeled faces in the wild: A database for studying face
  recognition in unconstrained environments. In:  . [S.l.: s.n.], 2008.}

\bibitem[Jiao and Zhao 2019]{jiao2019survey}
\abntrefinfo{Jiao and Zhao}{JIAO; ZHAO}{2019}
{JIAO, L.; ZHAO, J. A survey on the new generation of deep learning in image
  processing.
\emph{IEEE Access}, IEEE, 2019.}

\bibitem[Johnson, Alahi and Fei-Fei 2016]{johnson2016perceptual}
\abntrefinfo{Johnson, Alahi and Fei-Fei}{JOHNSON; ALAHI; FEI-FEI}{2016}
{JOHNSON, J.; ALAHI, A.; FEI-FEI, L. Perceptual losses for real-time style
  transfer and super-resolution. In:  SPRINGER. \emph{European conference on
  computer vision}. [S.l.], 2016. p. 694--711.}

\bibitem[Johnson et al. 1991]{johnson1991newborns}
\abntrefinfo{Johnson et al.}{JOHNSON et al.}{1991}
{JOHNSON, M.~H. et al. Newborns' preferential tracking of face-like stimuli and
  its subsequent decline.
\emph{Cognition}, Elsevier, v.~40, n.~1-2, p. 1--19, 1991.}

\bibitem[Kim et al. 2019]{kim2019progressive}
\abntrefinfo{Kim et al.}{KIM et al.}{2019}
{KIM, D. et al. Progressive face super-resolution via attention to facial
  landmark.
\emph{arXiv preprint arXiv:1908.08239}, 2019.}

\bibitem[Krizhevsky, Sutskever and Hinton 2012]{krizhevsky2012imagenet}
\abntrefinfo{Krizhevsky, Sutskever and Hinton}{KRIZHEVSKY; SUTSKEVER;
  HINTON}{2012}
{KRIZHEVSKY, A.; SUTSKEVER, I.; HINTON, G.~E. Imagenet classification with deep
  convolutional neural networks. In:  \emph{Advances in neural information
  processing systems}. [S.l.: s.n.], 2012. p. 1097--1105.}

\bibitem[LeCun, Bengio and Hinton 2015]{lecun2015deep}
\abntrefinfo{LeCun, Bengio and Hinton}{LECUN; BENGIO; HINTON}{2015}
{LECUN, Y.; BENGIO, Y.; HINTON, G. Deep learning.
\emph{nature}, Nature Publishing Group, v.~521, n.~7553, p.~436, 2015.}

\bibitem[Ledig et al. 2017]{ledig2017photo}
\abntrefinfo{Ledig et al.}{LEDIG et al.}{2017}
{LEDIG, C. et al. Photo-realistic single image super-resolution using a
  generative adversarial network. In:  \emph{Proceedings of the IEEE conference
  on computer vision and pattern recognition}. [S.l.: s.n.], 2017. p.
  4681--4690.}

\bibitem[Li, Feng and Kuo 2018]{li2018deep}
\abntrefinfo{Li, Feng and Kuo}{LI; FENG; KUO}{2018}
{LI, J.; FENG, J.; KUO, C.-C.~J. Deep convolutional neural network for latent
  fingerprint enhancement.
\emph{Signal Processing: Image Communication}, Elsevier, v.~60, p. 52--63,
  2018.}

\bibitem[Li et al. 2019]{li2019low}
\abntrefinfo{Li et al.}{LI et al.}{2019}
{LI, P. et al. On low-resolution face recognition in the wild: Comparisons and
  new techniques.
\emph{IEEE Transactions on Information Forensics and Security}, IEEE, v.~14,
  n.~8, p. 2000--2012, 2019.}

\bibitem[Liu et al. 2018]{liu2018intriguing}
\abntrefinfo{Liu et al.}{LIU et al.}{2018}
{LIU, R. et al. An intriguing failing of convolutional neural networks and the
  coordconv solution. In:  \emph{Advances in Neural Information Processing
  Systems}. [S.l.: s.n.], 2018. p. 9605--9616.}

\bibitem[Liu et al. 2015]{liu2015faceattributes}
\abntrefinfo{Liu et al.}{LIU et al.}{2015}
{LIU, Z. et al. Deep learning face attributes in the wild. In:
  \emph{Proceedings of International Conference on Computer Vision (ICCV)}.
  [S.l.: s.n.], 2015.}

\bibitem[Ma et al. 2018]{ma2018gan}
\abntrefinfo{Ma et al.}{MA et al.}{2018}
{MA, S. et al. Da-gan: Instance-level image translation by deep attention
  generative adversarial networks. In:  \emph{Proceedings of the IEEE
  Conference on Computer Vision and Pattern Recognition}. [S.l.: s.n.], 2018.
  p. 5657--5666.}

\bibitem[Muttu and Virani 2015]{muttu2015effective}
\abntrefinfo{Muttu and Virani}{MUTTU; VIRANI}{2015}
{MUTTU, Y.; VIRANI, H. Effective face detection, feature extraction \& neural
  network based approaches for facial expression recognition. In:  IEEE.
  \emph{2015 International Conference on Information Processing (ICIP)}.
  [S.l.], 2015. p. 102--107.}

\bibitem[Neves, Moreno and Proen{\c{c}}a 2017]{neves2017quis}
\abntrefinfo{Neves, Moreno and Proen{\c{c}}a}{NEVES; MORENO;
  PROEN{\c{C}}A}{2017}
{NEVES, J.; MORENO, J.; PROEN{\c{C}}A, H. Quis-campi: an annotated
  multi-biometrics data feed from surveillance scenarios.
\emph{IET Biometrics}, IET, v.~7, n.~4, p. 371--379, 2017.}

\bibitem[Neves and Proen{\c{c}}a 2016]{neves2016icb}
\abntrefinfo{Neves and Proen{\c{c}}a}{NEVES; PROEN{\c{C}}A}{2016}
{NEVES, J.; PROEN{\c{C}}A, H. Icb-rw 2016: International challenge on biometric
  recognition in the wild. In:  IEEE. \emph{2016 International Conference on
  Biometrics (ICB)}. [S.l.], 2016. p.~1--6.}

\bibitem[Nguyen and Bai 2010]{nguyen2010cosine}
\abntrefinfo{Nguyen and Bai}{NGUYEN; BAI}{2010}
{NGUYEN, H.~V.; BAI, L. Cosine similarity metric learning for face
  verification. In:  SPRINGER. \emph{Asian conference on computer vision}.
  [S.l.], 2010. p. 709--720.}

\bibitem[Nguyen et al. 2018]{nguyen2018super}
\abntrefinfo{Nguyen et al.}{NGUYEN et al.}{2018}
{NGUYEN, K. et al. Super-resolution for biometrics: A comprehensive survey.
\emph{Pattern Recognition}, Elsevier, v.~78, p. 23--42, 2018.}

\bibitem[Ouyang et al. 2018]{ouyang2018deep}
\abntrefinfo{Ouyang et al.}{OUYANG et al.}{2018}
{OUYANG, N. et al. Deep joint super-resolution and feature mapping for low
  resolution face recognition. In:  IEEE. \emph{2018 IEEE International
  Conference of Safety Produce Informatization (IICSPI)}. [S.l.], 2018. p.
  849--852.}

\bibitem[Peyrard, Mamalet and Garcia 2015]{peyrard2015comparison}
\abntrefinfo{Peyrard, Mamalet and Garcia}{PEYRARD; MAMALET; GARCIA}{2015}
{PEYRARD, C.; MAMALET, F.; GARCIA, C. A comparison between multi-layer
  perceptrons and convolutional neural networks for text image
  super-resolution. In:  \emph{VISAPP (1)}. [S.l.: s.n.], 2015. p. 84--91.}

\bibitem[Purkait, Pal and Chanda 2014]{purkait2014fuzzy}
\abntrefinfo{Purkait, Pal and Chanda}{PURKAIT; PAL; CHANDA}{2014}
{PURKAIT, P.; PAL, N.~R.; CHANDA, B. A fuzzy-rule-based approach for single
  frame super resolution.
\emph{IEEE Transactions on Image processing}, IEEE, v.~23, n.~5, p. 2277--2290,
  2014.}

\bibitem[Rasti et al. 2016]{rasti2016convolutional}
\abntrefinfo{Rasti et al.}{RASTI et al.}{2016}
{RASTI, P. et al. Convolutional neural network super resolution for face
  recognition in surveillance monitoring. In:  SPRINGER. \emph{International
  conference on articulated motion and deformable objects}. [S.l.], 2016. p.
  175--184.}

\bibitem[Reibman, Bell and Gray 2006]{reibman2006quality}
\abntrefinfo{Reibman, Bell and Gray}{REIBMAN; BELL; GRAY}{2006}
{REIBMAN, A.~R.; BELL, R.~M.; GRAY, S. Quality assessment for super-resolution
  image enhancement. In:  IEEE. \emph{2006 International Conference on Image
  Processing}. [S.l.], 2006. p. 2017--2020.}

\bibitem[Ribeiro and Uhl 2017]{ribeiro2017exploring}
\abntrefinfo{Ribeiro and Uhl}{RIBEIRO; UHL}{2017}
{RIBEIRO, E.; UHL, A. Exploring texture transfer learning via convolutional
  neural networks for iris super resolution. In:  IEEE. \emph{2017
  International Conference of the Biometrics Special Interest Group (BIOSIG)}.
  [S.l.], 2017. p.~1--5.}

\bibitem[S\'a 2019]{joao2019Automatic}
\abntrefinfo{S\'a}{S\'A}{2019}
{S\'A, J. M. D. d.~C.
\emph{Registro de Classe Automatizado Utilizando Reconhecimento Facial}.
74~p. Bachelor's Thesis --- Universidade Federal de Sergipe, 2019.}

\bibitem[Schroff, Kalenichenko and Philbin 2015]{schroff2015facenet}
\abntrefinfo{Schroff, Kalenichenko and Philbin}{SCHROFF; KALENICHENKO;
  PHILBIN}{2015}
{SCHROFF, F.; KALENICHENKO, D.; PHILBIN, J. Facenet: A unified embedding for
  face recognition and clustering. In:  \emph{Proceedings of the IEEE
  conference on computer vision and pattern recognition}. [S.l.: s.n.], 2015.
  p. 815--823.}

\bibitem[Shi et al. 2016]{shi2016real}
\abntrefinfo{Shi et al.}{SHI et al.}{2016}
{SHI, W. et al. Real-time single image and video super-resolution using an
  efficient sub-pixel convolutional neural network. In:  \emph{Proceedings of
  the IEEE conference on computer vision and pattern recognition}. [S.l.:
  s.n.], 2016. p. 1874--1883.}

\bibitem[Szegedy et al. 2017]{szegedy2017inception}
\abntrefinfo{Szegedy et al.}{SZEGEDY et al.}{2017}
{SZEGEDY, C. et al. Inception-v4, inception-resnet and the impact of residual
  connections on learning. In:  \emph{Thirty-First AAAI Conference on
  Artificial Intelligence}. [S.l.: s.n.], 2017.}

\bibitem[Szegedy et al. 2015]{szegedy2015going}
\abntrefinfo{Szegedy et al.}{SZEGEDY et al.}{2015}
{SZEGEDY, C. et al. Going deeper with convolutions. In:  \emph{Proceedings of
  the IEEE conference on computer vision and pattern recognition}. [S.l.:
  s.n.], 2015. p.~1--9.}

\bibitem[Tian and Ma 2011]{tian2011survey}
\abntrefinfo{Tian and Ma}{TIAN; MA}{2011}
{TIAN, J.; MA, K.-K. A survey on super-resolution imaging.
\emph{Signal, Image and Video Processing}, Springer, v.~5, n.~3, p. 329--342,
  2011.}

\bibitem[Tian, Suzuki and Koike 2010]{tian2010task}
\abntrefinfo{Tian, Suzuki and Koike}{TIAN; SUZUKI; KOIKE}{2010}
{TIAN, L.; SUZUKI, A.; KOIKE, H. Task-oriented evaluation of super-resolution
  techniques. In:  IEEE. \emph{2010 20th International Conference on Pattern
  Recognition}. [S.l.], 2010. p. 493--498.}

\bibitem[Timofte et al. 2018]{Timofte_2018_CVPR_Workshops}
\abntrefinfo{Timofte et al.}{TIMOFTE et al.}{2018}
{TIMOFTE, R. et al. Ntire 2018 challenge on single image super-resolution:
  Methods and results. In:  \emph{The IEEE Conference on Computer Vision and
  Pattern Recognition (CVPR) Workshops}. [S.l.: s.n.], 2018.}

\bibitem[Upadhyay, Singhal and Singh 2019]{upadhyay2019spinal}
\abntrefinfo{Upadhyay, Singhal and Singh}{UPADHYAY; SINGHAL; SINGH}{2019}
{UPADHYAY, U.; SINGHAL, B.; SINGH, M. Spinal stenosis detection in mri using
  modular coordinate convolutional attention networks. In:  IEEE. \emph{2019
  International Joint Conference on Neural Networks (IJCNN)}. [S.l.], 2019.
  p.~1--8.}

\bibitem[Vedadi and Shirani 2014]{vedadi2014map}
\abntrefinfo{Vedadi and Shirani}{VEDADI; SHIRANI}{2014}
{VEDADI, F.; SHIRANI, S. A map-based image interpolation method via viterbi
  decoding of markov chains of interpolation functions.
\emph{IEEE Transactions on Image Processing}, IEEE, v.~23, n.~1, p. 424--438,
  2014.}

\bibitem[Vezhnevets 2002]{detection2003}
\abntrefinfo{Vezhnevets}{VEZHNEVETS}{2002}
{VEZHNEVETS, V. Face and facial feature tracking for natural human-computer
  interface. In:  . [S.l.: s.n.], 2002.}

\bibitem[Wang and Deng 2018]{wang2018deep}
\abntrefinfo{Wang and Deng}{WANG; DENG}{2018}
{WANG, M.; DENG, W. Deep face recognition: A survey.
\emph{arXiv preprint arXiv:1804.06655}, 2018.}

\bibitem[Wang et al. 2016]{wang2016studying}
\abntrefinfo{Wang et al.}{WANG et al.}{2016}
{WANG, Z. et al. Studying very low resolution recognition using deep networks.
  In:  \emph{Proceedings of the IEEE Conference on Computer Vision and Pattern
  Recognition}. [S.l.: s.n.], 2016. p. 4792--4800.}

\bibitem[Wang, Chen and Hoi 2019]{wang2019deep}
\abntrefinfo{Wang, Chen and Hoi}{WANG; CHEN; HOI}{2019}
{WANG, Z.; CHEN, J.; HOI, S.~C. Deep learning for image super-resolution: A
  survey.
\emph{arXiv preprint arXiv:1902.06068}, 2019.}

\bibitem[Wang, She and Ward 2019]{wang2019generative}
\abntrefinfo{Wang, She and Ward}{WANG; SHE; WARD}{2019}
{WANG, Z.; SHE, Q.; WARD, T.~E. Generative adversarial networks: A survey and
  taxonomy.
\emph{arXiv preprint arXiv:1906.01529}, 2019.}

\bibitem[Xu, Chen and Jia 2019]{xu2019view}
\abntrefinfo{Xu, Chen and Jia}{XU; CHEN; JIA}{2019}
{XU, X.; CHEN, Y.-C.; JIA, J. View independent generative adversarial network
  for novel view synthesis. In:  \emph{Proceedings of the IEEE International
  Conference on Computer Vision}. [S.l.: s.n.], 2019. p. 7791--7800.}

\bibitem[Yang et al. 2017]{yang2017semi}
\abntrefinfo{Yang et al.}{YANG et al.}{2017}
{YANG, Z. et al. Semi-supervised qa with generative domain-adaptive nets.
\emph{arXiv preprint arXiv:1702.02206}, 2017.}

\bibitem[Yu et al. 2018]{yu2018generative}
\abntrefinfo{Yu et al.}{YU et al.}{2018a}
{YU, J. et al. Generative image inpainting with contextual attention. In:
  \emph{Proceedings of the IEEE Conference on Computer Vision and Pattern
  Recognition}. [S.l.: s.n.], 2018. p. 5505--5514.}

\bibitem[Yu et al. 2018]{yu2018face}
\abntrefinfo{Yu et al.}{YU et al.}{2018b}
{YU, X. et al. Face super-resolution guided by facial component heatmaps. In:
  \emph{Proceedings of the European Conference on Computer Vision (ECCV)}.
  [S.l.: s.n.], 2018. p. 217--233.}

\bibitem[Zafeiriou, Zhang and Zhang 2015]{zafeiriou2015survey}
\abntrefinfo{Zafeiriou, Zhang and Zhang}{ZAFEIRIOU; ZHANG; ZHANG}{2015}
{ZAFEIRIOU, S.; ZHANG, C.; ZHANG, Z. A survey on face detection in the wild:
  past, present and future.
\emph{Computer Vision and Image Understanding}, Elsevier, v.~138, p. 1--24,
  2015.}

\bibitem[Zafeirouli et al. 2019]{zafeirouli2019efficient}
\abntrefinfo{Zafeirouli et al.}{ZAFEIROULI et al.}{2019}
{ZAFEIROULI, K. et al. Efficient, lightweight, coordinate-based network for
  image super resolution. In:  IEEE. \emph{2019 IEEE International Conference
  on Engineering, Technology and Innovation (ICE/ITMC)}. [S.l.], 2019.
  p.~1--9.}

\bibitem[Zhang and Zhang 2010]{zhang2010survey}
\abntrefinfo{Zhang and Zhang}{ZHANG; ZHANG}{2010}
{ZHANG, C.; ZHANG, Z. A survey of recent advances in face detection.
2010.}

\bibitem[Zhang et al. 2016]{zhang2016joint}
\abntrefinfo{Zhang et al.}{ZHANG et al.}{2016}
{ZHANG, K. et al. Joint face detection and alignment using multitask cascaded
  convolutional networks.
\emph{IEEE Signal Processing Letters}, IEEE, v.~23, n.~10, p. 1499--1503,
  2016.}

\bibitem[Zhang et al. 2018]{zhang2018adaptive}
\abntrefinfo{Zhang et al.}{ZHANG et al.}{2018}
{ZHANG, L. et al. Adaptive importance learning for improving lightweight image
  super-resolution network.
\emph{arXiv preprint arXiv:1806.01576}, 2018.}

\end{thebibliography}

\postextual

\renewcommand{\chapnumfont}{\chaptitlefont}
\renewcommand{\afterchapternum}{}
\begin{apendicesenv}

\partapendices


\chapter{Perceptual Results of SR Algorithms for 4x Upscaling}
\label{appendix:imgs}

In this section it is possible to check the perceptual results for the SR experiment described in Section~\ref{sec:exp1}. All the images were upscaled from a grid of 40x40 to a size of 160x160.

\begin{figure}[!htb]
	\begin{center}
		\includegraphics[width=\textwidth,height=18cm]{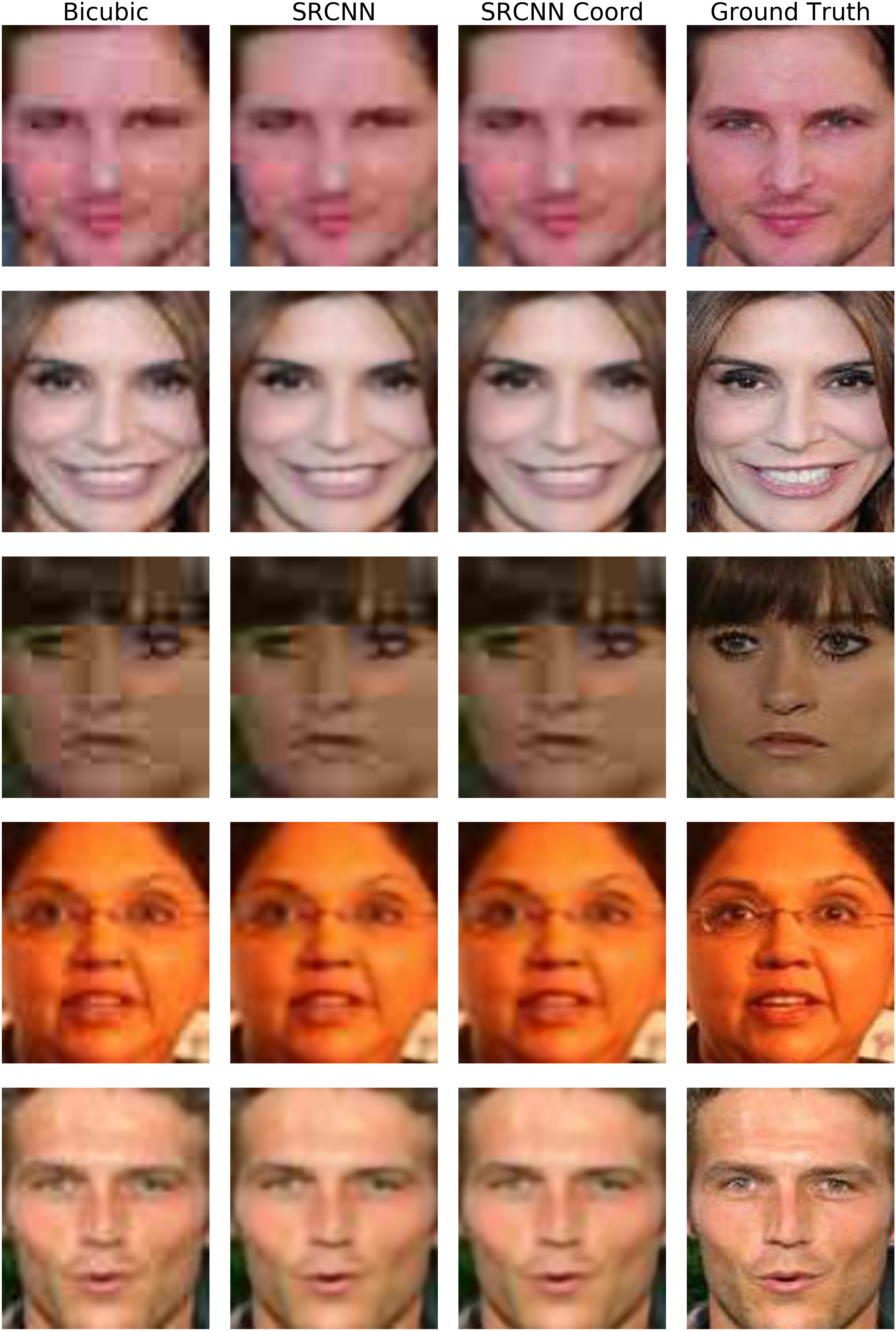}
	\end{center}
\end{figure}

\begin{figure}[!htb]
	\begin{center}
		\includegraphics[width=\textwidth]{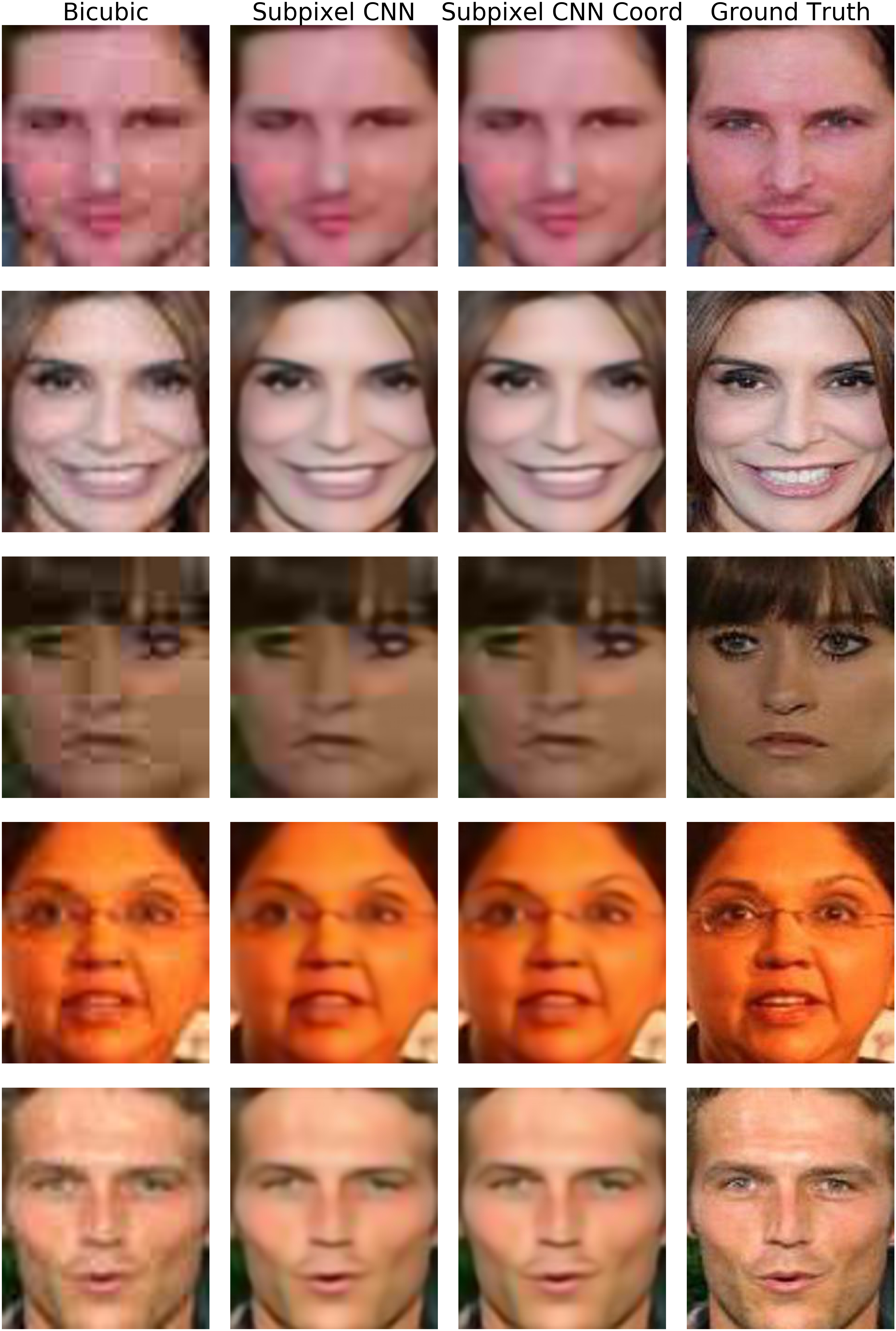}
	\end{center}
\end{figure}

\begin{figure}[!htb]
	\begin{center}
		\includegraphics[width=\textwidth]{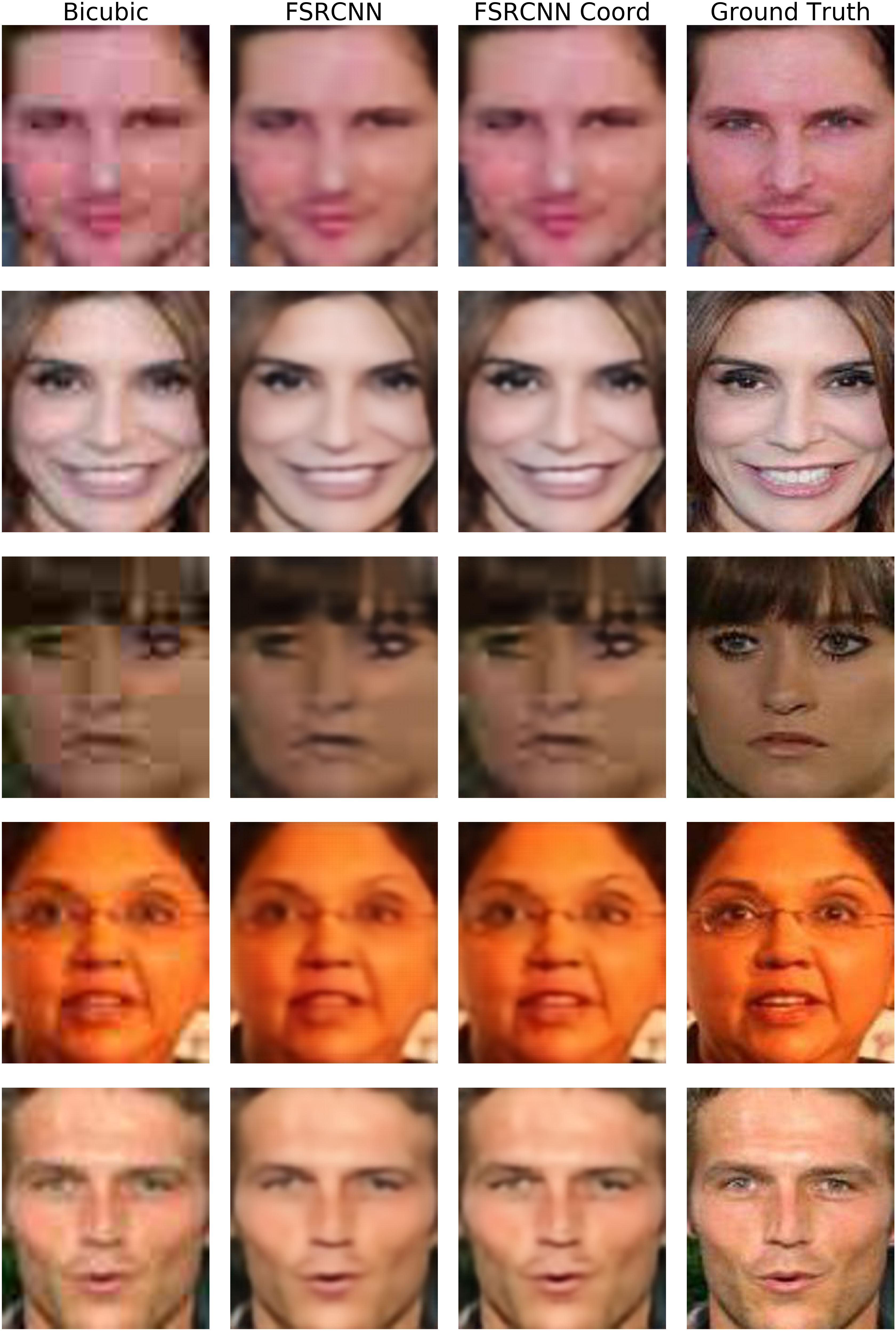}
	\end{center}
\end{figure}

\begin{figure}[!htb]
	\begin{center}
		\includegraphics[width=\textwidth]{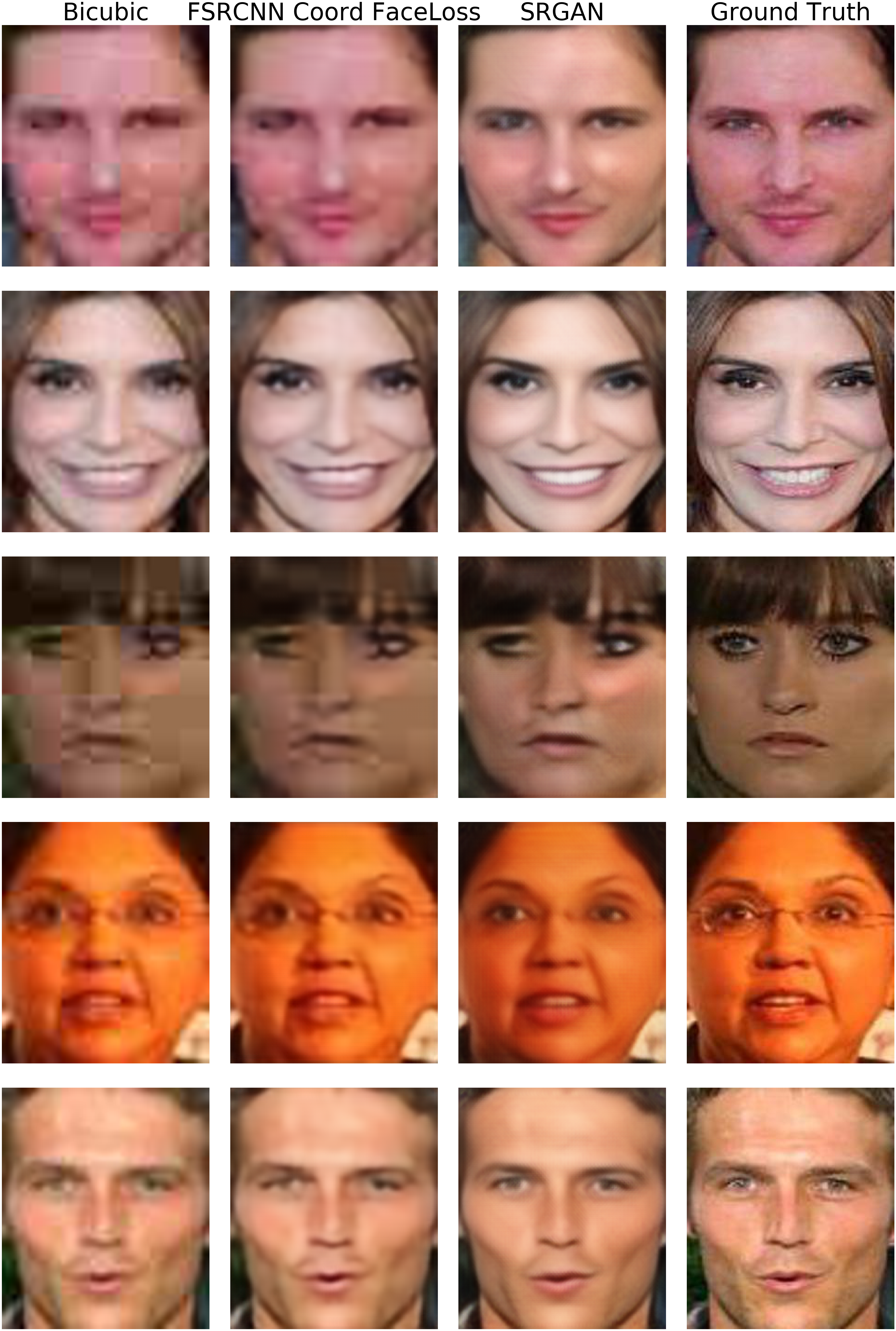}
	\end{center}
\end{figure}

\begin{figure}[!htb]
	\begin{center}
		\includegraphics[width=\textwidth]{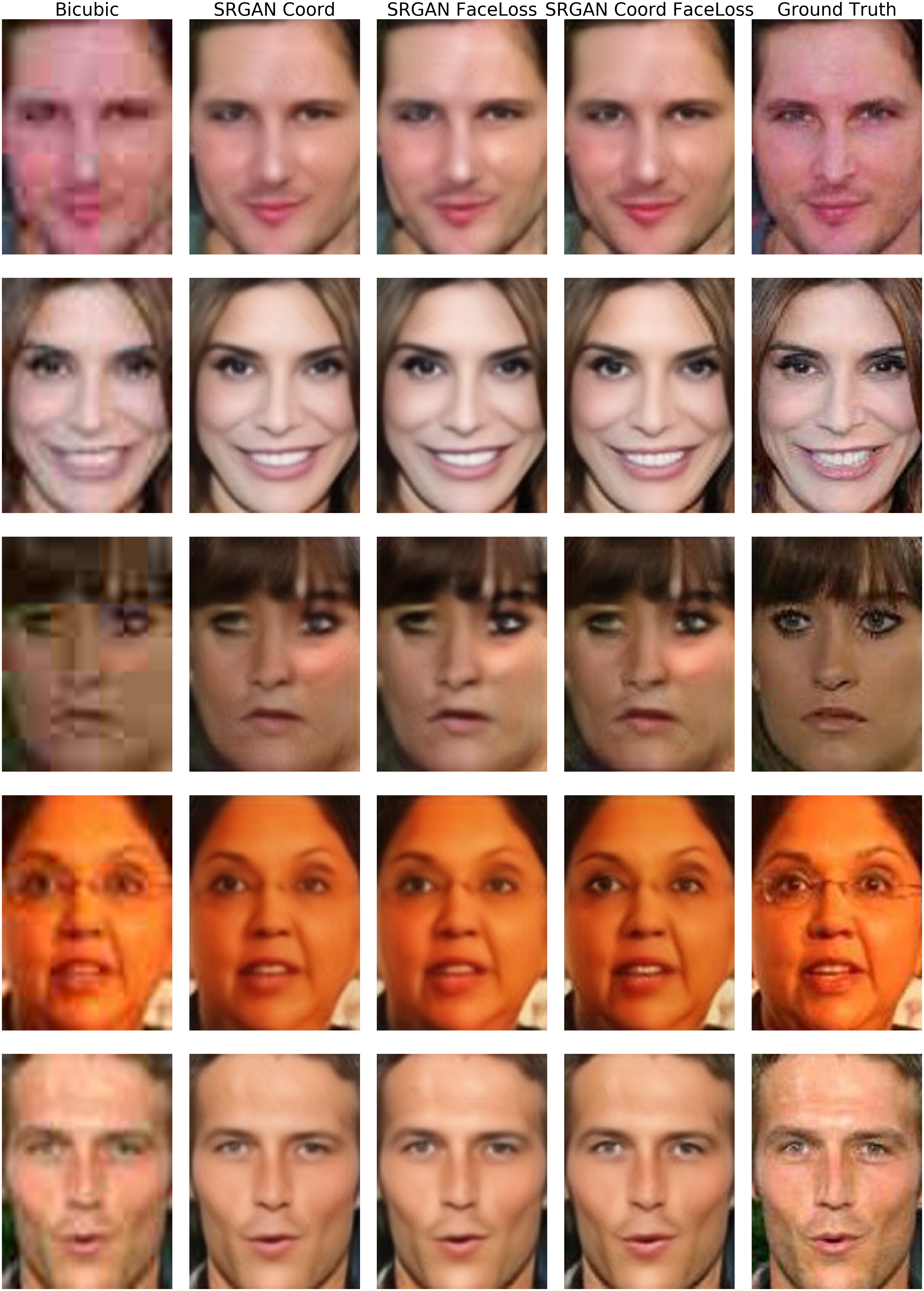}
	\end{center}
\end{figure}


\chapter{Average Training Losses for the SR Algorithms}
\label{appendix:graph}

This section presents the behavior for the training losses of each network obtained during the experiment described in Section \ref{sec:exp1}.

\begin{figure}[!htb]
  \centering
  \begin{minipage}[b]{0.49\textwidth}
    \includegraphics[width=\textwidth]{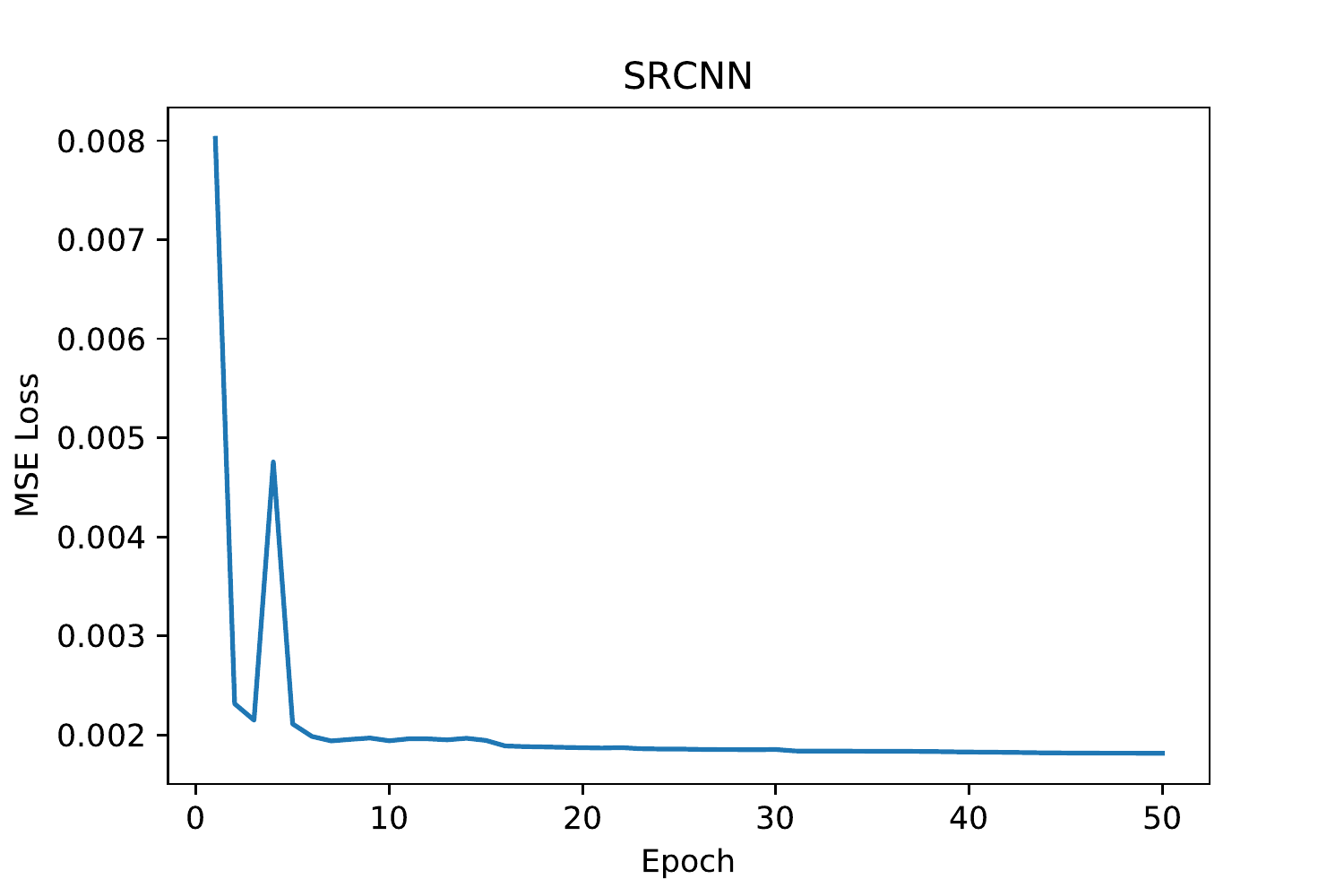}
  \end{minipage}
  \hfill
  \begin{minipage}[b]{0.49\textwidth}
    \includegraphics[width=\textwidth]{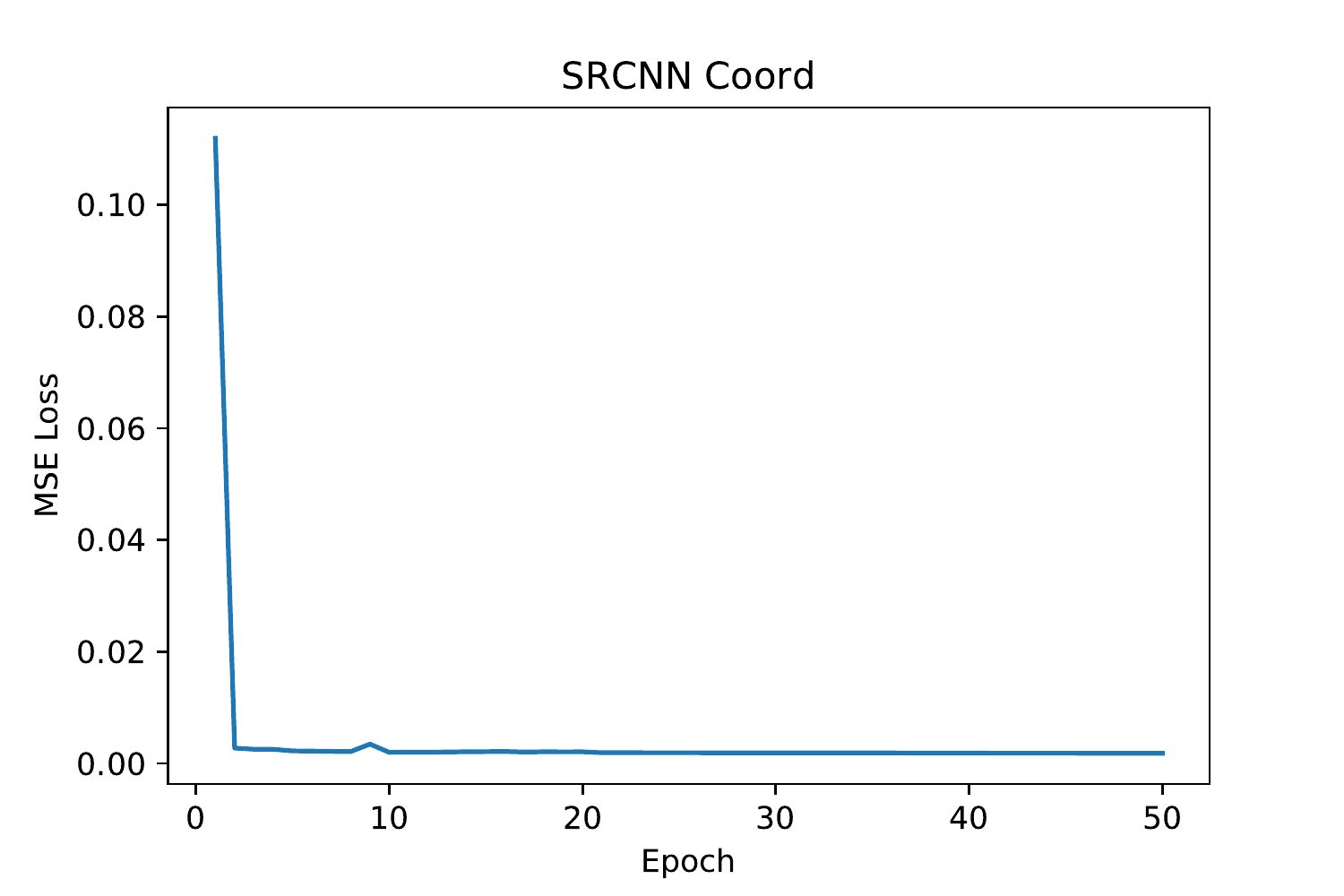}
  \end{minipage}
\end{figure}

\begin{figure}[!htb]
  \centering
  \begin{minipage}[b]{0.49\textwidth}
    \includegraphics[width=\textwidth]{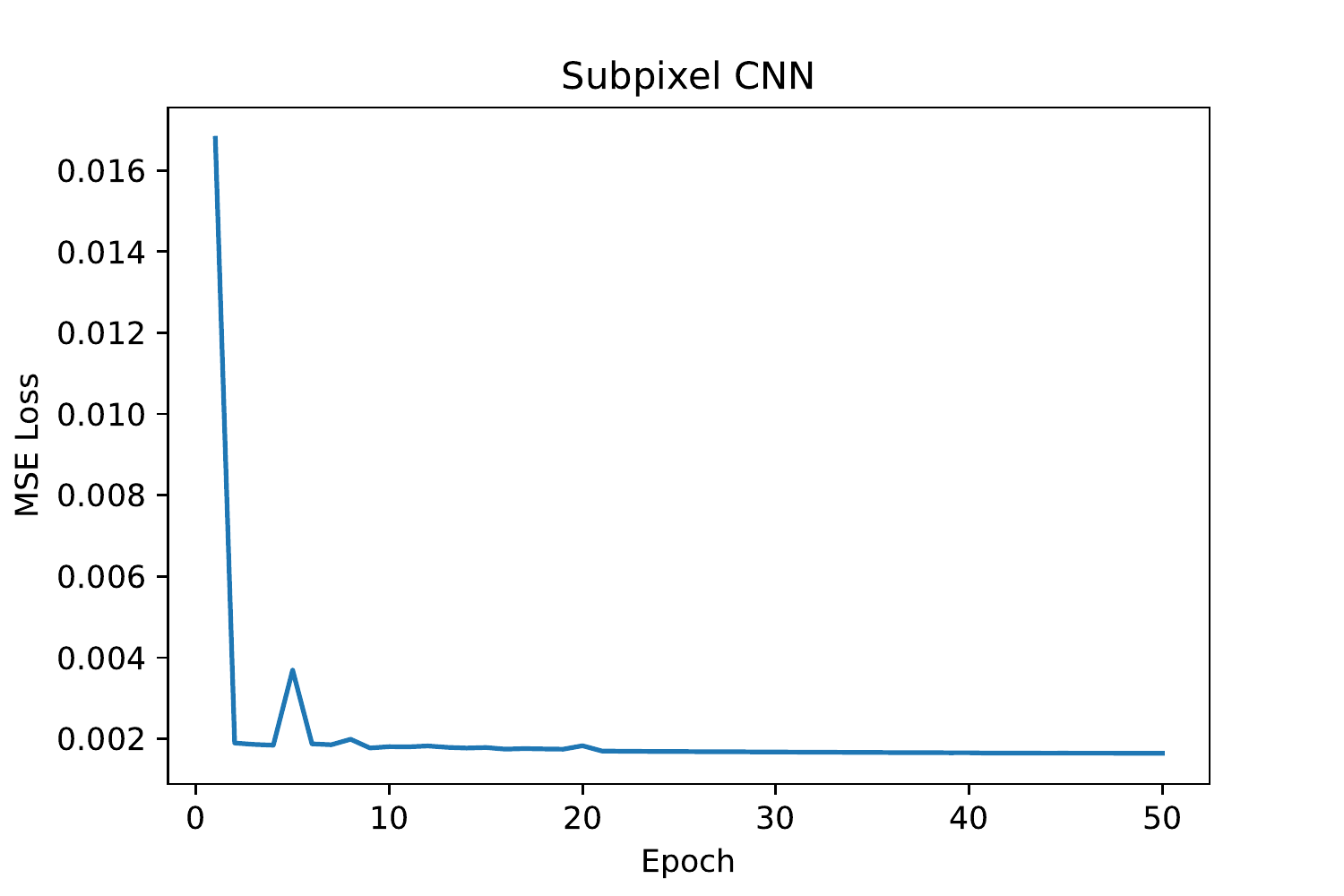}
  \end{minipage}
  \hfill
  \begin{minipage}[b]{0.49\textwidth}
    \includegraphics[width=\textwidth]{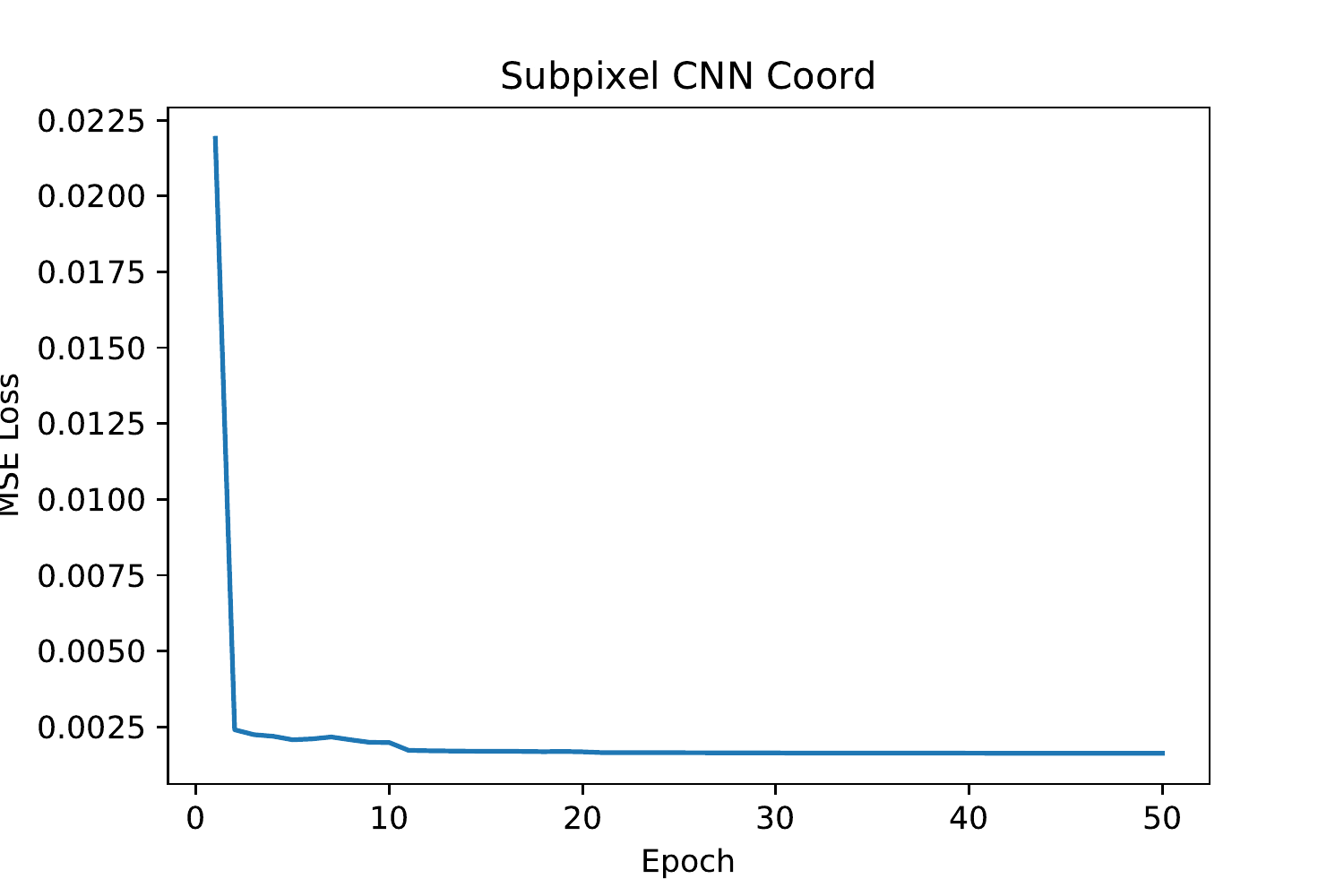}
  \end{minipage}
\end{figure}

\begin{figure}[!htb]
  \centering
  \begin{minipage}[b]{0.49\textwidth}
    \includegraphics[width=\textwidth]{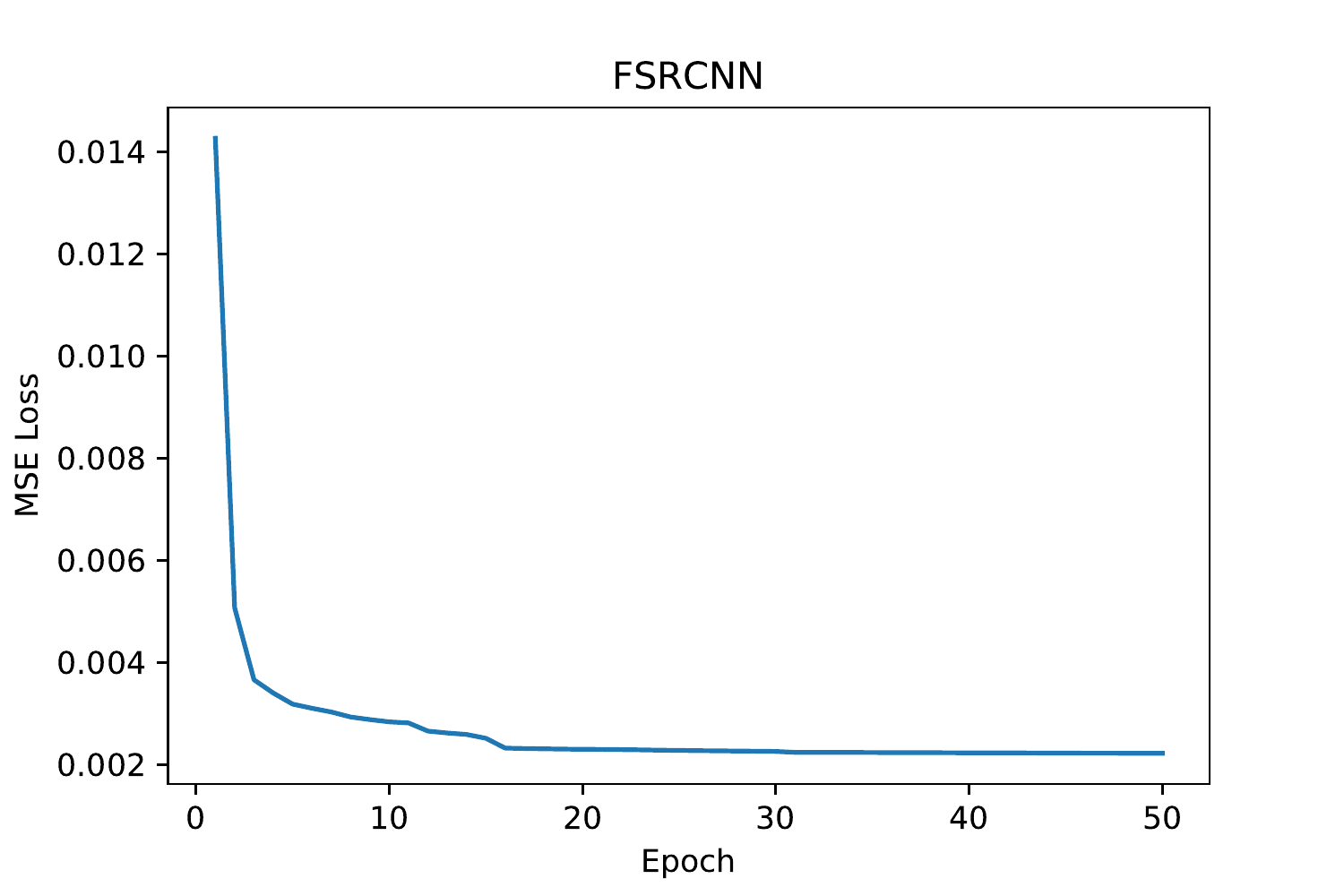}
  \end{minipage}
  \hfill
  \begin{minipage}[b]{0.49\textwidth}
    \includegraphics[width=\textwidth]{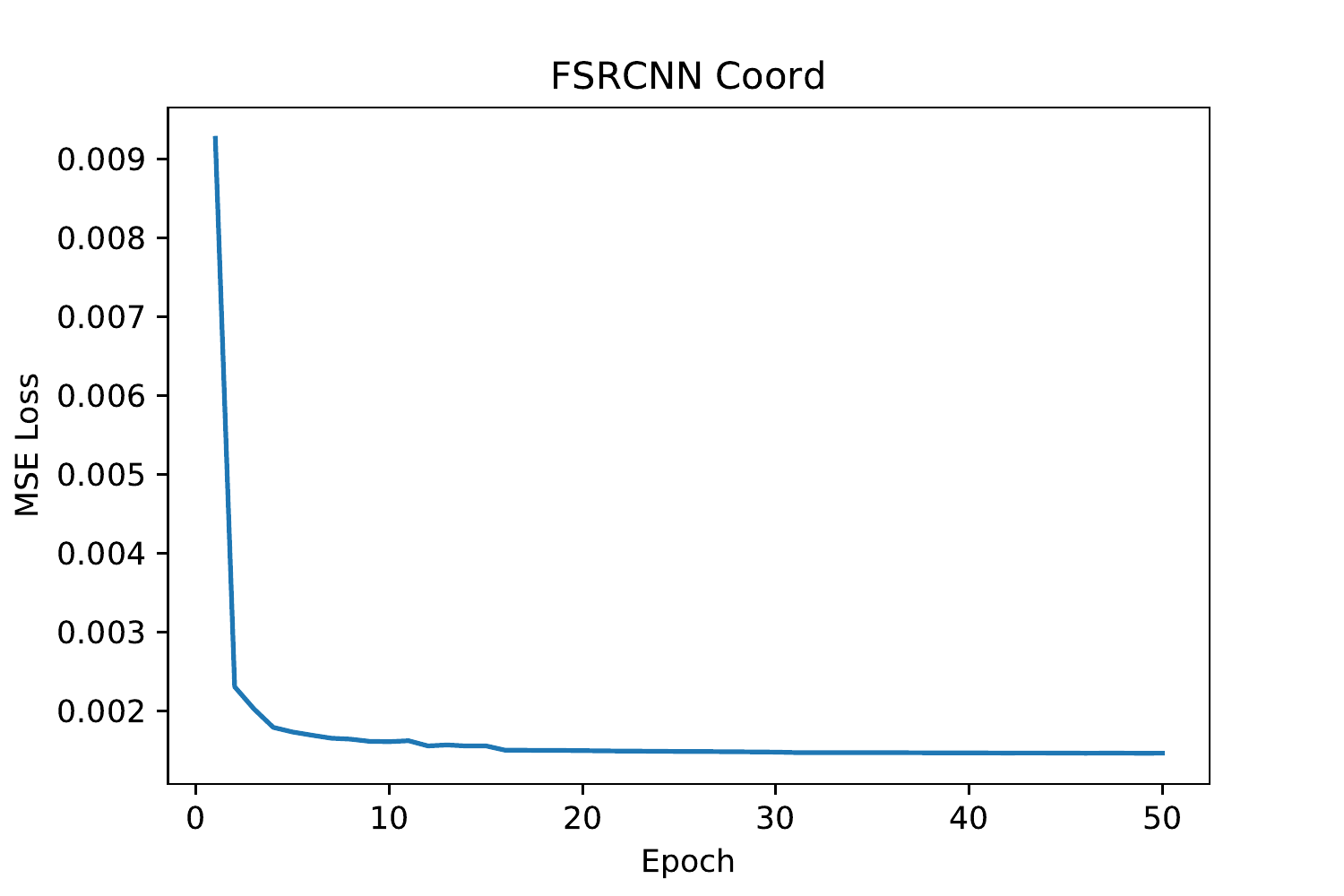}
  \end{minipage}
\end{figure}

\begin{figure}[!htb]
  \centering
  \begin{minipage}[b]{0.49\textwidth}
    \includegraphics[width=\textwidth]{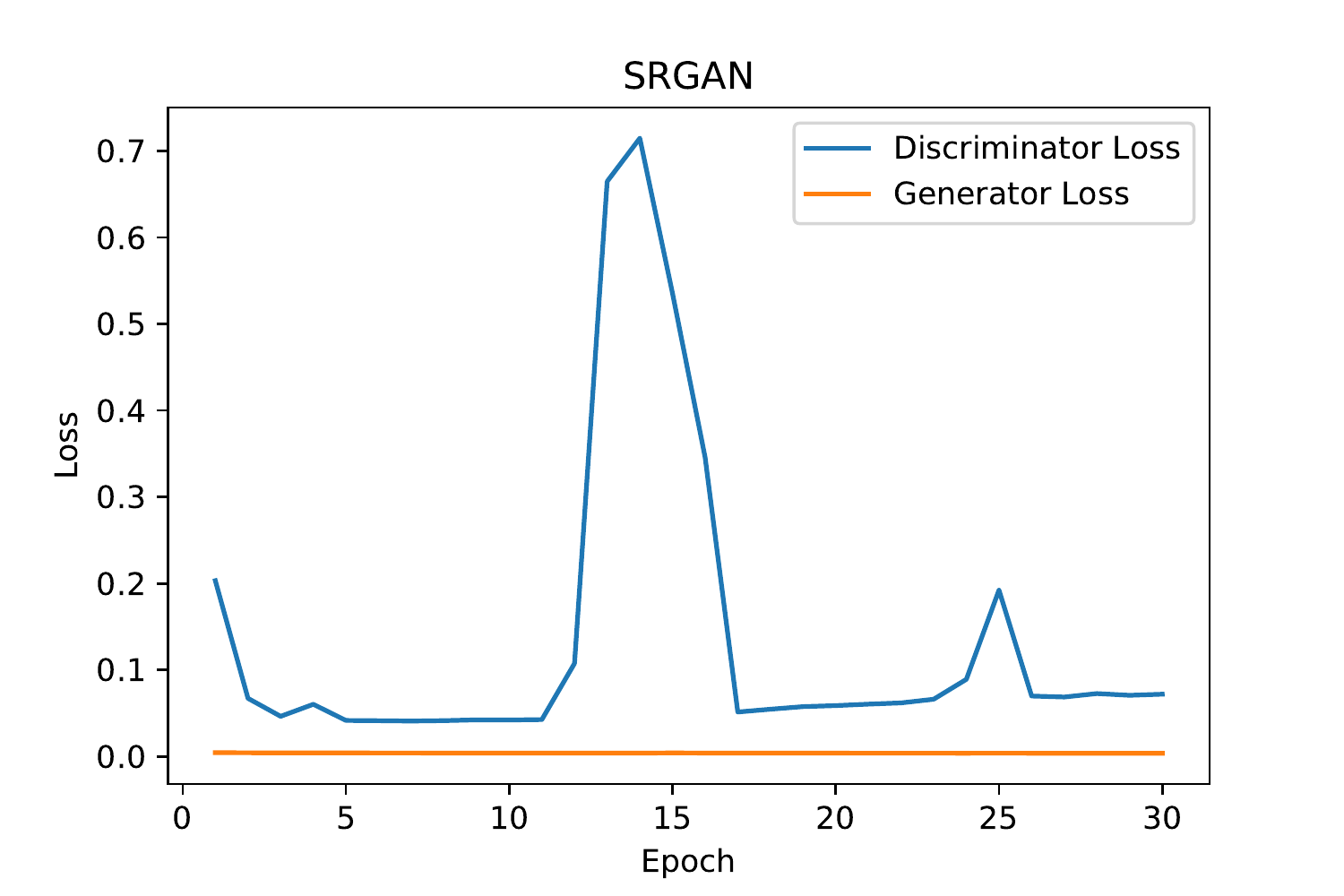}
  \end{minipage}
  \hfill
  \begin{minipage}[b]{0.49\textwidth}
    \includegraphics[width=\textwidth]{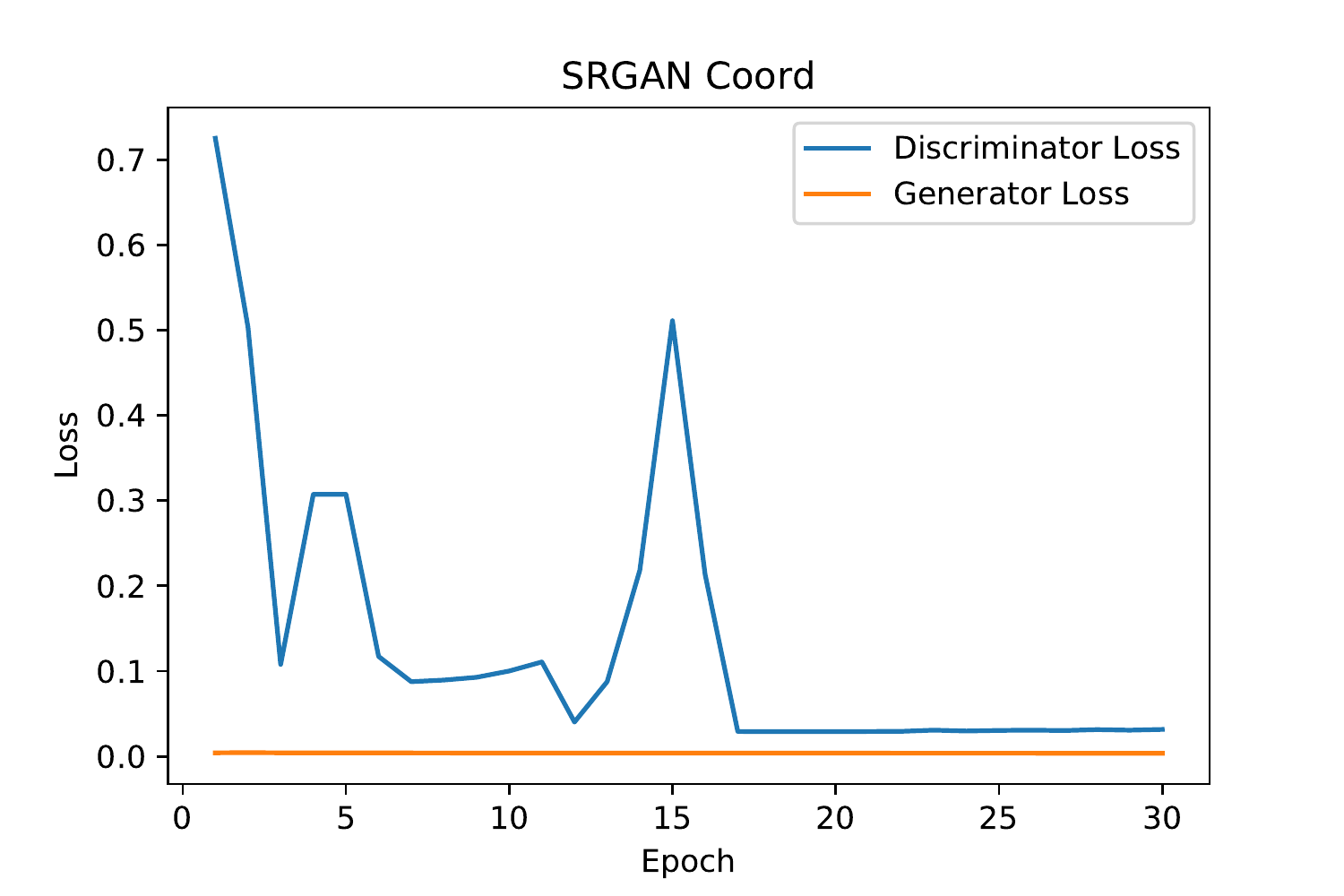}
  \end{minipage}
\end{figure}

\begin{figure}[!htb]
	\begin{center}
		\includegraphics[width=.49\textwidth]{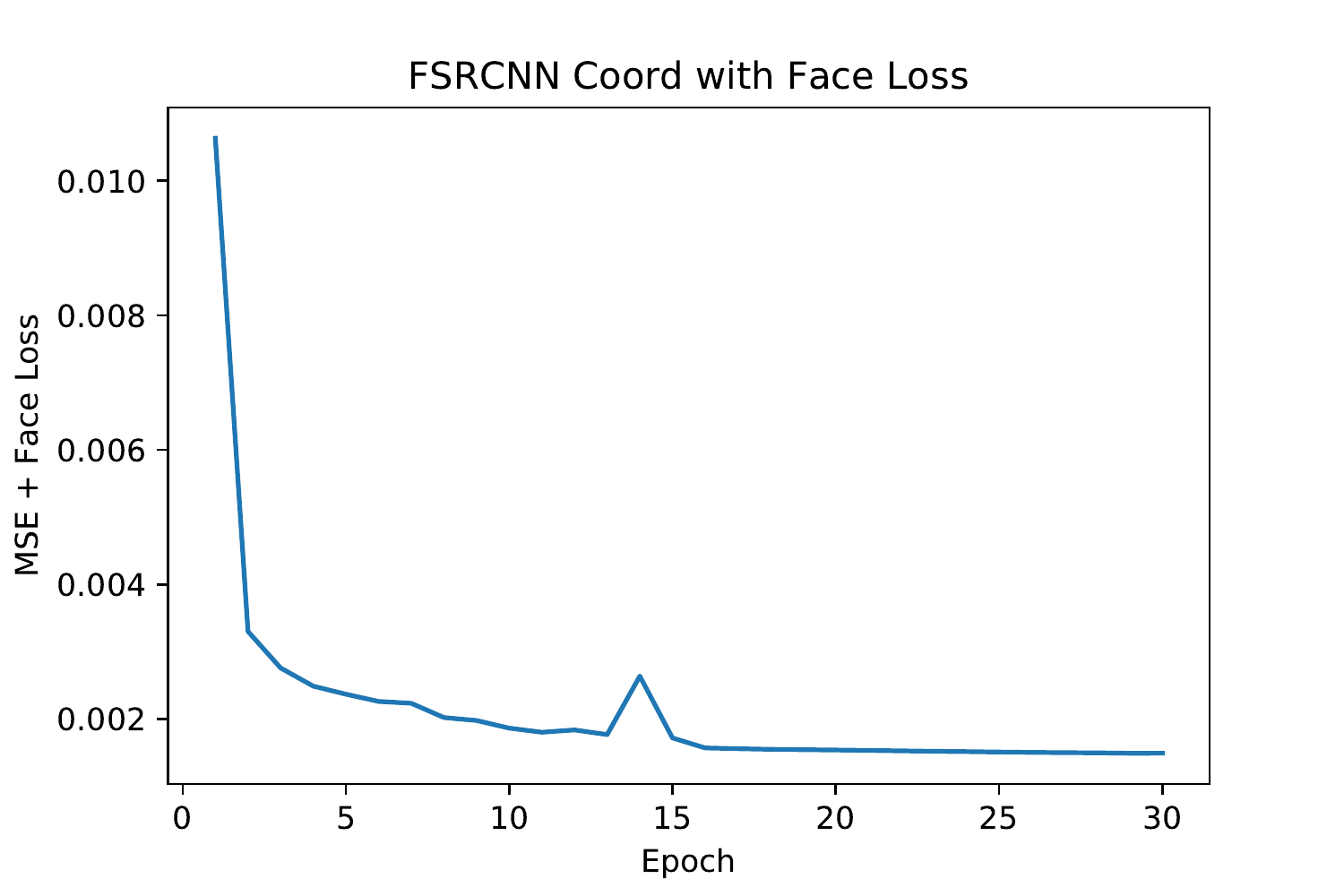}
	\end{center}
\end{figure}

\begin{figure}[!htb]
  \centering
  \begin{minipage}[b]{0.49\textwidth}
    \includegraphics[width=\textwidth]{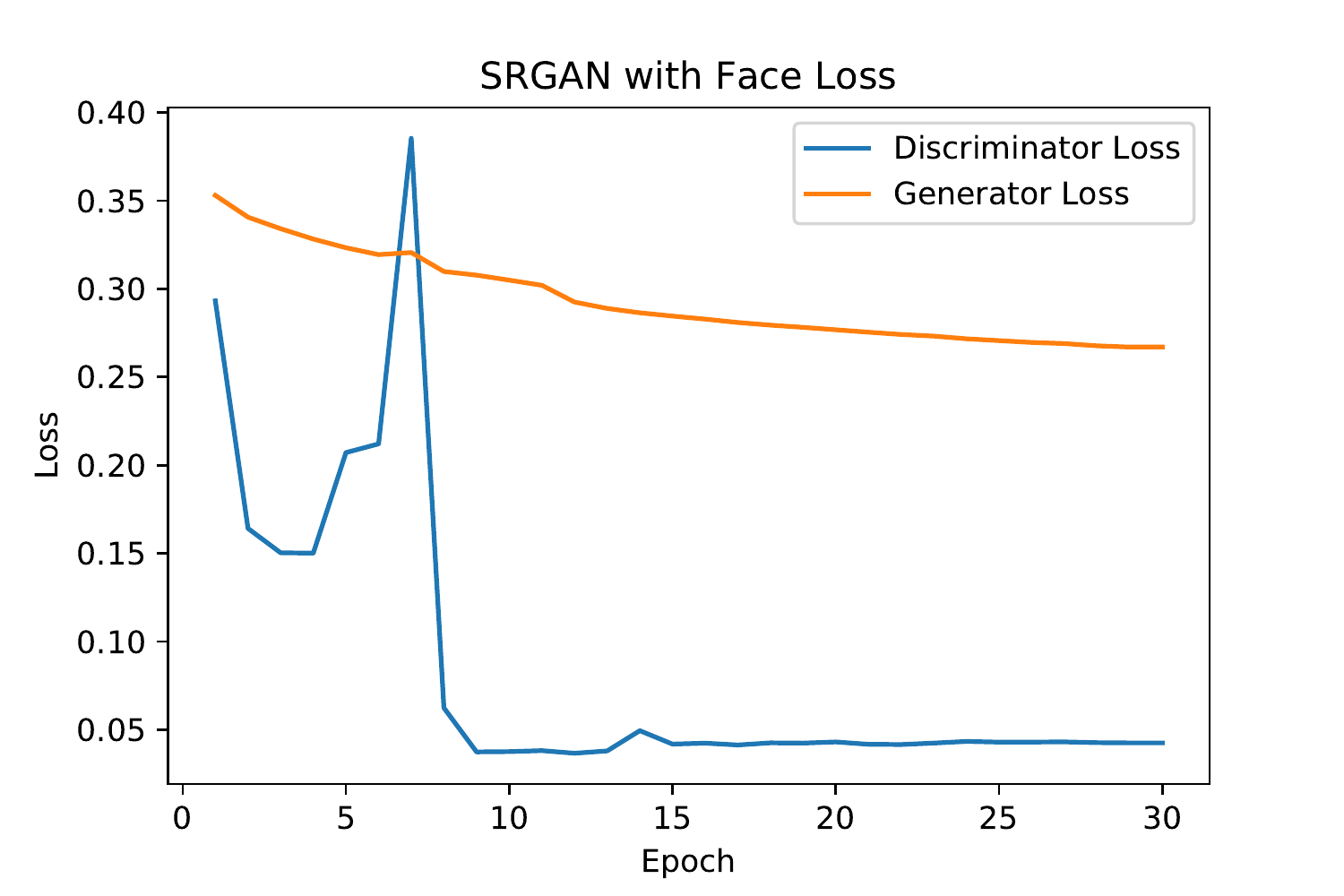}
  \end{minipage}
  \hfill
  \begin{minipage}[b]{0.49\textwidth}
    \includegraphics[width=\textwidth]{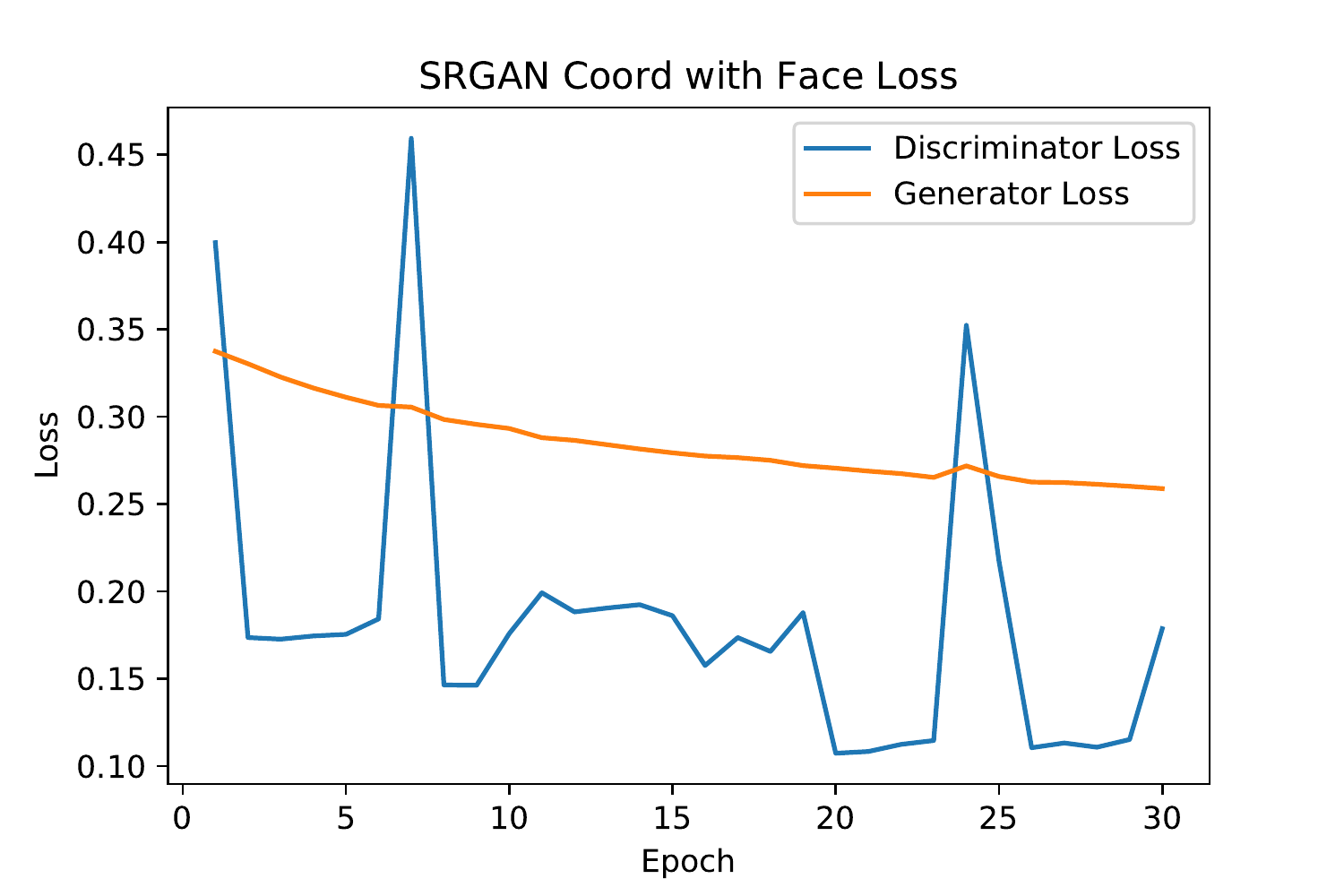}
  \end{minipage}
\end{figure}

\end{apendicesenv}

\end{document}